\documentclass[twoside,11pt]{article}

\usepackage{jmlr2e}
\usepackage[frozencache=true,cachedir=minted-cache]{minted}
\usepackage{amsmath}
\usepackage{amsfonts}
\usepackage{fancyvrb}
\usepackage{caption}
\usepackage{subcaption}
\usepackage{tikz}
\usetikzlibrary{arrows,matrix,positioning}
\usetikzlibrary{calc}

\newcommand{\tikzmark}[1]{\tikz[overlay,remember picture] \node (#1) {};}
\newcommand{\DrawBoxA}[1][]{%
    \tikz[overlay,remember picture]{
    \draw[red,#1]
      ($(left1)+(-0.2em,1.1em)$) rectangle
      ($(right1)+(0.7em,-0.3em)$);}
}
\newcommand{\DrawBoxB}[1][]{%
    \tikz[overlay,remember picture]{
    \draw[blue,#1]
      ($(left2)+(-0.5em,0.9em)$) rectangle
      ($(right2)+(0.5em,-0.5em)$);}
}
\newcommand{\DrawBoxC}[1][]{%
    \tikz[overlay,remember picture]{
    \draw[teal,#1]
      ($(left3)+(-0.0em,1.2em)$) rectangle
      ($(right3)+(0.5em,-0.2em)$);}
}
\newcommand{\DrawBoxD}[1][]{%
    \tikz[overlay,remember picture]{
    \draw[orange,#1]
      ($(left4)+(-0.2em,0.9em)$) rectangle
      ($(right4)+(0.7em,-0.4em)$);}
}

\newcommand{\fracpartial}[2]{\frac{\partial #1}{\partial  #2}}

\newcommand{\bdot}[2]{\langle #1, #2\rangle}
\newcommand{\eye}{\mathbf{o}}
\newcommand{\bx}{\mathbf{x}}

\newcommand{\Rho}{\mathrm{P}}

\ShortHeadings{$FC^2T^2$: The Fast Continuous Convolutional Taylor Transform}{Lange and Kutz}
\firstpageno{1}

\begin{document}

\title{$\mathbf{FC^2T^2}$: The Fast Continuous Convolutional Taylor Transform with Applications in Vision and Graphics}

\author{\name Henning Lange \email helange@uw.edu \\
       \addr Department of Applied Mathematics\\
       University of Washington\\
       Seattle, WA 98195-4322, USA
       \AND
       \name J. Nathan Kutz \email kutz@uw.edu \\
       \addr Department of Applied Mathematics\\
       University of Washington\\
       Seattle, WA 98195-4322, USA}

\editor{Someone}

\maketitle

\begin{abstract}
Series expansions have been a cornerstone of applied mathematics and engineering for centuries. In this paper, we revisit the Taylor series expansion from a modern Machine Learning perspective. Specifically, we introduce the Fast Continuous Convolutional Taylor Transform ($FC^2T^2$), a variant of the Fast Multipole Method (FMM), that allows for the efficient approximation of low dimensional convolutional operators in continuous space. We build upon the FMM which is an approximate algorithm that reduces the computational complexity of N-body problems from $\mathcal{O}(NM)$ to $\mathcal{O}(N+M)$ and finds application in e.g. particle simulations. As an intermediary step, the FMM produces a series expansion for every cell on a grid and we introduce algorithms that act directly upon this representation. These algorithms analytically but approximately compute the quantities required for the forward and backward pass of the backpropagation algorithm and can therefore be employed as (implicit) layers in Neural Networks. Specifically, we introduce a root-implicit layer that outputs surface normals and object distances as well as an integral-implicit layer that outputs a rendering of a radiance field given a 3D pose. In the context of Machine Learning, $N$ and $M$ can be understood as the number of model parameters and model evaluations respectively which entails that, for applications that require repeated function evaluations which are prevalent in Computer Vision and Graphics, unlike regular Neural Networks, the techniques introduce in this paper scale gracefully with parameters. For some applications, this results in a 200x reduction in FLOPs compared to state-of-the-art approaches at a reasonable or non-existent loss in accuracy.\footnote{A video abstract is available at: \url{https://youtu.be/e6gXoMA5te4}} 
\end{abstract}


\section{Introduction}

\begin{figure}
    \centering
    \begin{subfigure}[t]{0.3\textwidth}
        \centering
        \includegraphics[width=0.95\linewidth]{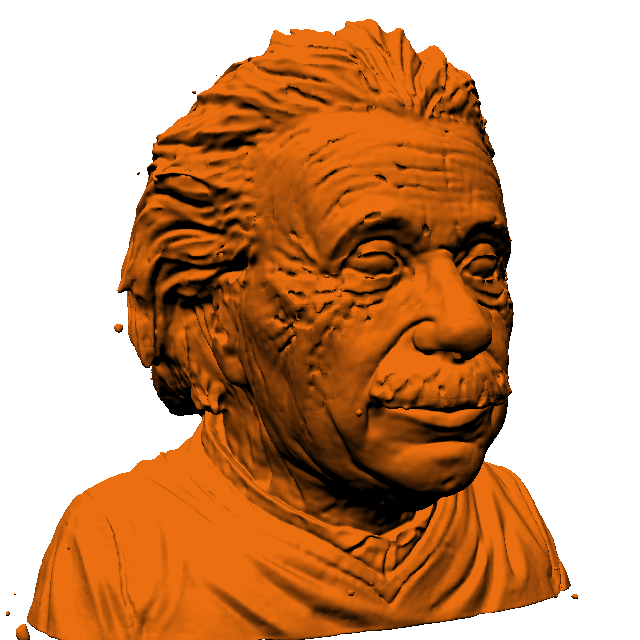} 
        \caption{\textbf{Explicit Layer:} Trained on 10m samples from a signed distance function for 1000 epochs in approximately 20min in total.} \label{fig1:eplicit}
    \end{subfigure}
    \hfill
    \begin{subfigure}[t]{0.3\textwidth}
        \centering
        \includegraphics[width=0.95\linewidth]{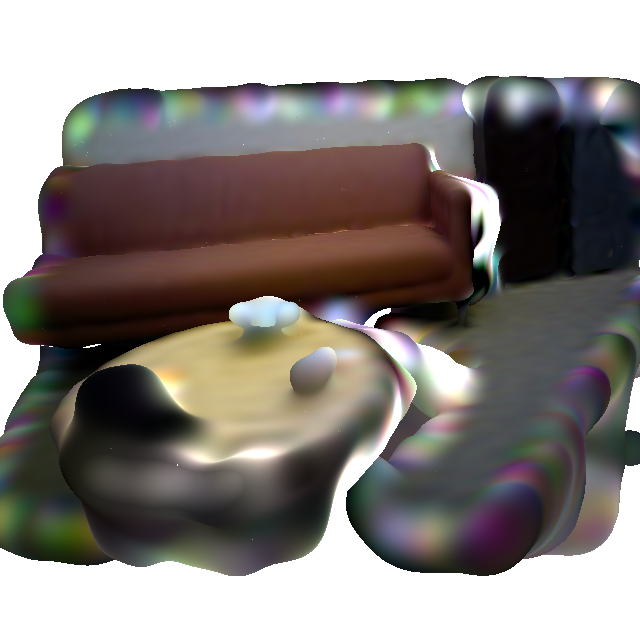} 
        \caption{\textbf{Root Implicit Layer:} Trained on a single RGBD image and rendered from novel view point in approximately 2.5min.} \label{fig1:rgbd}
    \end{subfigure}
    \hfill
    \begin{subfigure}[t]{0.32\textwidth}
        \centering
        \includegraphics[width=0.95\linewidth]{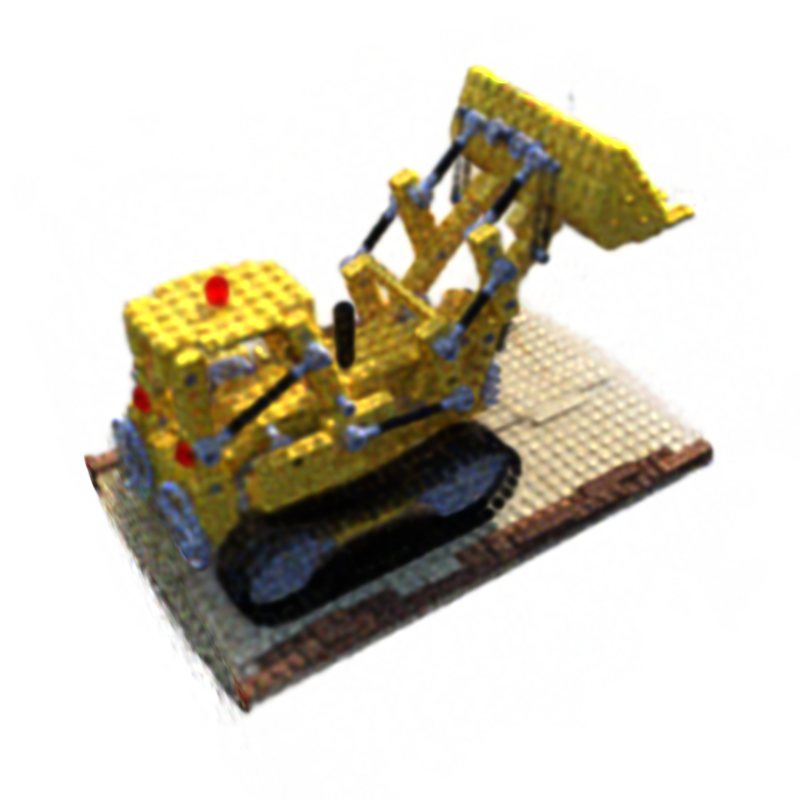} 
        \caption{\textbf{Integral Implicit Layer:} Trained on multiple RGB images annotated with view points in approximately 20min.} \label{fig1:radiance}
    \end{subfigure}
    \caption{An overview of the computational layers introduced in this paper. Wall time based on a single NVIDIA GeForce RTX 2080 Ti and a suboptimal implementation. The current implementation is only approximately 2.5\% efficient depending on layer. Assuming a FLOP efficient implementation, the training time could still be reduced dramatically, i.e. to 28s for the Integral Implicit Layer and to 30s for the Explicit Layer.}
    \label{fig:intro_fig}
\end{figure}

The Fast Multipole Method~\citep{greengard1987fast} is an algorithm that allows for the approximation of $y_m = \sum_{n=1}^N \phi(p_n,q_m) w_n$ with $1 \leq m \leq M$ in $\mathcal{O}(N+M)$ instead of $\mathcal{O}(NM)$ for some kernel functions $\phi$ under the assumption that $p_n$ and $q_m$ are low-dimensional. The algorithm is a two-step process. In its simplest form, the first step expands $p$ and $w$ into a representation that locally expresses $f(q_m) = \sum_{n=1}^N \phi(p_n,q_m) w_n$ by a series expansion. Because $f$ is being expressed locally (i.e. on a grid) this step requires $p$ and $q$ to be low dimensional. Popular series expansions that have previously been used are e.g. the Multipole (hence the name), Chebyshev or Taylor expansion \citep{messner2012optimized}. This first step has a computational cost of $\mathcal{O}(C_{expand}N)$ where $C_{expand}$ is a controllable constant coefficient that increases with the required accuracy. Traditionally, the FMM is employed in high-precision particle simulations that is why only far-field interactions, i.e. interactions for which $d(p_n,q_m) > \epsilon$ hold for some metric $d$, are approximated based on a series expansion. Near field interactions are typically resolved exactly. Let $F$ be the expansion of far-field interactions evaluated at spatial location $q_m$, then mathematically speaking:
\begin{align*}
    y_m = \sum_{i \in \{n | d(q_m,p_n) < \epsilon\}} \phi(q_m, p_i) w_i + F(q_m)
\end{align*}
Thus, in the traditional formulation of the FMM, after the price for expanding the far-field interactions has been paid, evaluating $q_m$ is in $\mathcal{O}((C_{eval} + C_{near})M)$ where $C_{eval}$ denotes the controllable cost of evaluating the series expansion representing far-field interactions and $C_{near}$ denotes the cost of exactly resolving near-field interactions. In this paper, we aggressively trade accuracy for advantageous computational properties and speed. That is why we approximate all interactions by a series expansion. By doing so, we not only reduce the constant coefficient for evaluation to $C_{eval}$, we can also design algorithms that act directly on the series expansion $F$ and exploit its mathematical properties.

\subsection{(Implicit) Gradient Based Learning}
In recent years, Gradient Based Learning has emerged as a dominant approach in Machine Learning and optimization. The general idea is simple: Given a function $f: X \times \Theta \rightarrow Y$ parameterized by $\theta \in \Theta$, a loss function $l: Y \times Y \rightarrow \mathbb{R}$ and a set of pairs of inputs and desired outputs $\{(x_i,y_i) | x_i \in X, y_i \in Y\}$ usually referred to as the training set, the empirical risk $\mathcal{L} = \sum_i l(f(x_i,\theta), y_i)$ is minimized by following $\theta$ in the opposite direction of the gradient of $\mathcal{L}$ w.r.t. $\theta$, i.e.
\begin{align*}
    \theta^{(n+1)} = \theta^{(n)} - \fracpartial{\mathcal{L}}{\theta}.
\end{align*}
One of the most prominent and currently successful incarnations of Gradient Based Learning are Neural Networks or Deep Learning~\citep{lecun2015deep}. For Neural Networks, $f$ is assumed to be a composition of simple computational operations, i.e. $f = f_N \circ f_{N-1} \circ ... f_1$ with $f_i$ usually referred to as a layer. Even though in practice some layers are not parameterized, assume for simplicity that layer $i$ is parameterized by $\theta_i$. Because $f$ is assumed to be a composition of functions, computing partial derivatives $\fracpartial{\mathcal{L}}{\theta_i}$ requires the chain-rule of derivatives. Let $y_i$ denote the output of layer $i$, then:
\begin{align*}
    \fracpartial{f}{\theta_i} &= \fracpartial{f_N}{y_{N-1}} \fracpartial{f_{N-1}}{y_{N-2}} ... \fracpartial{f_{i+1}}{y_{i}}\fracpartial{f_i}{\theta_{i}}\\
    &= \overline{y}_i \fracpartial{f_i}{\theta_i}
\end{align*}
$\overline{y}_i \fracpartial{f_i}{\theta_i}$ is usually referred to as the Jacobian Vector Product or JVP for short and the backpropagation algorithm~\citep{rumelhart1986learning} is a dynamic programming approach to speed up the computation of $\overline{y_i}$ by exploiting the recursion property: $\overline{y}_{i-1} = \overline{y}_{i} \displaystyle \fracpartial{f_{i}}{y_{i-1}}$. Thus, in order for a computational layer $f$ to be viable in the context of Gradient Based Learning, computationally efficient strategies to compute or approximate the following quantities need to be provided:
\begin{align*}
\centering
    \overline{y}_{i-1} = \overline{y}_{i} \fracpartial{f_{i}}{y_{i-1}}; \ \ 
    \overline{\theta}_i = \overline{y}_{i} \fracpartial{f_{i}}{\theta_{i}}
\end{align*}
Note that even for simple computational layers such as a convolutional layer~\citep{lecun1995convolutional}, writing out the Jacobian would be computationally too expensive but directly computing the JVP is nevertheless possible in a computationally efficient way. For the techniques introduced in this paper, even direct computations of the JVP are intractable. In order to overcome this, we will introduce computationally efficient strategies to closely approximate the JVP.\\

\textbf{Implicit Layers:} Recently, computational layers for Gradient Based Learning have been proposed whose input-output relationships are defined implicitly. These approaches usually define an auxiliary function $g: X \times Z \times \Theta \rightarrow \mathbb{R}$ imbued with an auxiliary input $z$. The process of computing the output of these layers usually eliminates the auxiliary variable $z$. We make the distinction between root-implicit~\citep{agrawal2019differentiable,amos2017optnet} and integral-implicit layers~\citep{chen2018neural,kidger2020neural}. Root-implicit layers usually determine and output quantities related to a root of $g$ w.r.t. $z$ in the forward pass and require an additional projection step in the backward whereas integral-implicit layers solve an integral during the forward and backward pass.

Mathematically speaking,
\begin{align*}
    \text{Root implicit: \ \ }&y_i = z\\
    \text{s.t.: \ \ } &g(x_i, z, \theta_i) = 0\\
    \text{Integral implicit: \ \ }&y_i = \int_a^b g(x_i, z, \theta_i) dz.
\end{align*}

In this paper, we will propose multiple approximate computational layers for Gradient Based Learning that internally make use of a variant of the Fast Multipole Method taylored to Machine Learning in order to approximate their outputs and JVPs. In particular, we introduce an explicit, a root-implicit and an integral-implicit layer. We showcase potential applications of these layers in the realm of Computer Vision and Graphics. Figure \ref{fig:intro_fig} summarizes the introduced computational layers and some of their potential applications.

\section{Related Work}

The application of fast (summation) algorithms in the context of Machine Learning is not new. Variants of the FMM have been applied to e.g. the problem of kernel density estimation and statistical learning more broadly~\citep{gray2001n,ram2009linear,lee2006dual}. The authors of~\citep{yang2004efficient} propose a Fast Gauss Transform that makes use of the ideas of Multipole expansions. The approaches introduced in this paper are also, to some degree, related to Geometric Deep Learning~\citep{bronstein2017geometric,cao2020comprehensive}. Geometric Deep Learning aims to generalize Deep Learning to non-Euclidean geometries. Even though the techniques introduced in this paper assume Euclidean spaces, it is straight-forward to generalize the techniques to e.g. periodic boundary conditions. Because the approaches introduced in this paper reduce the computational cost by exploiting low-rank structure, it is related to approaches like~\citep{jaderberg2014speeding} that however speed up discrete convolutions. Another line of work that is related to our work is operator learning \citep{li2020fourier,li2020multipole,lange2021fourier}. These approaches aim to learn a solution operator for a given partial differential equation and internally make use of series or Multipole expansions. 

We believe the approaches described in ~\citep{carr2001reconstruction} to be the closest relative to our approaches. The authors propose using fast summation algorithms to interpolate point clouds. This paper extends these ideas to the general case of gradient based learning which allows us to use the techniques as layers in Neural Networks.

The applications of fast summation algorithms discussed in this paper are in the realm of Computer Vision and Graphics. They can be understood as instances of Differentiable Rendering (DR). For a recent review of DR the reader is referred to~\citep{kato2020differentiable}. Current approaches to DR mostly differ in how shape is being parameterized or represented. There seem to be four rivaling approaches based on either meshes, voxels, point clouds and implicit representations. Mesh-based approaches ~\citep{loper2014opendr} update triangle meshes and other scene dependent properties based on gradient information but struggle with non-differentiability resulting from discontinuities of the rendering equation and computational issues in general for which numerous heuristics and tricks have been proposed~\citep{rhodin2015versatile,liu2019soft,chen2018neural,kato2018neural,zhang2020path}. Voxel-based approaches circumvent the problem of non-differntiability by e.g. modeling occupancy or occupancy probabilities within 3D voxels. In order to compute the color value of a pixel, these approaches aggregate some quantity along a ray~\citep{tulsiani2017multi,henzler2019escaping,lombardi2019neural}. Note that these voxels are usually spatially discrete which in turn can lead to artifacts and potential discontinuities at voxel boundaries. Because of the large memory requirements, these approaches have limited resolution which is sometimes alleviated by employing warping fields. The approach described in \cite{lombardi2019neural} e.g. uses an autoencoder structure that produces a voxel grid containing RGB$\alpha$ values conditioned on input images that is trained by a differentiable raymarching algorithm.
Because of their abundant availability, many approaches that internally make use of point clouds have been proposed~\citep{roveri2018pointpronets,yifan2019differentiable,wiles2020synsin,lin2018learning,li2020end,insafutdinov2018unsupervised}. These approaches usually require inferring the influence of each point within the point cloud on each pixel. This step can either lead to sparse images when this influence is assumed to be a Dirac delta or to large computational costs and blurriness when the influence is assumed to be large. These approaches oftentimes do not allow for the inference of a coherent surface, i.e. object boundaries might not be able to be extracted. The approach described in ~\citep{insafutdinov2018unsupervised} can be understood as a hybrid between voxel and point cloud based techniques.
More recently, implicit representations of shape and other scene dependent quantities have been proposed. These approaches oftentimes make use of Neural Networks to represent these quantities~\citep{chen2019learning2,mescheder2019occupancy,mildenhall2020nerf}. Two instances of implicit 3D representations are discussed in more detail in the remainder of the paper.

The approach introduced in this paper is to some degree related to voxel, point cloud and implicit approaches. The relation to mesh-based approaches is less pronounced. As we will show shortly, the $FC^2T^2$ expansion returns quantities on a voxel grid. However, instead of assuming discrete space, our approach lives in continuous space. This entails that instead of storing a value for each voxel, parameters of a function are stored within each cell which in turn allows for continuity across voxel boundaries. Since our approach is based on continuous-in-space convolutions (similar to radial basis functions \citep{broomhead1988radial}) and the position of these kernels are points in 3D, there is a connection to point cloud based approaches. However, because each point in this `point cloud' is associated with a kernel and parameterizes a function (or operator), it is easy to extract surface information. Because of this property, the relation to implicit approaches becomes apparent. To some degree, the approaches introduced in this paper could be understood as a potentially faster drop-in alternative to Neural Networks for implicit scene representations. However, because our approach does not make use of Neural Networks internally, it does not share the disadvantageous inductive bias of `smoothness' that neural approaches seem to exhibit and also come at a much lower FLOP footprint, i.e. our approach is considerably less demanding in terms of computational cost. Furthermore, because every voxel contains parameters of a 3D polynomial, we can device algorithms that exploit properties of polynomials such as e.g. fast root finding and integration.

\section{Fast Continuous Convolutional Taylor Transform}
\label{sec:fc2t2}

\begin{figure}
    \begin{subfigure}[t]{0.26\textwidth}
        \centering
        \includegraphics[width=\linewidth]{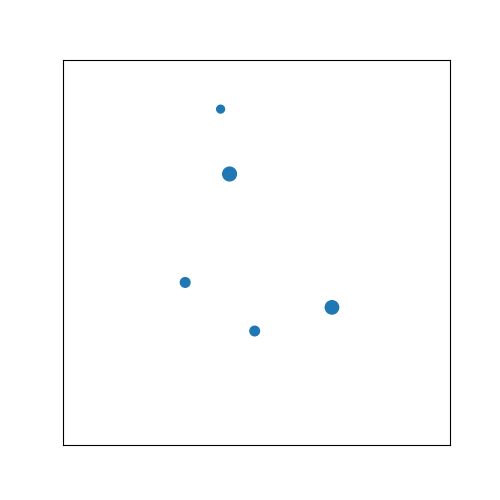} 
        \caption{\textbf{Input}: The inputs are continuous spatial locations and weights. Weights represented by dot size in the graphic.}
    \end{subfigure}
    \hfill
    \begin{subfigure}[t]{0.4\textwidth}
        \centering
        \includegraphics[width=\linewidth]{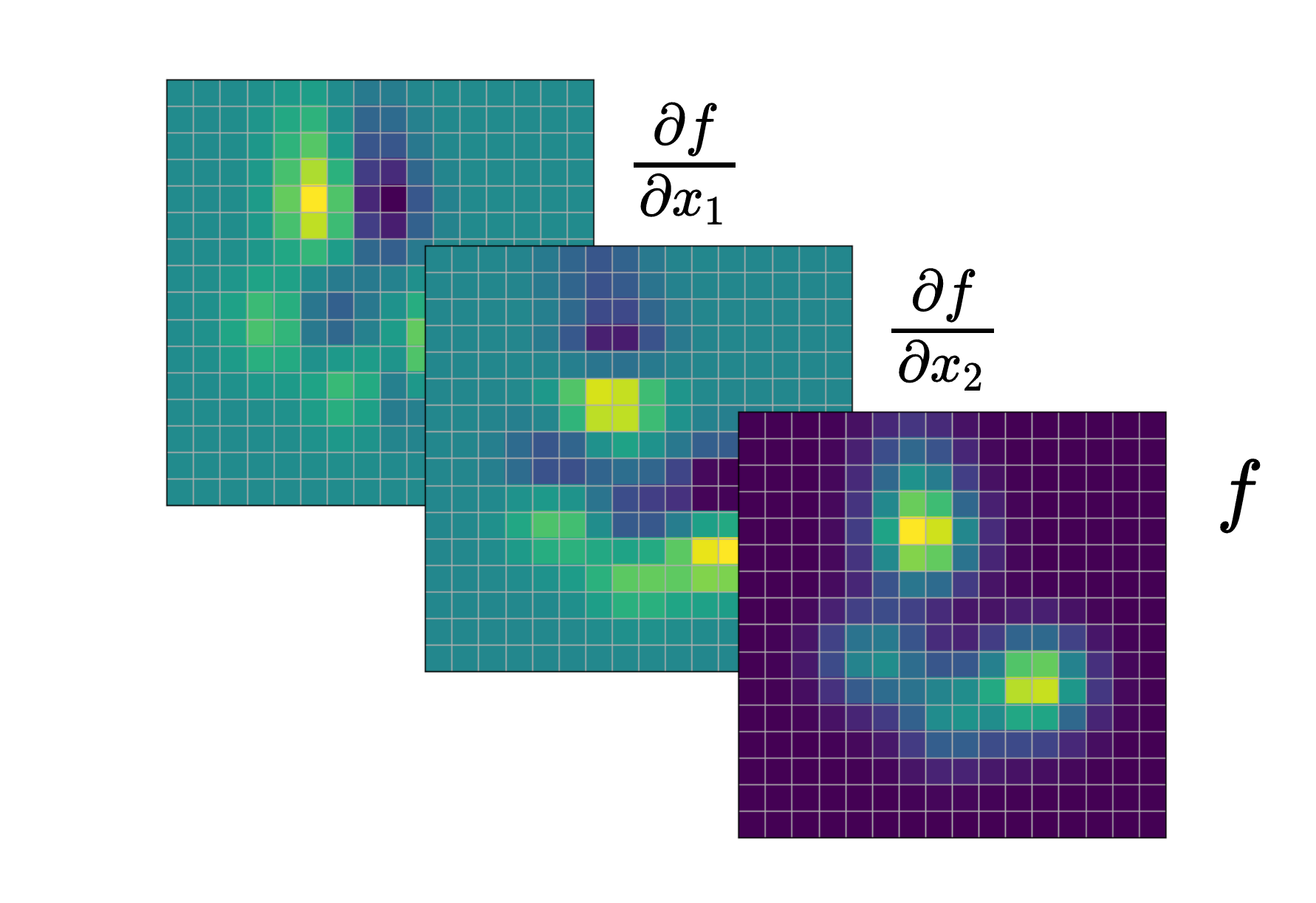} 
        \caption{\textbf{Expansion}: The $FC^2T^2$ discretizes space and computes a series expansion for every cell within the grid. Given $N$ input locations, this step is in $\mathcal{O}(N)$. The graphic illustrates a first order 2D expansion.}
    \end{subfigure}
    \hfill
    \begin{subfigure}[t]{0.26\textwidth}
        \centering
        \includegraphics[width=\linewidth]{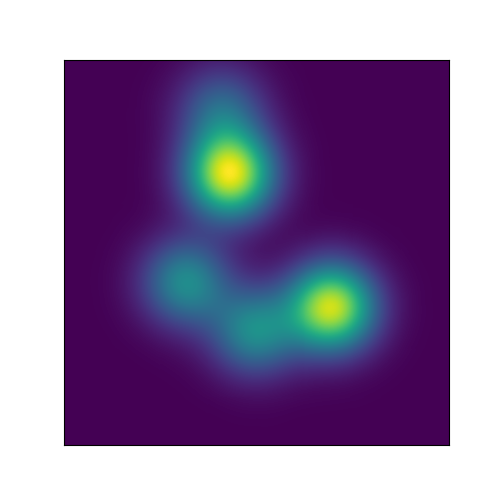}
        \caption{\textbf{Output:} Given the expansion, computing the output at $M$ target locations is independent of $N$, i.e. in $\mathcal{O}(M)$ resulting in $\mathcal{O}(N + M)$ for the entire procedure.}
    \end{subfigure}
    \caption{The $FC^2T^2$ computes the convolution of a kernel located at the source locations and weighted by the input weights in a two-step process. First, for every cell on a discrete grid a local representation based on series expansion is generated. After this initial setup, computing function values is fast and independent of the number of source locations.}
    \label{fig:input_expansion_output}
\end{figure}

The Fast Multipole Method~\citep{greengard1987fast} can be understood in different ways. One way of understanding it is in relation or comparison to the Fast Fourier Transform~\citep{beatson1997short} or from the perspective of Linear Algebra~\citep{yokota2015fast}. In this paper, we adopt a Machine Learning perspective. From a Machine Learning perspective, it can be understood as an efficient algorithm to approximate low dimensional convolutional operators in continuous space. The regular convolutional layer that is employed in many Neural Networks designed for applications in image processing usually assumes a discrete and fixed spatial grid~\citep{lecun1995convolutional}. Every grid cell is associated with a weight and computing the output of the convolutional layer requires convolving the weights distributed over this fixed grid with a kernel. The FMM performs a similar operation but in continuous space. Every weight is associated with a position in a low dimensional space and computing the output requires convolving the spatially distributed weights with a kernel. Whilst the regular convolutional layer assumes fixed locations of the weights and learns an optimal kernel, the techniques introduced in this paper assume a fixed kernel and learn optimal spatial locations and weights. Internally, similar to the regular convolutional layer, the FMM computes quantities on a grid. However, instead of storing the function value at every location of the grid, the FMM stores a set of series expansion coefficients that enable us to efficiently compute the value at any location within the grid cell. This process is sketched in Figure \ref{fig:input_expansion_output}. In this paper, we build upon \citep{challacombe1995recurrence} and store the coefficients of a 3D Taylor series expansion in every grid cell. As described earlier, the algorithms introduced in this paper not only exploit function values at specific locations but also directly act upon this intermediary representation. Specifically, we exploit the following mathematical properties of the 3D Taylor expansion:
\begin{itemize}
    \item \textbf{Line to polynomial}: Any line (or ray) through a 3D Taylor series expansion can be converted to a 1D polynomial efficiently.
    \item \textbf{Root finding}: Given a polynomial of order $\leq 4$, analytical closed-form solutions for its roots exist and are fast to evaluate ~\citep{ferrari}.
    \item \textbf{Integration and differentiation}: Integrating and differentiating polynomials is trivial and fast. The ability for quickly computing partial derivatives is particularly useful in the context of Gradient Based Learning.
    \item \textbf{Polynomial closure}: If $g(x)$ and $f(x)$ are polynomials, then so are $f(x) + g(x)$, $f(x)g(x)$ and $f(g(x))$. Adding, multiplying and composing polynomials is also reasonably fast if the degree of the polynomials is sufficiently small.
    \item \textbf{Polynomial to 3D Taylor}: Whilst the traditional FMM inserts points associated with a weight into the far-field expansion, the strategies introduced in this work allow us to insert functions of lines into the expansion.
\end{itemize}

We begin by explaining the general idea of the FMM following the structure of ~\citep{beatson1997short} with the Taylor series as the underlying expansion in mind. As described earlier, the FMM is an approximate technique to evaluate:
\begin{align}
    y_m = \sum_{n=1}^N \phi(p_n,q_m) w_n \label{eq:sum}
\end{align}
In the context of the FMM, $p$ and $q$ are oftentimes referred to as source and target locations respectively and a na\"ive approach to evaluating (\ref{eq:sum}) at $M$ target locations is obviously in $\mathcal{O}(NM)$. The FMM reduces this cost to $\mathcal{O}(N+M)$ at the expense of accuracy, i.e. by being approximate in nature.

For the sake of the argument, assume that $\phi$ is a degenerate kernel in the sense that it can be decomposed as $\phi(p,q) = \sum_{k=1}^\rho f_k(p)g_k(q)$. For such a kernel, evaluating $y_m$ can be sped up to $\mathcal{O}(\rho N + \rho M)$ in a trivial way by simple arithmetic, i.e. 
\begin{align*}
     y_m &= \sum_{n=1}^N \phi(p_n,q_m) w_n = \sum_{n=1}^N \sum_{k=1}^\rho f_k(p_n) g_k(q_m) w_n\\
     &= \sum_{k=1}^\rho (\sum_{n=1}^N f_k(p_n) w_n) g_k(q_m) = \sum_{k=1}^\rho A_k g_k(q_m)
\end{align*}
Because $A_k = \sum_{n=1}^N f_k(p_n) w_n$ is independent of any information of the target locations $q$, it can be computed once in $\mathcal{O}(\rho N)$. After computing and storing $A_k$, evaluating $y_m$ is in $\mathcal{O}(\rho M)$ therefore rendering the entire strategy to be in $\mathcal{O}(\rho N + \rho M)$. However, choosing $\phi$ to be degenerate significantly limits the expressiveness of $y$ and therefore the type of functions that could potentially be approximated by $y$.

The FMM algorithm exploits the same idea: Approximating the kernel $\phi$ by a truncated series expansion allows for the separation of the effects of target and source locations which ultimately will allow us to collect terms in the same manner as the degenerate kernel example above. The order of the series expansion then controls or trades off computation, memory and accuracy. Assume we approximate the kernel $\phi$ by a 3D Taylor series expansion, i.e. $p \in \mathbb{R}^3$. Furthermore, assume a kernel for which the following holds: $\phi(p,q) = \psi(p_1-q_1, p_2-q_2, p_3-q_3)$.

For the 3D Taylor series expansion centered at $\mathbf{c} = [c_1, c_2, c_3]$ truncated to order $\rho$ the following holds:
\begin{align*}
    &\text{Let } \partial^{n_1,n_2,n_3} f(x_1,x_2,x_3) = \frac{\partial^{n_1+n_2+n_3}f}{\partial x_1^{n_1} \partial x_2^{n_2} \partial x_3^{n_3}}(x_1,x_2,x_3).\\
    &f(x_1, x_2, x_3) \approx \sum_{n_1 + n_2 + n_3 \leq \rho} \partial^{n_1,n_2,n_3} f(c_1,c_2,c_3)\prod_{i=1}^3\frac{(x_i-c_i)^{n_i}}{n_i!}
\end{align*}
Assume that $p_i - p'_i$ and $q_i - q'_i$ is sufficiently small such that the Taylor expansion centered at $\mathbf{c}$ converges. Futhermore, let  $c_i = p_i' - q_i'$, $d_{p,i} = p_i - p_i'$ and $d_{q,i}$ analogously. Applying the 3D Taylor series expansion to $\phi$ then yields:
\begin{align}
    &\psi(p_1-q_1, p_2-q_2, p_3-q_3) \nonumber\\
     \approx& \sum_{n_1 + n_2 + n_3 \leq \rho} \partial^{n_1,n_2,n_3} \psi(c_1,c_2,c_3)\prod_{i=1}^3\frac{(p_i-q_i-c_i)^{n_i}}{n_i!} \nonumber\\
    =& \sum_{n_1 + n_2 + n_3 \leq \rho} \partial^{n_1,n_2,n_3} \phi(p', q')\prod_{i=1}^3\frac{(d_{p,i}-d_{q,i})^{n_i}}{n_i!} \nonumber
\intertext{Making use of the binomial theorem which reads $(x+y)^n = \sum_{k=0}^n \frac{n!}{(n-k)!k!} x^{n-k}y^k$ and some arithmetic gives us:}
    =& \sum_{n_1 + n_2 + n_3 \leq \rho} \partial^{n_1,n_2,n_3} \phi(p', q')\prod_{i=1}^3 \sum_{k=0}^{n_i} \frac{d_{q,i}^{k}d_{p,i}^{n_i-k}}{(n_i-k)!k!} \nonumber\\
    =& \sum_{n_1 + n_2 + n_3 \leq \rho} 
    \left(\sum_{k_1=0}^{n_1} \frac{d_{q,1}^{k_1}d_{p,1}^{n_1-k_1}}{(n_1-k_1)!k_1!}\right) 
    \left(\sum_{k_2=0}^{n_2} \frac{d_{q,2}^{k_2}d_{p,2}^{n_2-k_2}}{(n_2-k_2)!k_2!}\right) 
    \left(\sum_{k_3=0}^{n_3} \frac{d_{q,3}^{k_3}d_{p,3}^{n_3-k_3}}{(n_3-k_3)!k_3!}\right) \partial^{n_1,n_2,n_3} \phi(p', q') \nonumber \\
    =& \sum_{k_1 + k_2 + k_3 \leq \rho} 
    \left(\sum_{n_1=0}^{\rho - k_1} \frac{d_{q,1}^{k_1}d_{p,1}^{n_1}}{n_1!k_1!}\right) 
    \left(\sum_{n_2=0}^{\rho - k_2} \frac{d_{q,2}^{k_2}d_{p,2}^{n_2}}{n_2!k_2!}\right) 
    \left(\sum_{n_3=0}^{\rho - k_3} \frac{d_{q,3}^{k_3}d_{p,3}^{n_3}}{n_3!k_3!}\right) \partial^{n_1+k_1,n_2+k_2,n_3+k_3} \phi(p', q') \nonumber \\
    =& \sum_{k_1 + k_2 + k_3 \leq \rho} 
    \frac{d_{q,1}^{k_1}}{k_1!}\frac{d_{q,2}^{k_2}}{k_2!}\frac{d_{q,3}^{k_3}}{k_3!}\left(\sum_{n_1=0}^{\rho - k_1} \frac{d_{p,1}^{n_1}}{n_1!}\right) 
    \left(\sum_{n_2=0}^{\rho - k_2} \frac{d_{p,2}^{n_2}}{n_2!}\right) 
    \left(\sum_{n_3=0}^{\rho - k_3} \frac{d_{p,3}^{n_3}}{n_3!}\right) \partial^{n_1+k_1,n_2+k_2,n_3+k_3} \phi(p', q')\nonumber \\
    =& \sum_{k_1 + k_2 + k_3 \leq \rho} \underbrace{ \vphantom{\sum_{n_1,n_2,n_3=0}^{\rho - k_1,\rho - k_2,\rho - k_3}}
    \frac{d_{q,1}^{k_1}}{k_1!}\frac{d_{q,2}^{k_2}}{k_2!}\frac{d_{q,3}^{k_3}}{k_3!}}_{L(\mathbf{k}, \mathbf{d}_q)} \underbrace{
    \sum_{n_1,n_2,n_3=0}^{\rho - k_1,\rho - k_2,\rho - k_3} \partial^{n_1+k_1,n_2+k_2,n_3+k_3} \phi(p', q')}_{M2L(\mathbf{n},\mathbf{k}, p', q')} \underbrace{\frac{d_{p,1}^{n_1}}{n_1!}\frac{d_{p,2}^{n_2}}{n_2!}\frac{d_{p,3}^{n_3}}{n_3!} \vphantom{\sum_{n_1,n_2,n_3=0}^{\rho - k_1,\rho - k_2,\rho - k_3}}}_{M(\mathbf{n}, \mathbf{d}_p)} \label{eq:expansion}
\end{align}

Therefore:
\begin{align*}
    y_m \approx \sum_{\mathbf{k} \leq \rho} L(\mathbf{k}, \mathbf{d}_{q_m}) \sum_i^N \sum_{\mathbf{n}=0}^{\rho - \mathbf{k}} M2L(\mathbf{n},\mathbf{k},p',q') M(\mathbf{n}, \mathbf{d}_{p_i}) w_i
\end{align*}
Note that all terms in $M$ and $L$ are independent of target and source locations respectively. Thus we have achieved a separation similar to the example of the degenerate kernel by approximating the kernel $\phi$ with a truncated series expansion. The derivations above could already be turned into an approximate algorithm that resolves $y_m$. Assume a discretization of the domain into non-overlapping boxes and denote the centers by $p'$ and $q'$. Let $p'_i$ be the box that contains particle $p_i$ and $q'_i$ analogously. Furthermore, let $I(p')$ be the set of all indices of particles contained in box $p'$. Because the boxes are non-overlapping, $\sum_i^N f(p_i) = \sum_{p'} \sum_{i \in I(p')} f(p_i)$.
\begin{align*}
    y_m &\approx \sum_{\mathbf{k} \leq \rho} L(\mathbf{k}, \mathbf{d}_{q_m}) \sum_i^N \sum_{\mathbf{n}=0}^{\rho - \mathbf{k}} M2L(\mathbf{n},\mathbf{k},p'_i,q'_m) M(\mathbf{n}, \mathbf{d}_{p_i}) w_i\\
    &\approx \sum_{\mathbf{k} \leq \rho} L(\mathbf{k}, \mathbf{d}_{q_m}) \sum_{p'} \sum_{\mathbf{n}=0}^{\rho - \mathbf{k}} M2L(\mathbf{n},\mathbf{k},p'_i,q'_m) \sum_{i\in I(p')} M(\mathbf{n}, \mathbf{d}_{p_i}) w_i\\
    &\approx \sum_{\mathbf{k} \leq \rho} L(\mathbf{k}, \mathbf{d}_{q_m}) \sum_{p'} \sum_{\mathbf{n}=0}^{\rho - \mathbf{k}} M2L(\mathbf{n},\mathbf{k},p'_i,q'_m) \mathcal{M}(\mathbf{n}, p')\\
    \text{L2P:\ \ \ \ \ \ \ } &\approx \sum_{\mathbf{k} \leq \rho} L(\mathbf{k}, \mathbf{d}_{q_m}) \mathcal{L}(\mathbf{k}, p') \\
\end{align*}
With:
\begin{align*}
    \text{P2M:\ }\mathcal{M}(\mathbf{n}, p') &= \sum_{i\in I(p')} M(\mathbf{n}, \mathbf{d}_{p_i}) w_i\\
    \text{M2L:\ } \mathcal{L}(\mathbf{k}, q') &= \sum_{p'} \sum_{\mathbf{n}=0}^{\rho - \mathbf{k}} M2L(\mathbf{n},\mathbf{k},p',q') \mathcal{M}(\mathbf{n}, p')
\end{align*}

This algorithm collects the effects of all source points into their respective boxes $p'$. Traditionally, this step is called $P2M$. Then, in order to obtain a Taylor expansion at location $q'$, all cells $p'$ are convolved with the $M2L$ kernel. The pseudo-code below illustrates how such an algorithm:

\begin{minted}{python}
def slow_expand(source_locations, weights):
    for p,w in zip(source_locations,weights):
        g = find_box(p, grid_cells)
        d1,d2,d3 = p.x-g.x, p.y-g.y, p.z-g.z
        for n1 + n2 + n3 <= rho
            M[g,n1,n2,n3] += d1**n1/fac(n1) * \
                             d2**n2/fac(n2) * \
                             d3**n3/fac(n3) * w
    for g_q in grid_cells:
        for g_p in grid_cells:
            for k1+k2+k3 <= rho:
                for n1,n2,n3 <= rho-k1,k2,k3:
                    L[g_q,k1,k2,k3] = M2L(n,k,g_p,g_q)M[g_p,n1,n2,n3]
    return L
\end{minted}
In order to query data from the expansion, one would employ the following algorithm:
\begin{minted}{python}
def L2P(target_locations, L):
    ys = []
    for q in target_locations:
        g = find_box(q, grid_cells)
        d1,d2,d3 = q.x-g.x, q.y-g.y, q.z-g.z
        y = 0
        for k1+k2+k3 < rho:
            y += d1**k1/fac(k1) * \
                 d2**k2/fac(k2) * \
                 d3**k3/fac(k3) * L[g,k1,k2.k3]
        ys.append(y)
    return ys
\end{minted}

The proposed algorithm successfully separates the number of source and target locations in terms of computational complexity but is still too slow to be viable in practice. Let $G$ be the number of grid cells. It is easy to see that the computational complexity is $\mathcal{O}(G^2 + N + M)$. In practice, the number of grid cells is oftentimes on the order of the number of source locations, rendering the algorithm above quadratic in $N$ in practice and therefore too slow for most applications. In order to devise an algorithm that is truly in $\mathcal{O}(N + M)$ a crucial assumptions is missing, namely that the radius of convergence of $\phi$ grows exponentially with the distance from its center. This assumptions allows us to devise a multi-level variant of the algorithm above whose expansion step is significantly faster by resolving longer range interactions between boxes at a lower spatial resolution.

The assumption that the radius of converges increases exponentially with distance from the center of the kernel allows us to resolve the $M2L$ procedure for boxes that are further apart from each other at a lower spatial resolution. In practice, we collect the $\mathcal{M}$-expansion of adjacent boxes into a single larger box. Let $p''$ and $q''$ denote the centers of these larger boxes. The maximum size of the larger boxes depends of the radius of convergence of $\phi$ at the desired distance. Thus, we do not perform a Taylor expansion at $p' - q'$ but at a distance of $p'' - q''$ with $p'' = p' + d_p'$ and $q''$ analogously. This entails that $d_{p,i} + d_p' = p_i - p_i'' =: d_{p,i}''$ and $d_{q,i} + d_q' = q_i - q'' =: d_{q,i}''$. Plugging this into equation (\ref{eq:expansion}), yields:
\begin{align*}
    \sum_{k_1 + k_2 + k_3 \leq \rho} 
    \frac{{d_{q,1}''}^{k_1}}{k_1!}\frac{{d_{q,2}''}^{k_2}}{k_2!}\frac{{d_{q,3}''}^{k_3}}{k_3!} 
    \sum_{n_1,n_2,n_3=0}^{\rho - k_1,\rho - k_2,\rho - k_3} \partial^{n_1+k_1,n_2+k_2,n_3+k_3} \phi(p'', q'') \frac{{d_{p,1}''}^{n_1}}{n_1!}\frac{{d_{p,2}''}^{n_2}}{n_2!}\frac{{d_{p,3}''}^{n_3}}{n_3!}
\end{align*}

Investigating $M$-terms independently and applying the binomial theorem yields:
\begin{align*}
   M(\mathbf{n}, \mathbf{d}'_p) =  &\frac{{d_{p,1}''}^{n_1}}{n_1!}\frac{{d_{p,2}''}^{n_2}}{n_2!}\frac{{d_{p,3}''}^{n_3}}{n_3!} = 
    \frac{{(d_{p,1} + d_{p,1}')}^{n_1}}{n_1!}\frac{{(d_{p,2} +d_{p,2}')}^{n_2}}{n_2!}\frac{{(d_{p,3}+d_{p,3}')}^{n_3}}{n_3!}\\
    =& \sum_{k_1=0}^{n_1} d_{p,1}^{n_1} \frac{d_{p,1}'^{n_1-k_1}}{(n_1-k_1)!k_1!} 
       \sum_{k_2=0}^{n_2} d_{p,2}^{n_2} \frac{d_{p,2}'^{n_2-k_2}}{(n_2-k_2)!k_2!} 
       \sum_{k_3=0}^{n_3} d_{p,3}^{n_3} \frac{d_{p,3}'^{n_3-k_3}}{(n_3-k_3)!k_3!}\\
    =& \sum_{k_1,k_2,k_3=0}^{n_1,n_2,n_3} d_{p,1}^{n_1} d_{p,2}^{n_2} d_{p,3}^{n_3}
        \frac{d_{p,1}'^{n_1-k_1}}{(n_1-k_1)!k_1!}
        \frac{d_{p,2}'^{n_2-k_2}}{(n_2-k_2)!k_2!}
        \frac{d_{p,3}'^{n_3-k_3}}{(n_3-k_3)!k_3!}\\
    =& \sum_{k_1,k_2,k_3=0}^{n_1,n_2,n_3}
        \frac{d_{p,1}'^{n_1-k_1}}{(n_1-k_1)!}
        \frac{d_{p,2}'^{n_2-k_2}}{(n_2-k_2)!}
        \frac{d_{p,3}'^{n_3-k_3}}{(n_3-k_3)!} M(\mathbf{k}, \mathbf{d}_p)\\
    =& \sum_{\mathbf{k}}^\mathbf{n} M2M(\mathbf{d'}_p,\mathbf{n},\mathbf{k}) M(\mathbf{k}, \mathbf{d}_p)
\end{align*}
Let $I(p'')$ bet the set of all boxes $p'$ contained in the box $p''$. This implies that:
\begin{align*}
    \mathcal{M}(\mathbf{n}, p'') &= \sum_{p' \in I(p'')} \sum_{i\in I(p')} M(\mathbf{n}, \mathbf{d'}_{p_i}) w_i\\
    &= \sum_{p' \in I(p'')} \sum_{\mathbf{k}}^\mathbf{n} M2M(\mathbf{d'}_p,\mathbf{n},\mathbf{k}) \sum_{i\in I(p')} M(\mathbf{k}, \mathbf{d}_{p_i}) w_i\\
    &= \sum_{p' \in I(p'')} \sum_{\mathbf{k}}^\mathbf{n} M2M(\mathbf{d'}_p,\mathbf{n},\mathbf{k}) \mathcal{M}(\mathbf{k}, p')
\end{align*}
Once the $M2M$ procedure has been applied, a lower resolution $\mathcal{L}$ expansion can be obtained by applying the $M2L$ procedure. This $\mathcal{L}$ expansion is only valid if the distance between boxes supplied to the $M2L$ kernel is large enough, i.e. every spatial resolution is associated with a minimum distance at which interactions can be resolved. Although computationally inefficient, high spatial resolutions allow for resolving long range interactions but low spatial resolution expansions cannot accurately resolve short range interactions.

In order to avoid having to loop over $\mathcal{L}$ expansions at multiple spatial resolutions when performing $L2P$, we would like to sort the $\mathcal{L}$ expansions for large boxes into those of smaller boxes. This can be achieved by an operation analogous to $M2M$, called $L2L$:
\begin{align*}
    L(\mathbf{n}, \mathbf{d}_q) &= \sum_{k_1,k_2,k_3=n_1,n_2,n_3}^\rho
        \frac{d_{p,1}'^{k_1-n_1}}{(k_1-n_1)!}
        \frac{d_{p,2}'^{k_2-n_2}}{(k_2-n_2)!}
        \frac{d_{p,3}'^{k_3-n_3}}{(k_3-n_3)!} L(\mathbf{k}, \mathbf{d}'_q)\\
        &= \sum_{\mathbf{k}=\mathbf{n}}^\rho L2L(\mathbf{d'}_q, \mathbf{n},\mathbf{k}) L(\mathbf{k}, \mathbf{d}'_q)\\
    \mathcal{L}(\mathbf{n}, q') &= \sum_{\mathbf{k}=\mathbf{n}}^\rho L2L(\mathbf{d'}_q, \mathbf{n},\mathbf{k}) \mathcal{L}(\mathbf{k}, q'')
\end{align*}
\begin{figure}
    \begin{subfigure}[t]{0.48\textwidth}
        \centering
        \includegraphics[width=\linewidth]{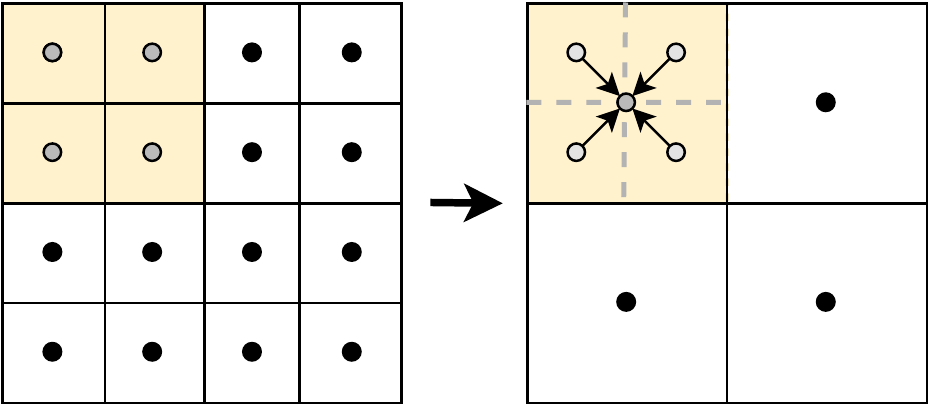} 
        \caption{\textbf{M2M:} The $M2M$ kernels allows to recursively obtain an $\mathcal{M}$ expansion at a lower spatial resolution. Lower resolution $\mathcal{M}$ expansions can be used to resolve longer range $M2L$ interactions more efficiently because there is less boxes.} \label{fig:m2m}
    \end{subfigure}
    \hfill
    \begin{subfigure}[t]{0.48\textwidth}
        \centering
        \includegraphics[width=\linewidth]{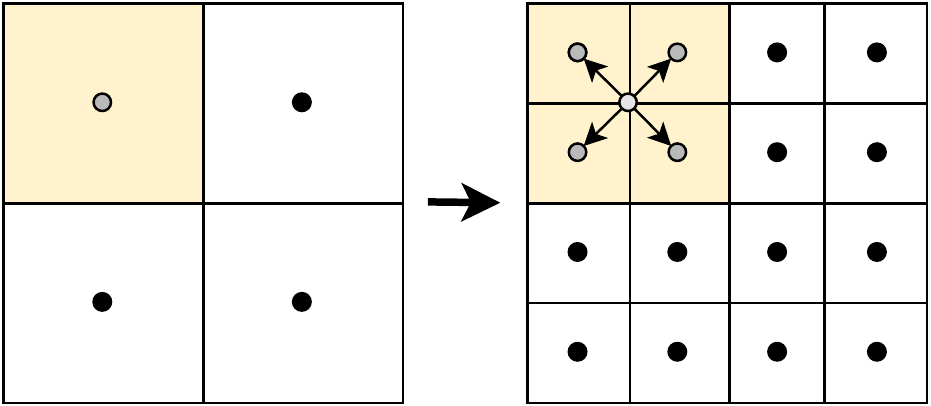} 
        \caption{\textbf{L2L:} The $L2L$ kernel can be used to hierarchically sort $\mathcal{L}$ expansions of a lower spatial resolution into a higher resolution $\mathcal{L}$ expansion to ultimately obtain a single $\mathcal{L}$ expansion that represents interactions of all ranges.} \label{fig:l2l}
    \end{subfigure}
    \caption{The $M2M$ and $L2L$ kernels allow for a multilevel variant that significantly speeds up the $M2L$ procedure by resolving longer range interactions at lower spatial resolutions.}
    \label{fig:x2x}
\end{figure}
In practice, the $M2M$ procedure is employed recursively to obtain $M$-expansions at varying spatial resolutions to resolve interactions at a specific range while the $L2L$ procedure is used to collect lower resolution $L$-expansions into higher resolution ones. For uniform grids and assuming that the radius of convergence doubles when doubling the distance from the kernel center, the $M2M$ and $L2L$ kernels can be implemented by a convolution and transpose-convolution with stride 2, respectively. The channel dimensions of the kernel depend on the order of the expansion. For example, the convolution kernel for the $M2M$ or $L2L$ operation for a 2D FMM with $\rho = 1$ has a shape of $3 \times 3 \times 2 \times 2$ and for $\rho=2$ has shape $6 \times 6 \times 2 \times 2$. The leading dimension signifies the order of the partial derivative, i.e. for $\rho = 1$ the channel dimensions denote $f$, $\partial^{1,0} f$ and $\partial^{0,1} f$. Increasing $\rho$ to 2, adds $\partial^{2,0} f$, $\partial^{0,2} f$ and $\partial^{1,1} f$. Let $\Rho$ be the number of leading dimensions dependent on $\rho$. Note that $\Rho$ is dependent on the number of spatial dimensions (2D vs 3D). Furthermore, because some arbitrary but consistent ordering of the partial derivatives needs to be assumed, let $i:\mathbb{N} \times \mathbb{N}\ (\times \mathbb{N}) \rightarrow \mathbb{N}$ be the indexer that induces this ordering and $I$ be the indicator function that is 1 if its argument is True and 0 otherwise, then for the convolutional kernel of $M2M$ and $L2L$ operation in 2D the following holds:

\begin{align*}
    K^{M2M}_{i[\mathbf{n}], i[\mathbf{k}]}(d) = \begin{bmatrix}
    \frac{(-d)^{k_1-n_1}}{(k_1-n_1)!}\frac{(-d)^{k_2-n_2}}{(k_2-n_2)!} & \frac{(-d)^{k_1-n_1}}{(k_1-n_1)!}\frac{(d)^{k_2-n_2}}{(k_2-n_2)!}\\
    \frac{(d)^{k_1-n_1}}{(k_1-n_1)!}\frac{(-d)^{k_2-n_2}}{(k_2-n_2)!} & \frac{(d)^{k_1-n_1}}{(k_1-n_1)!}\frac{(d)^{k_2-n_2}}{(k_2-n_2)!} 
    \end{bmatrix} I[k_1 \leq n_1, k_2\leq n_2]\\
    K^{L2L}_{i[\mathbf{n}], i[\mathbf{k}]}(d) = \begin{bmatrix}
    \frac{(-d)^{n_1-k_1}}{(n_2-k_2)!}\frac{(-d)^{n_1-k_1}}{(n_2-k_2)!} & \frac{(-d)^{n_1-k_1}}{(n_2-k_2)!}\frac{(d)^{n_1-k_1}}{(n_2-k_2)!}\\
    \frac{(d)^{n_1-k_1}}{(n_1-k_1)!}\frac{(-d)^{n_2-k_2}}{(n_2-k_2)!} & \frac{(d)^{n_1-k_1}}{(n_1-k_1)!}\frac{(d)^{n_2-k_2}}{(n_2-k_2)!}
    \end{bmatrix}  I[k_1 \geq n_1, k_2\geq n_2]
\end{align*}

Assuming a uniform grid, the $M2L$ operation can also be expressed as a convolution. Special care needs to be taken not to resolve one source point multiple times at different spatial resolutions for the same grid cell. The implementation employed in this paper, uses a convolution with shape $\Rho \times \Rho \times 6 \times 6\ (\times 6)$. In order not to resolve the same source point multiple times, the convolutional kernel has a hole in its middle to be filled by the $M2L$ operation at a higher spatial resolution. In order to fill this hole snugly, the kernel needs to change depending on whether the index of the grid cell is even or odd which can be implemented reasonably efficiently using strides and interleaving. Let $\odot$ be element-wise multiplication and $\partial \psi$ be applied element-wise, then the example below shows the convolutional kernel for a 2D FMM. Assuming zero-based numbering, the red box shows the kernel that is applied to a grid cell with an odd $x$ and $y$ index. Green, blue and orange boxes show the kernel for odd/even, even/odd and even/even cells respectively.
\begingroup
\renewcommand*{\arraystretch}{1.3}
\begin{align*}
     D_d &= \begin{bmatrix}
     \\[-1.2em]
    \ \tikzmark{left1}(-3d,-3d) & \tikzmark{left2}(-3d,-2d) & (-3d,-d) & (-3d,0) & (-3d,d) & (-3d,2d) & (-3d,3d)\ \ \\
    \ \tikzmark{left3}(-2d,-3d) &\tikzmark{left4}(-2d,-2d) & (-2d,-d) & (-2d,0) & (-2d,d) & (-2d,2d) & (-2d,3d) \\
    \ (-d,-3d) &(-d,-2d) & (-d,-d) & (-d,0) & (-d,d) & (-d,2d) & (-d,3d) \\
    \ (0,-3d) &(0,-2d) & (0,-d) & (0,0) & (0,d) & (0,2d) & (0,3d) \\
    \ (d,-3d) &(d,-2d) & (d,-d) & (d,0) & (d,d) & (d,2d) & (d,3d) \\
    \ (2d,-3d) &(2d,-2d) & (2d,-d) & (2d,0) & (2d,d) & (2d,2d)\tikzmark{right1} &  (2d,3d)\tikzmark{right2} \\
    \ (3d,-3d) &(3d,-2d) & (3d,-d) & (3d,0) & (3d,d) & (3d,2d)\tikzmark{right3} & (3d,3d)\tikzmark{right4} \\[0.3em]
    \end{bmatrix}\\
\DrawBoxA[thick]
\DrawBoxB[thick]
\DrawBoxC[thick]
\DrawBoxD[thick]
\end{align*}
\endgroup
\begin{align*}
    \overline{D} &= \begin{bmatrix}
    1 & 1 & 1 & 1 & 1 & 1 & 1\\
    1 & 1 & 1 & 1 & 1 & 1 & 1\\
    1 & 1 & 0 & 0 & 0 & 1 & 1\\
    1 & 1 & 0 & 0 & 0 & 1 & 1\\
    1 & 1 & 0 & 0 & 0 & 1 & 1\\
    1 & 1 & 1 & 1 & 1 & 1 & 1\\
    1 & 1 & 1 & 1 & 1 & 1 & 1\\
    \end{bmatrix} \hspace{50pt}
    K^{M2L}_{i[\mathbf{n}], i[\mathbf{k}]}(d) = \partial^{\mathbf{n} + \mathbf{k}} \psi (D_d) \odot \overline{D}
\end{align*}

\begin{figure}
    \centering
    \includegraphics[width=0.8\linewidth]{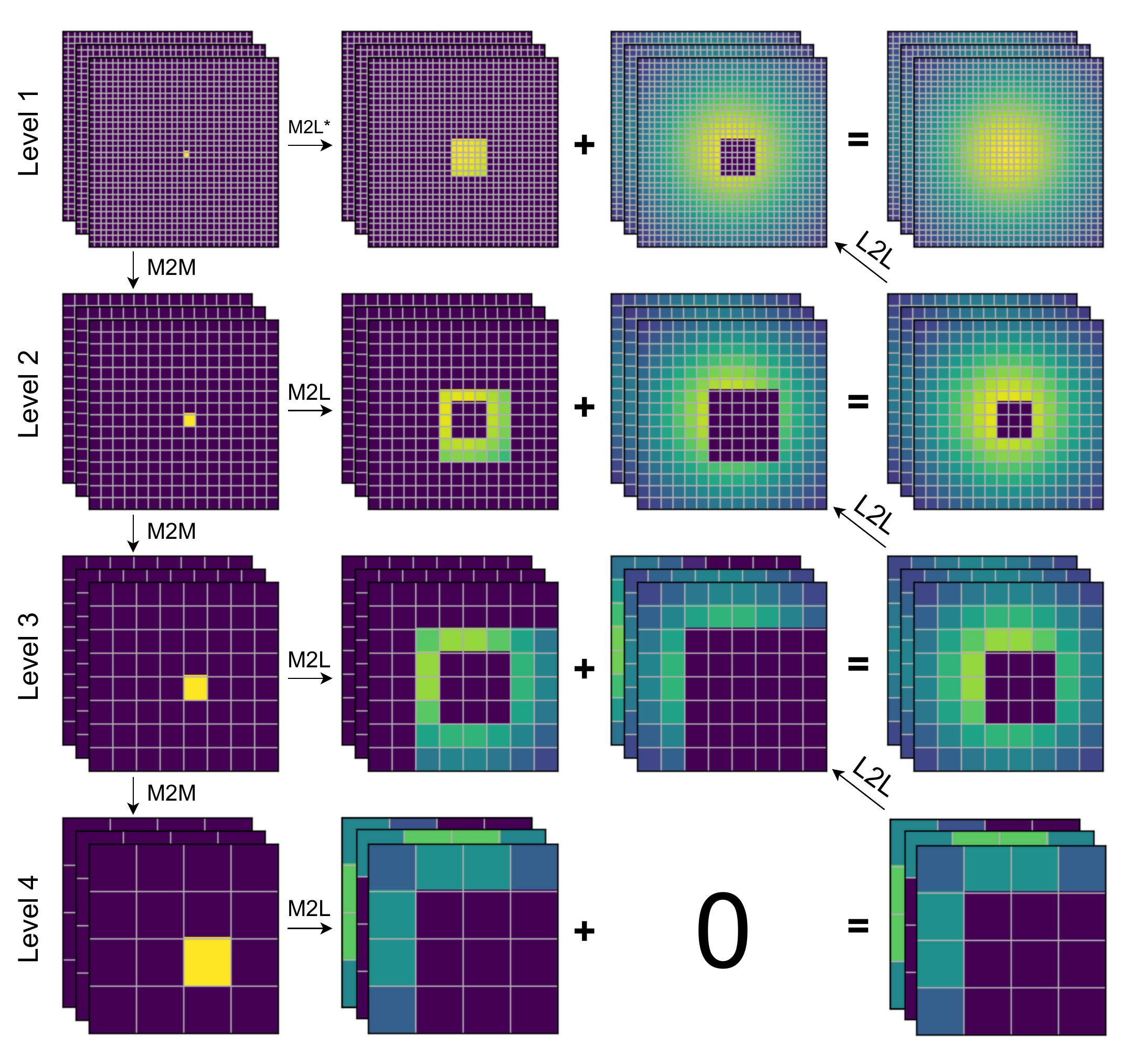}
    \caption{The data flow of a 4 level 2D FMM with a single source point located approximately in the center of the spatial domain. The three layers represent the function and its partial derivative in both spatial directions, i.e. $\rho=1$ in this example. At Level 1 we depart from the original formulation of the FMM and approximate near field interaction by a series expansion as well.}
    \label{fig:fmm_dataflow}
\end{figure}

The process of sorting source points into the largest $M$-expansion is traditionally referred to as $P2M$ whereas evaluating the function value at a specific location is called $L2P$. The overall flow of data and operation is as follows and depicted graphically in Figure \ref{fig:fmm_dataflow}. The initial step sorts source locations into their respective boxes ($P2M$) at the highest grid granularity. The computational graph then branches off: the $M2L$ procedure can be applied to compute the $L$-expansion at the highest level but at the same time, the $M$-expansion at the next lower spatial resolution can be computed. This branching allows for a large degree of parallelization. Computing the highest resolution $L$-expansion is the computationally most demanding step and can be done at the same time as the $M2M$, $M2L$ and $L2L$ procedures at lower spatial resolutions.

\subsection{Least Squares}
\begin{figure}
    \centering
    \includegraphics[width=\linewidth]{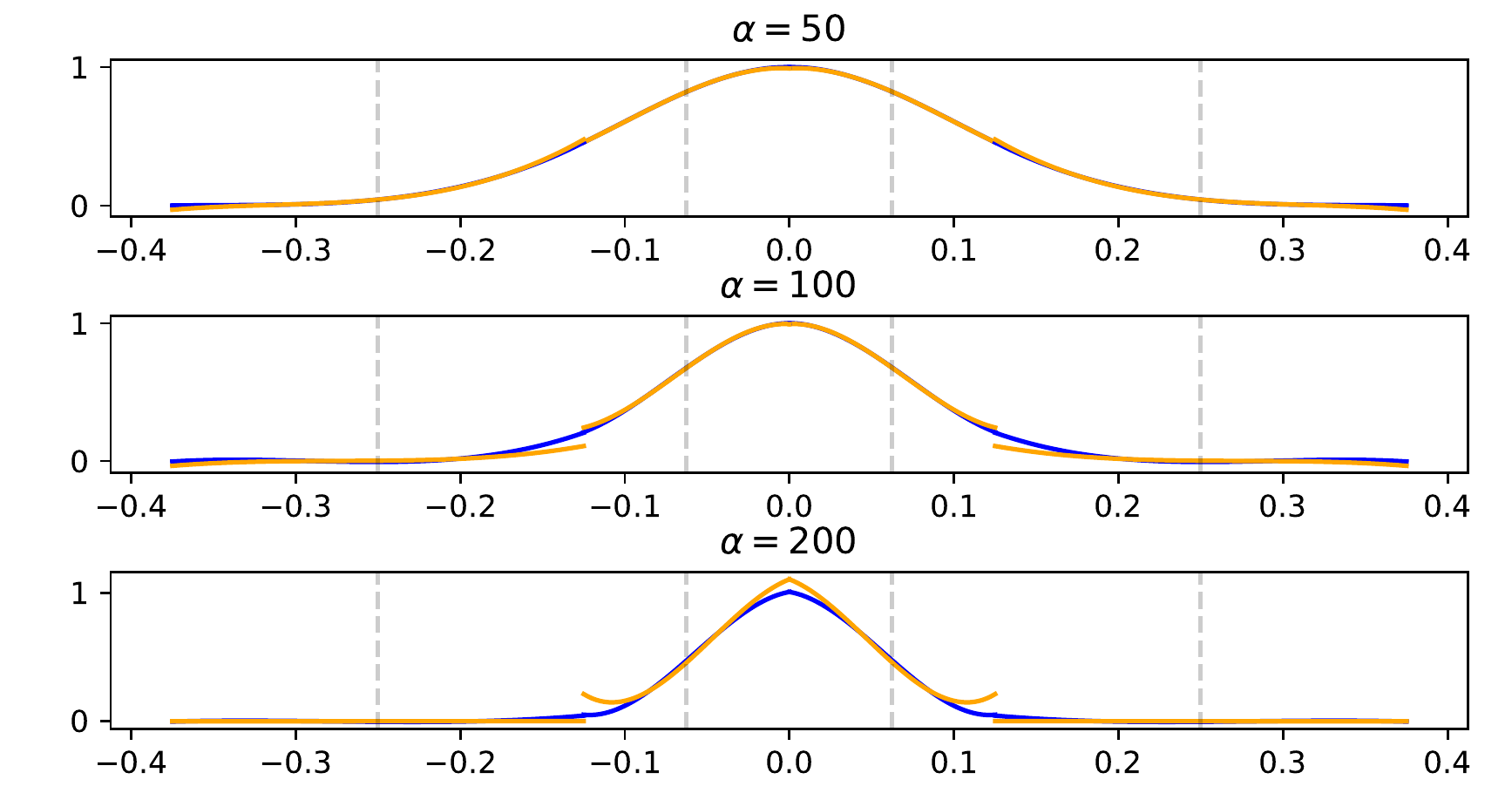}
    \caption{The effect of fitting polynomials to the kernel and its partial derivatives instead of using partial derivatives directly. The order of the expansion in this example is $\rho = 4$ and the kernel is $\phi(x) = \exp(-\alpha x^2)$ for various $\alpha$. Orange lines show the approximation with regular Taylor series whilst blue lines show the approximation by first performing a least-squares polynomial fit. Grey dashed lines denote locations of expansions.}
    \label{fig:least_squares}
\end{figure}
Unlike Fourier series that expand a function into an orthogonal basis in the data limit, Taylor series are not imbued with this notion of orthogonality. This can lead to odd and undesirable behavior. Consider $f(x) = \exp(-a x)$ with $a \gg 1$. The $n$th derivative of $f$ is $\nabla^n f(x) = (-a)^n \exp(-a x)$. If we were to perform a Taylor series expansion of $f$, the magnitude of derivatives increases exponentially while oscillating around the $x$-axis. Colloquially speaking, the odd-ordered derivatives overshoot towards $-\infty$ whilst the even-ordered derivatives overshoot toward $\infty$. Furthermore, we would like to inform the series expansion of the desired radius of convergence and order of the expansion, i.e. we would like to expand $f$ in a polynomial basis that is optimal in the least-squares sense for a given radius of convergence and expansion order. In order to achieve this, we employ a simple idea that is similar to the strategies introduced in ~\citep{fong2009black} even though motivated differently. In ~\citep{fong2009black} the authors propose fitting Chebyshev polynomials to the kernel $\phi$ in order to generalize the FMM to kernels that are only known numerically and not symbolically. In this paper, in order to imbue the $FC^2T^2$ expansion with notions of optimality in the least-squares sense for a given desired radius of convergence and order of expansion, we fit regular polynomials to the kernel but also all of its partial derivatives, i.e. we require the assumption that the kernel is provided symbolically. This trick only affects the $M2L$ kernel and significantly improves memory and computational requirements because the number of grid cells and $\rho$ can be kept smaller in comparison to the ordinary Taylor expansion for a similar degree of accuracy. Note that fitting polynomials to the kernel and its partial derivatives only needs to be performed once in a pre-processing step. The effects of trick are demonstrated qualitatively in Figure \ref{fig:least_squares}.

\subsection{How to use the implementation?}
Currently, the techniques are implemented in JAX~\citep{jax2018github} and python \citep{python}. The $FC^2T^2$ algorithm requires three inputs. First, the kernel $\psi$, the number of levels that controls the grid granularity and the order of the expansion. Because the kernel and its partial derivatives are approximated by a polynomial fit, it needs to be provided symbolically. We employ the python package sympy~\citep{sympy} for this. Given these three inputs, a functions that returns the expansion given source locations and weights and an accessor object that allows to query function values and partial derivatives are returned. The code below shows an example of a 4-level fourth-order $FC^2T^2$ with a Gaussian kernel.
\begin{minted}{python}
from fc2t2.base import initialize

func = lambda pkg: lambda x,y,z: pkg.exp(-5*(x**2 + y**2 + z**2))
expand, A = initialize(func, levels=4, rho=4)

#B is batch dimension, C is channel dimension
p = random_matrix((B,N,3))
w = random_matrix((B,C,N))

L = expand(p,w)
#N2 = 2**(levels+1), P is function of rho, in case of rho=4, P = 35
#L.shape = (B,C,N2,N2,N2,P)
\end{minted}

The accessor object implements a similar \verb+__getitem__+ API as a 5D numpy array~\citep{harris2020array} with shape $(B,C,N,N,N)$. The last three dimensions denote spatial dimensions and are handled differently whilst the batch and channel dimensions behave exactly like those of a numpy array. In contrast to a numpy array, for the spatial dimensions, the accessor object allows for querying data at continuous locations but requires an input for every dimension. The code below shows some examples of how to query data from the expansion in comparison to a numpy array.

\begin{minted}{python}
N = np.zeros((B,C,N,N,N)) #a numpy array
S = A(L) #the accessor object

y = N[0,0,0,0,0] #y is scalar
y = S[0,0, 0.,0.,0.] #y is scalar

y = N[:,:,0,0,0] #y.shape = (B,C)
y = S[:,:,0.,0.,0.] #y.shape = (B,C)
\end{minted}
The spatial dimensions accept any combination of a float scalar, a one-dimensional float vector or a slice. The inputs to the spatial dimensions need to be broadcastable onto each other. For slices, instead of behaving like \verb+arange+, slices behave like \verb+linspace+. We assume the domain to be $(-1,+1)$ for all spatial dimensions. All source and target locations are required to be within the domain.
\begin{minted}{python}
y = S[0,0,a:b:c,a:b:c,a:b:c] #y.shape = (c,)
#equivalent to:
z = linspace(a,b,c)
y = S[0,0,z,z,z] #y.shape = (c,)

y = S[0,0,::128,::128,::128] #y.shape = (128,)
#shorthand for
y = S[0,0,-1:1:128,-1:1:128,-1:1:128]
\end{minted}
Additionally, the API provides shortcuts to query data in a volume. When querying data in a volume, the meshgrid of the individual inputs to the spatial dimension is formed. Thus, inputs do not necessarily need to be broadcastable onto each other but large amounts of data might be queried by accident.
\begin{minted}{python}
y = S.vol[...,-1:1:256,-1:1:512,0.] #y.shape = (B,C,256,512,1)
\end{minted}
The API also allows to extract gradients and all partial derivatives of order 2. They could also be extracted in a volume if necessary.
\begin{minted}{python}
y = S.partials[..., 0., 0., 0.] #y.shape = (B,C,3)
y = S.partials2[..., 0., 0., 0.] #y.shape = (B,C,6)
y = S.partials.vol[0,0,::128,::64,0.] #y.shape = (128, 64, 1, 3)
\end{minted}

\subsection{Benchmarks}
\begin{figure}
    \centering
    \begin{subfigure}[t]{0.86\textwidth}
        \centering
        \includegraphics[width=\linewidth]{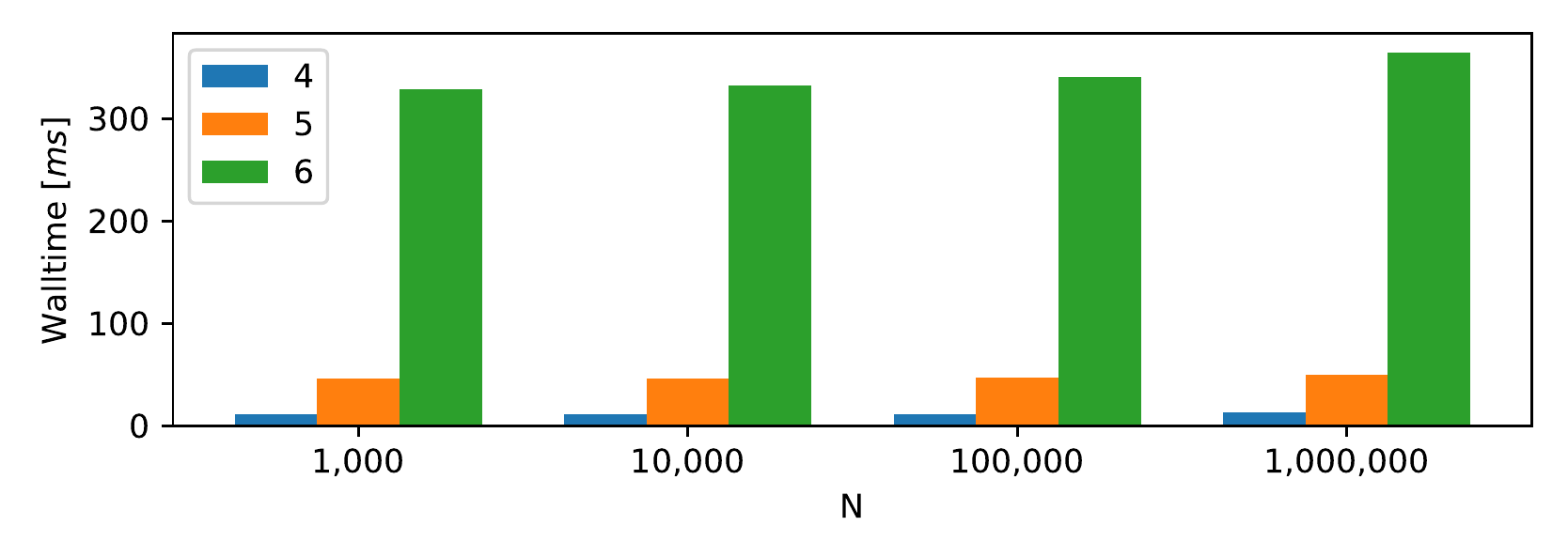} 
        \caption{\textbf{Expansion}: Wall time in milliseconds for the expansion step of the proposed algorithm. The number of levels that controls the granularity of the grid is color coded. Note that the granularity of the grid as opposed to the number of source locations determines the majority of the run-time of the expansion step.}
    \end{subfigure}
    \begin{subfigure}[t]{0.86\textwidth}
        \centering
        \includegraphics[width=\linewidth]{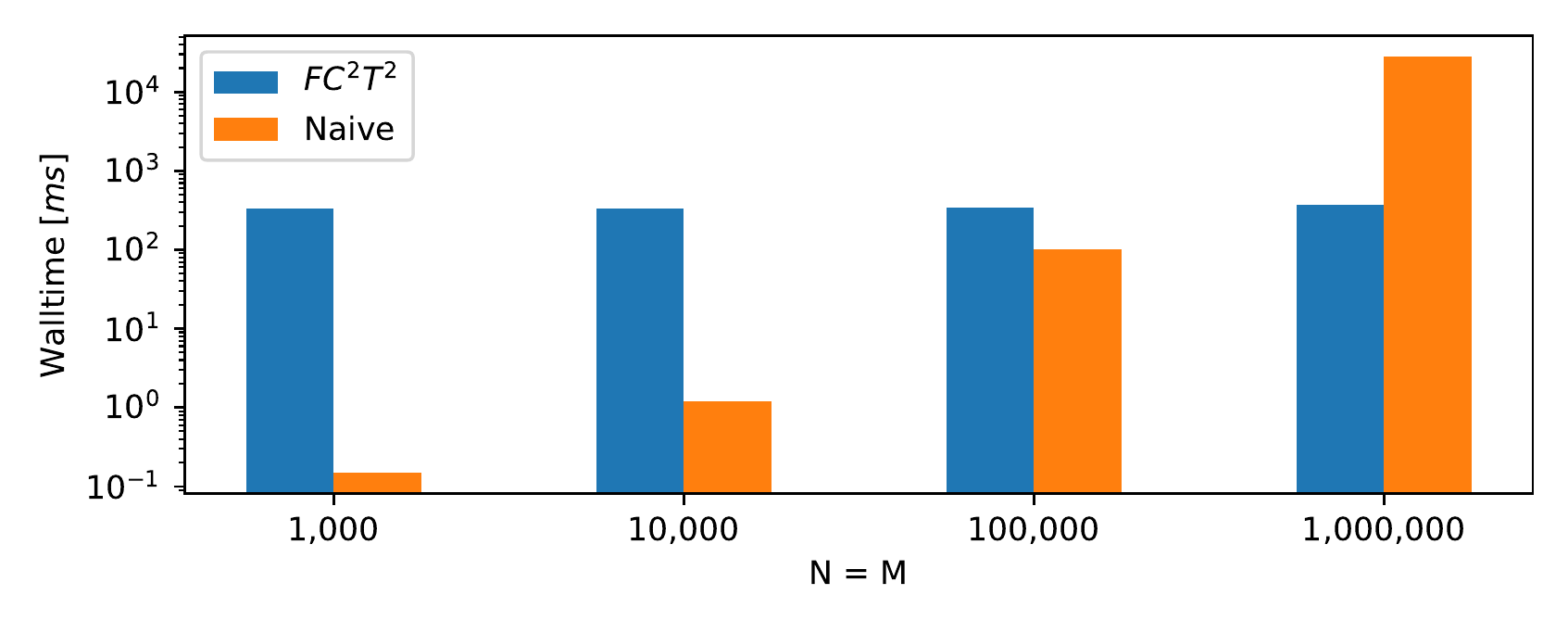}
        \caption{\textbf{Expand + Extract:} A comparison between the na\"ive implementation and the proposed algorithm with grid granularity level 6 in log-scale assuming $N=M$. Even for a modest amount of parameters and evaluations ($N=M=1m$) in the context of ML, $FC^2T^2$ is already approximately 75x faster.}
    \end{subfigure}
    \caption{A comparison of the wall time required for the expansion and extraction step for different levels of the expansion and number of source and target locations. All measurements are based on \emph{jit}ted JAX implementations executed on a NVIDIA RTX2080 Ti.}
    \label{fig:benchmark_expansion}
\end{figure}

The $FC^2T^2$ expansion in its current inception was designed based on the principles of robustness, ease of use/implementation and generality. It is robust because its speed is independent of the distribution of source points and general because it allows for any symmetric kernel ($\phi(q,p) = \phi(p,q))$ whose radius of convergence increases exponentially with distance from center. If a specific kernel would be assumed further speed-ups could be achieved by e.g. decomposing the $M2L$ kernel for each spatial dimension for kernels that allow this like e.g. the Gaussian kernel or by using an adaptive instead of a fixed grid. We believe that the following performance benchmarks can be improved significantly by future work and should serve as a competitive lower bound. The techniques scale linearly with batch and channel dimension and results are shown for $B = C = 1$ and a Gaussian kernel $\phi(x,y,z) = \exp(-\alpha (x^2 + y^2 + z^2))$ for $\alpha$ depending on the number of levels because expansions on a finer grid (more levels) allow for smaller or tighter Gaussian kernels. We keep $\rho$ fixed at four. Figure \ref{fig:benchmark_expansion} shows the effects of changing the granularity of the grid by varying the levels of the expansion and varying the number of source and target locations. For the applications in Computer Graphics and Vision introduced later, we oftentimes use 8m source locations evaluated at roughly 10m target locations. In that case, the $FC^2T^2$ expansion is 10,000x faster compared to the na\"ive implementation. Note that the memory requirements are fairly modest. A 4-level, 5-level and 6-level expansion require storing $32 \times 32 \times 32 \times 35$ (4.5MB), $64 \times 64 \times 64 \times 35$ (36.7MB) and $128 \times 128 \times 128 \times 35$ (293.6MB) values per channel respectively.

When the performance of the algorithms is dissected, we find that the algorithm scales gracefully with $N$ and $M$. Only the $P2M$ and $L2P$ steps of the algorithm are dependent on $N$ or $M$ respectively. Locating the box of a single source or target particle can be done by computing $int((p_i + 1) N/2)$ for each spatial dimension, therefore requiring 3 FLOPs per spatial dimension assuming that casting to an integer requires one FLOP and $N/2$ was precomputed. Then distances to the center of the box need to be computed which requires 1 FLOP per spatial dimension and $\rho$-many powers weighted by factorials are computed resulting in 8 FLOPs per spatial dimension. For each partial derivative of a certain order, two FLOPs are required to compute the product of distances and in case of $L2P$ another FLOP to multiply with the respective coefficient and $\Rho - 1$ to sum weighted coefficients up. A 3D expansion with $\rho = 4$ implies that $\Rho = 35$. This entails that for $P2M$ and $L2P$ a total of $9 +3 + 24 + 35*2 = 106$FLOPs and $9 + 3 + 24 + 35*4 = 176$FLOPs per source and target location are needed respectively. \textbf{The majority of the FLOPs is spent on work that is independent of $N$ and $M$}. Per grid cell, the $M2L$ and $L2L$ operations require $35^2 * 2^3 * 2$FLOPs  whereas $M2L$ requires $35^2 * 6^3 * 2$. This implies that $M2L$ is the most expensive operation of the algorithm by a large margin when making the not unreasonable assumption that the number of grid cells is roughly on the order of $M$ and $N$.
\begin{figure}
    \centering
    \includegraphics[width = 0.65\linewidth]{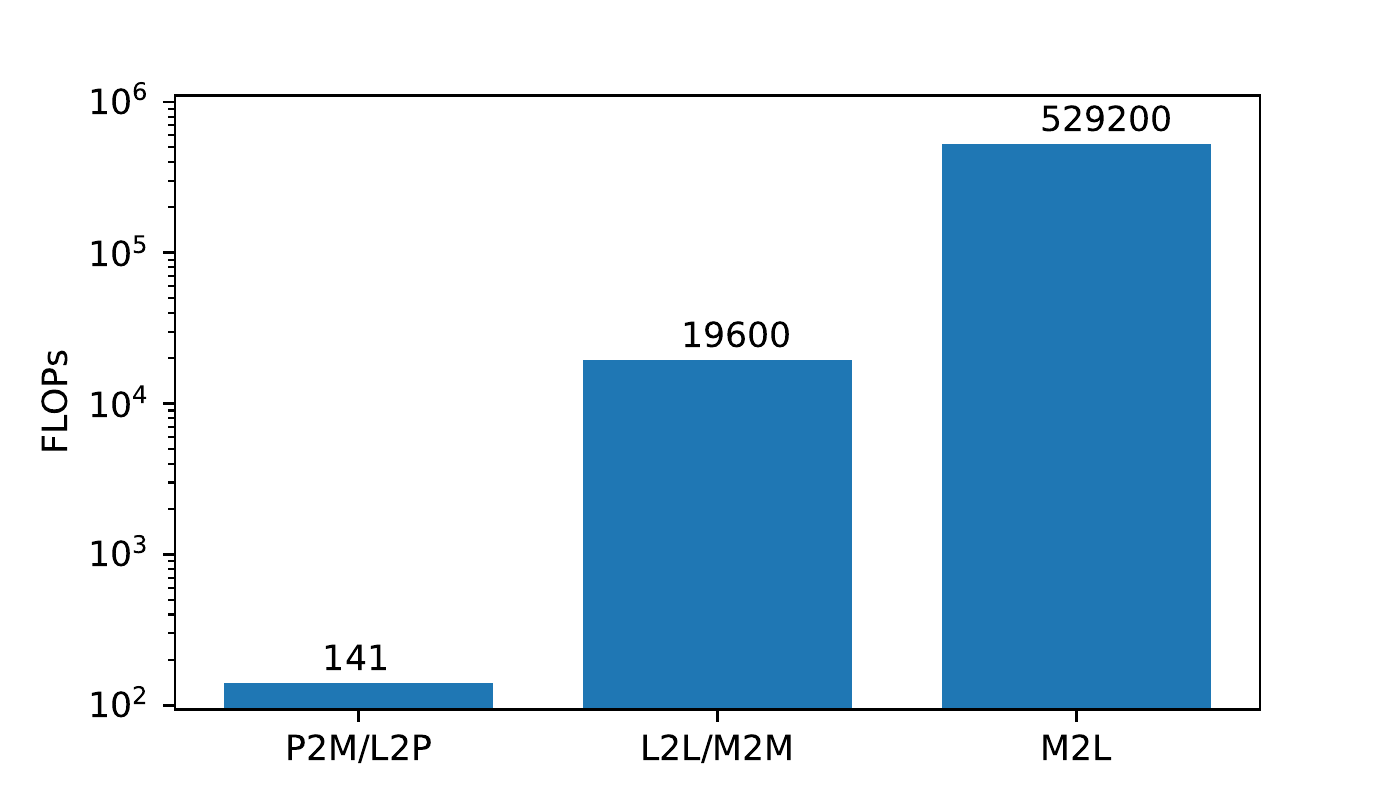}
    \caption{Under the assumption that the number of grid cells is roughly the same as $N$ and $M$, the graph shows a comparison of the number of FLOPs in $log$-space required for each step of the algorithm. The $M2L$ procedure requires the largest amount of computation by far.}
    \label{fig:flop_compare}
\end{figure}
Thus, in reality, the computational complexity of the $FC^2T^2$ expansion is in $\mathcal{O}(106N + 176M + C)$ with a very large constant coefficient $C$ that is mostly affected by the grid granularity level. This has direct implications on which types of problems are suitable for the algorithm. In general, expanding source locations and weights can be seen as trading off memory for computation. In the case that a model would need to be served to millions of users, the respective expansion could just be kept in memory and potentially large performance gains could be achieved in comparison to e.g. Neural Networks. Even a modestly sized Neural Network oftentimes requires $>1$m FLOPs for a single evaluation. Thus, in a scenario where a single static model needs to serve a large amount of requests, performance gains of $5,000\times$ could potentially be achieved in comparison to a modestly sized Neural Network. More generally, the $FC^2T^2$ expansion is suitable for problems that require repeated evaluations. As we will show later, there seems to be an abundance of problems in Graphics and Vision that have this property.

\subsection{Future Work}
\label{fucking_kernels}
\subsubsection{Decomposable Kernels and Spatially Separable Convolutions}
The fact that the $M2L$ subroutine is by far the most expensive is well known in the Fast Multipole Method literature and strategies to mitigate this have been proposed. The arguably easiest strategy is to assume the kernel function $\psi$ to be separable along the spatial dimensions, i.e. $\psi(x,y,z) = h(x)h(y)h(z)$. This assumption is true for e.g. the Gaussian kernel, since $\exp(-\alpha(x^2 + y^2 + z^2)) = \exp(-\alpha x^2)\exp(-\alpha y^2)\exp(-\alpha z^2)$. Once again, the strategy to achieve this speed up is to collect terms.
\begin{align*}
    L[p, x,y,z] &= \sum_{p'}^\Rho \sum_i \sum_j \sum_k M[p', x+i, y+j, z+k] k[p,p',i,j,k]\\
            &= \sum_{p'}^\Rho \sum_i \sum_j \sum_k M[p', x+i, y+j, z+k] k_x[p,p',i]k_y[p,p',j] k_z[p,p',k]\\
            &= \sum_{p'}^\Rho \sum_i k_x[p,p',i] \sum_j k_y[p,p',j] \underbrace{\sum_k M[p', x+i, y+j, z+k] k_z[p,p',k]}_{ = L_z[p,p',x+i,y+j,z]}\\
            & = \sum_{p'}^\Rho \sum_i k_x[p,p',i] \underbrace{\sum_j k_y[p,p',j] L_z[p,p',x+i,y+j,z]}_{ = L_y[p,p',x+i,y,z]}\\
            & = \sum_{p'}^\Rho \sum_i k_x[p,p',i] L_y[p,p',x+i,y,z] \\
\end{align*}
Assuming a 3D expansion, computing $L_z$, $L_y$ and $L$ requires $3*6 \Rho^2$FLOPs in total as opposed to $6^3 \Rho^2$FLOPs resulting in a potential speed-up of factor $12\times$. In the context of Machine Learning, the assumption of spatial separability seems to do little harm. In fact, all experiments conducted in this paper are based on Gaussian kernels that are spatially separable. However, an efficient implementation of a convolution with separable kernels is not trivial. There seems to be a common misconception (or misnomer) in the Machine Learning literature that spatially separable convolutions can be implemented by chaining 1D convolutions. However, chaining 1D convolutions assumes a different factorization as the one introduced above. To give a 2D example, the composition of two 1D convolutions computes:
\begin{align*}
    L[p, x,y] = \sum_{p'} \sum_i \sum_j (\sum_{p''} k_x[p,p'',i] k_y[p'',p',j]) M[p', x+i, y+j]
\end{align*}
In a sense, chaining 1D convolutions results in a spatially separable but depth-wise intertwined convolution because inner products instead of regular products between kernels are computed (at no additional cost). Thus, unless the depth of the kernel is 1 like in the case of the Sobel filter that is commonly used as an illustrative example of a spatially separable convolution, chaining convolutions requires a different factorization of the kernel. In fact, to the best of our knowledge, the $M2L$ kernel cannot be factored in such a way that it can be expressed as a composition of 1D convolutions. Furthermore, an efficient GPU implementation of a spatially separable convolution in the sense described above most likely requires a lower level than is currently possible with JAX primitives, most likely in CUDA or another low-level framework like e.g. Triton~\citep{tillet2019triton}. 

\subsubsection{Even/Odd-Indexed Convolutions}
As described above, the $M2L$ procedure requires a convolution with a kernel that depends on whether or not the voxel index is even or odd in each spatial dimension. In 3D, this is currently implemented by convolving the $M$ expansion with 8 different kernels (one for each combination of odd/even in 3D) with a stride of $(2,2,2)$ and subsequent interleaving of the results. In theory this operation requires the same number of FLOPs as a regular convolution, in practice however, it leads to a slow down of approximately $2\times$. In a similar manner as the speed-up that could potentially be gained from decomposing the kernel $\phi$, we believe an efficient GPU implementation could significantly improve the wall time of the approach. 
To sum up, if put together, we believe that the $M2L$ step responsible for approximately 90\% of the wall time\footnote{depending on the number of source and target locations} could be sped up by a factor of up to $24\times$ assuming a decomposable $\phi$ and careful implementation. A $24\times$ speed up of the $M2L$ subroutine, assuming a reasonable number of source and target locations, could lead to an overall speed up of $7$ to $8\times$.

\newpage
\section{Explicit Taylor Layer}

In the previous section, we introduced algorithms that allow for the efficient computation of local 3D Taylor expansions on a fixed grid of a continuous convolutional operator. For the remainder of the paper, we will introduce algorithms that make use of this algorithm in the context of Gradient Based Learning. Specifically, we derive the JVPs required for the backpropagation algorithm for computational layers that could potentially be used within Neural Networks. As we will show shortly, not only is it computationally intractable to compute the Jacobian similarly to the ordinary convolutional layer, but unlike the ordinary convolutional layer, directly computing the JVPs is also computationally intractable. However, we will show that the JVPs can be approximated by the $FC^2T^2$ expansion introduced previously. Similarly to the ordinary convolutional layer, the JVPs will be computed by an operation similar to transposition.\\

We begin by introducing an explicit layer, the simplest computational layer that internally makes use of the $FC^2T^2$ expansion described earlier. This layer outputs:
\begin{align*}
    y = f(q;p,w) = \begin{bmatrix}
    \sum_n^N \phi(q_1,p_n) w_n, & \dots, & \sum_n^N \phi(q_m,p_n) w_n
    \end{bmatrix}
\end{align*}
with $q \in \mathbb{R}^{M,3}$, $p \in \mathbb{R}^{N,3}$ and $w \in \mathbb{R}^{N,C}$. Note that $y$ has the functional form required for the $FC^2T^2$ and can therefore be approximated efficiently. In general, any combination of $p$, $q$ and/or $w$ could be used as data, model parameters or inputs from a previous computational layer.\\

We will exploit the following properties of the intermediate expansion for additional speed-ups:
\begin{itemize}
    \item Expansion Recycling: Only a single expansion is required to evaluate $f(q_1; p,w)$ and $f(q_2; p,w)$. In general, if the arguments behind the semi-colon of $f$ remain unchanged, the previously computed expansion can be recycled.
    \item Partial Derivatives: If the computational cost for expanding $p$ and $w$ has been paid, not only can $f$ be evaluated efficiently at any location $q$ but also $\nabla f$ (with lower precision however).
\end{itemize}

\subsection{Jacobian Vector Product}
\label{jvp:explicit}
Let $\overline{y} \in \mathbb{R}^M$ be the tangent vector (the incoming error propagated backwards from the subsequent layer) at which the JVP needs to be evaluated, then computationally efficient strategies to obtain the following quantities are required for the explicit layer to be used in the context of Gradient Based Learning:
\begin{align*}
    \overline{q} = \overline{y} \fracpartial{f}{q}; \ \overline{p} = \overline{y} \fracpartial{f}{p}; \ \overline{w} = \overline{y} \fracpartial{f}{w}
\end{align*}
The general strategy is the following: Even direct computation of the JVP will be intractable but the JVP can be brought into the functional form of $f$ and can therefore be approximated using the $FC^2T^2$ expansion. Because of the simplicity of $f$, deriving the JVPs is trivial assuming a symmetric kernel in the sense that $\phi(p,q) = \phi(q,p)$:
\begin{align*}
    \overline{q} &=
        \begin{bmatrix} \sum_i^N \nabla \phi(q_1,p_i)w_i & \dots & 0 \\
            \dots & \dots & \dots \\
            0& \dots& \sum_i^N \nabla \phi(q_N,p_i)w_i
        \end{bmatrix} \begin{bmatrix} \overline{y}_1 \\ \dots \\ \overline{y}_M\end{bmatrix} 
        = \overline{y} \odot \nabla f(q; p,w)\\
    \overline{w} &= \begin{bmatrix} \phi(q_1,p_1) & \dots & \phi(q_M,p_1) \\
            \dots & \dots & \dots \\
            \phi(q_1,p_N)& \dots& \phi(q_M,p_N)
            \end{bmatrix} \begin{bmatrix} \overline{y}_1 \\ \dots \\ \overline{y}_M\end{bmatrix} = f(p; q,\overline{y})\\
 \overline{p} &=
     \begin{bmatrix} \nabla \phi(q_1,p_1)w_1 & \dots & \nabla \phi(q_M,p_1)w_1 \\
        \dots & \dots & \dots \\
        \nabla \phi(q_1,p_N)w_N& \dots& \nabla \phi(q_M,p_N)w_N
    \end{bmatrix} \begin{bmatrix} \overline{y}_1 \\ \dots \\ \overline{y}_M\end{bmatrix} = w \odot \nabla f(p;q,\overline{y})
\end{align*}

Note that this entails that for every iteration of the backpropagation algorithm two expansions need to be computed. Because the gradients w.r.t. $q$ require an expansion for $p$ and $w$, the expansion computed in the forward pass can be recycled. A second expansion is required for gradients w.r.t. $p$ and $w$. In the forward pass, weights are inserted into the expansion at source locations to be evaluated at target locations whilst during the backward pass, errors are inserted into the expansion at target locations to be evaluated at source locations.

\subsection{How to use the implementation?}
In the following we will quickly explain how to use the JAX implementation of the explicit layer. As described earlier, we provide a high level API that should be easy to use given some experience with auto-differentiation frameworks. In a first step, the kernel, level and order of the $FC^2T^2$ algorithm that is used internally needs to be specified. After that, the resulting layer can be thought of as a differentiable transformation similar to an activation function with multiple inputs. The following code is pseudo but not far from a working example:
\begin{minted}{python}
    from fc2t2.explicit import get_layer
    
    func = lambda pkg: lambda x,y,z: pkg.exp(-5*(x**2 + y**2 + z**2))
    layer = get_layer(func, levels = 4, rho = 4)

    def loss(weights, x, y):
        q,p,w = neural_net1(x, weights[0])
        y1 = layer(q,p,w)
        y_hat = neural_net2(y1, weights[1])
        return jnp.mean((y-yhat)**2)
        
    grad = jax.grad(loss, argnums=(1,))(weights, x, y)
\end{minted}

\subsection{Applications}
The explicit layer introduced above has many potential applications and instantiations based on which inputs are assumed to be trainable parameters, data and inputs from previous layers. In the following, we discuss some applications theoretically but show empirical results in the context of Vision and Graphics. Specifically, we show how the explicit layer can be used to fit a signed distance function similar to the DeepSDF~\citep{Park_2019_CVPR} approach but orders of magnitudes faster. Note that fast summation algorithms have previously been applied to learn object descriptors as described in \citep{carr2001reconstruction}.

\subsubsection{Rank Constrained High dimensional Linear Layer}

When adopting the linear algebra view on the FMM algorithm~\citep{yokota2015fast}, a linear layer can be devised that is similar to the regular convolutional layer or a low-rank layer. The idea is the following: The regular convolutional layer is a sparse linear layer whose sparsity patterns are induced by the kernel shape. For example, if a kernel size of 3 is used, then the corresponding convolutional layer could be implemented as a sparse linear layer with $3$-element blocks on the diagonal in the case of a 1D convolution. This layer naturally is low-rank. Similarly, one could envision a linear layer whose rank is constrained in order to gain computational speed-ups in the same vein as the degenerate kernel example introduced earlier. Without non-linearities, such a layer would make little sense because the output of the layer would also be low-rank and therefore have redundant information. In a sense, the non-linearity that typically follows a linear layer that is either low-rank or increases the output dimensionality inflates the rank of its outputs.
\begin{align*}
    y_1 = W_{lr}x, \ \ y_2 = \sigma(W_{lr}x), \ \ y_3 = \psi(W_{lr})x
\end{align*}
In general, given a suitable non-linearity $\sigma$, the rank of $y_1$ is always smaller or equal compared to the rank of $y_2$ if $W_{lr}$ is low-rank. When choosing $p$ and $q$ as model parameters, the explicit layer could be used in a similar fashion. However, the non-linearity enters in a different way. Specifically, the non-linearity would be applied to the low-rank matrix before multiplying with the input $x$ but even though $\psi(W_{lr})$ is full rank given a suitable kernel, the price for multiplying $x$ and $\psi(W_{lr})$ would be reduced significantly. However, because such a layer is still linear in its inputs, in a multi-layer setting, a non-linearity would still need to be applied afterwards. In the experience of the authors, using the explicit layer in such a way turned out to be fruitless. A simple low-rank layer seemed to converge faster and to better solutions. Adding a second source of non-linearity did not seem improve convergence or solution quality.

\subsubsection{Compressed Sensing}
If we assume $p$ and $w$ to be data and $q$ to be parameters, the explicit layer could potentially be used for compressed sensing~\citep{baraniuk2007compressive} or optimal sensor placement~\citep{meo2005optimal,chmielewski2002theory}. Assume a training set consisting of pairs of $p$ and $w$ describing a low-dimensional phenomenon of interest, such as e.g. concentrations of chemicals or pollutants in the atmosphere on a specific day~\citep{mao2013ozone}. Since $q$ determines spatial locations where the phenomenon is being measured, optimizing $q$ yields optimal measurement locations. In practice, the explicit layer would be used as the input layer and the output would be fed into a classifier. Such a layer would also be approximately twice as fast because no additional expansion is required for the backward step as $\overline{q}$ only requires information that can be extracted from the expansion of the forward pass.

\subsubsection{Fitting a Signed Distance Function}
\label{sec:sdf}
\begin{figure}
    \centering
    \begin{subfigure}[t]{0.48\textwidth}
        \centering
        \includegraphics[width=\linewidth]{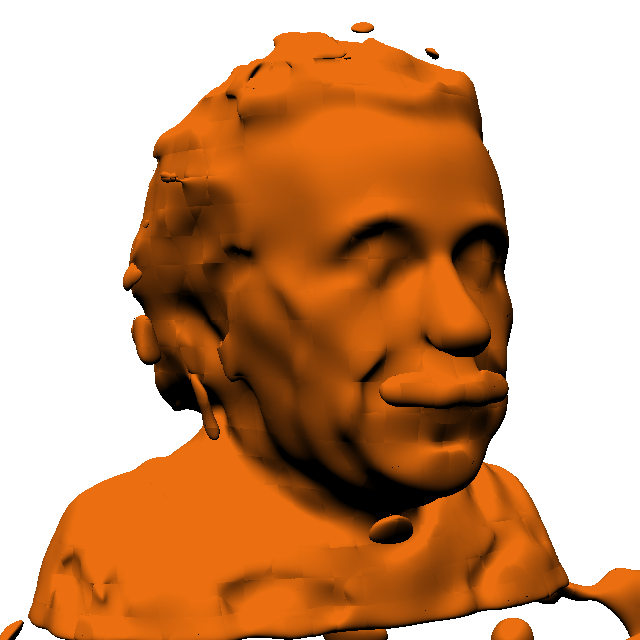} 
        \caption{Grid granularity level 4, $N=8m$, $M=10m$, $\alpha=200$, trained for 1400 epochs for approximately 3min (130ms/epoch) to an average MAE of $10.6E-4$} \label{fig:einstein4}
    \end{subfigure}
    \hfill
    \begin{subfigure}[t]{0.48\textwidth}
        \centering
        \includegraphics[width=\linewidth]{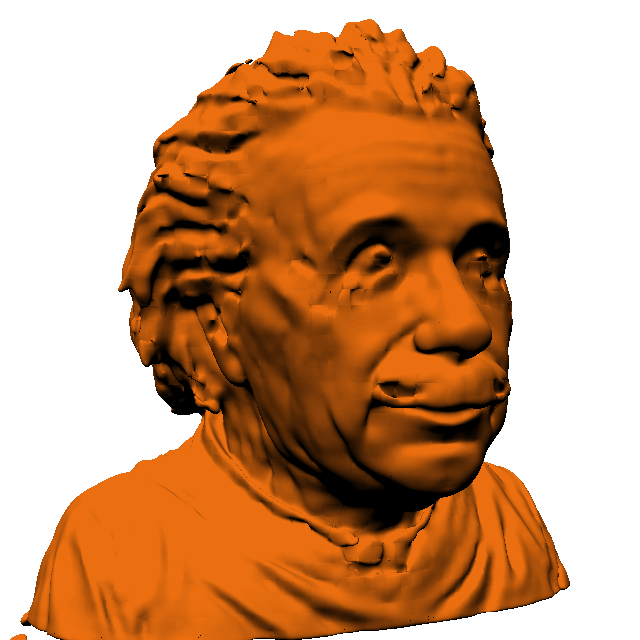} 
        \caption{Grid granularity level 5, $N=8m$, $M=10m$, $\alpha=1200$, trained for 2000 epochs for approximately 7.5min (225ms/epoch) to an average MAE of $7.2E-4$} \label{fig:einstein5}
    \end{subfigure}
    \begin{subfigure}[t]{0.48\textwidth}
        \centering
        \includegraphics[width=\linewidth]{result_plots/einstein_6.png} 
        \caption{Grid granularity level 6, $N=8m$, $M=10m$, $\alpha=4000$, trained for 1000 epochs for approximately 20min (1200ms/epoch) to an average MAE of $5.6E-4$} \label{fig:einstein6}
    \end{subfigure}
    \hfill
    \begin{subfigure}[t]{0.48\textwidth}
        \centering
        \includegraphics[width=\linewidth]{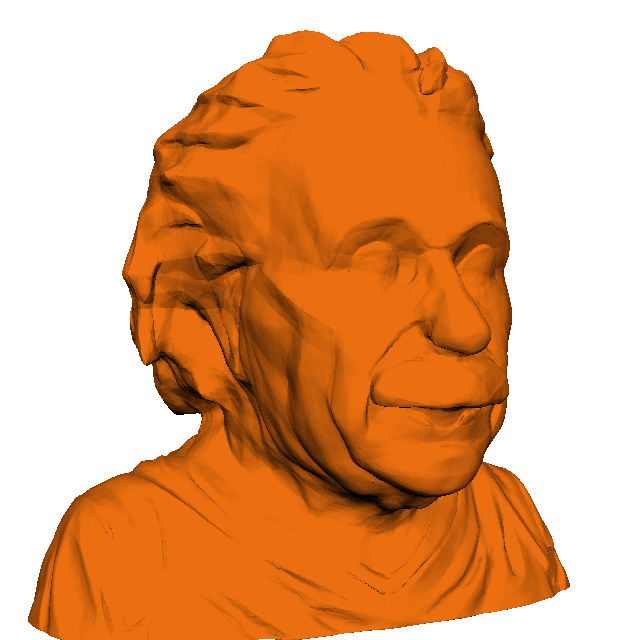} 
        \caption{The result of fitting a Neural Network with the architecture described in \citep{Park_2019_CVPR} trained to an MAE comparable to (c) in approximately 5.5 hrs (1333 epochs of 13s each).} 
        \label{fig:deepsdf}
    \end{subfigure}
    \caption{A visual comparison of the results when using the explicit layer to fit a SDF trained using $FC^2T^2$ expansion with different levels of the expansion and a Neural Network. Note that the inductive biases of the explicit layer and the NN approach are seem to differ as the NN seems to be biased towards smoothness. Larger versions of the resulting images are available in the appendix.}
    \label{fig:sdf}
\end{figure}

As described earlier, in this paper we focus on applications in Computer Vision and Graphics. By choosing $q$ to be data and $p$ and $w$ to be model parameters, we use the explicit layer to fit a signed distance function. Signed distance functions (SDFs) find applications as shape descriptors~\citep{carr2001reconstruction}. The value of the SDF signifies the shortest distance to the object. This implies that roots of a SDF determine the surface of the object. As the name suggests, this distance function is signed and negative values imply that a point is within an object. Neural Networks have recently been proposed to model SDFs~\citep{Park_2019_CVPR} but fitting SDFs has a long history and fast solvers have been employed for this task in the past~\citep{carr2001reconstruction}. We use the python package \verb+mesh_to_sdf+\footnote{\url{https://github.com/marian42/mesh_to_sdf}} that allows for sampling signed distance functions given a polygon mesh. The technique was proposed in~\citep{Park_2019_CVPR} and open-sourced by the authors of \citep{kleineberg2020adversarial}. We generate $10m$ samples from a triangle mesh describing a bust of Albert Einstein\footnote{\url{http://3dmag.org/en/market/item/3573/}}. Let $\hat{q}$ denote the sample locations and $\hat{d}$ distances of the locations to the object. We minimize the mean absolute error w.r.t. to $p$ and $w$, i.e. $|f(\hat{q}; p,w) - \hat{d}|$ and train on all 10m points jointly or to use Neural Network nomenclature, we use a batch size of 10m. One epoch takes between 130ms and 1.2s depending on the level of the expansion. We use a Gaussian kernel $\phi(x,y,z) = \exp(-\alpha (x^2 + y^2 + z^2)$ with varying $\alpha$ depending on expansion level. We do not exploit the fact that the Gaussian kernel is decomposable into its spatial dimensions, i.e. because the Gaussian kernel can be factored as $\phi(x,y,z)=h(x)h(y)h(z)$, $M2L$ could be decomposed, which could potentially lead to another $5$-$10\times$ speed-up. However, as described earlier, the implementation of this decomposition is non-trivial. Figure \ref{fig:sdf} shows the results and the reader is referred to the code for implementation details. For a granularity level of 4, the general shape of the object can be discerned but most details are missing. Increasing the granularity level to 5 reveals more features such as wrinkles in the face and shirt but fine details are still underresolved. At a granularity level of 6, fine details such as eye lids, wrinkles and facial hair are resolved much better. We compare to a Neural Network with the architecture used for DeepSDF~\citep{Park_2019_CVPR}. With a batch size of 16384 one epoch takes approximately 13s. Thus, the explicit layer based on $FC^2T^2$ is approximately 10x faster with a suboptimal implementation of the $M2L$ subroutine. Training the Neural Network for one epoch requires $36,741.12$GFLOPs whilst the explicit layer with granularity level 6 requires $1,317$GFLOPs (and $154$ GFLOPs for separable kernels) therefore resulting in a $28\times$ (and $237\times$ for separable kernels) reduction in FLOPs. However, because the algorithms optimize models that are imbued with different inductive biases such a comparison might make little sense. Indeed, we found that Neural Networks and the explicit layer to have vastly different inductive biases but believe the bias of the proposed layer to be preferable in the context of Vision and Graphics. Neural Networks tend to struggle to learn high frequency variations~\citep{rahaman2019spectral}, i.e. they tend to learn approximations with a low norm in function space. Whilst this property might be beneficial for some applications, it seems to be harmful in this context. Note that if we initialize $p$ and $w$ uniformly random between $(-1, 1)$, the output of the explicit layer will fluctuate rapidly. The explicit layer seems to learn high frequency features quickly. Indeed if we terminate the learning procedure prematurely, the resulting rendering seems to exhibit holes in Einstein's face suggesting that low frequency information is corrupted as opposed to a premature rendering of the Neural Network approach that exhibits closed surfaces that however lack fine details.\\

\textbf{The batch size conjecture}: Training a Neural Network requires careful consideration of the interdependence between batch size and learning rate. When fixing the learning rate, one epoch with a smaller batch size usually results in longer training times but also a lower error. We did not experience such an interaction when using the explicit layer in the way described above. We usually used the largest batch size that seemed reasonable. We believe this to be due to two properties of the explicit layer: 1.) Gradients have local support and 2.) the layer is linear and therefore convex in $w$. Because during the backward pass, errors are inserted into the expansion at target locations and the kernel has a considerably small and local support, the error at one location does not interact with errors at locations far enough away. Thus, model parameters only need to agree locally unlike Neural Network parameters whose support is potentially global (a single weight might influence outputs over the entire domain). Furthermore, the application of Gradient Descent in the context of Neural Networks induces \emph{non-linear} dynamics of model parameters. Thus, using a different size (or even just ordering) of mini-batches might lead to convergence to a different local minimum. The explicit layer on the other hand is linear and therefore convex in $w$. This entails that assuming fixed $p$ and an appropriate learning rate, Gradient Descent will converge to the same unique solution irrespective of the batch size.

\newpage
\section{Root Implicit Taylor Layer}
In this section, another computational layer that internally makes use of the $FC^2T^2$ expansion is introduced. This layer is implicit in the sense that its output is implicitly defined. To be precise, the layer outputs quantities related to the root of a function along a line, or to use Graphics/Vision nomenclature, a ray. Let $\mathbf{r}, \eye \in R^3$ be a direction and position vector respectively. $\eye$ can be understood as the position of a pinhole camera and $\mathbf{r}$ as a viewing direction. For most applications, one can safely assume multiple $\mathbf{r}$, usually one for each pixel in a 2D image, but the following derivations assume a single $\mathbf{r}$, however generalization is trivial.
Similar to the application of the explicit layer to model signed distance functions described in section \ref{sec:sdf}, we assume a function $f$ whose roots define the surface of a 3D object. The root-implicit layer can in principle be used to output any combination of two quantities related to roots of $f$. First, the distance between the position of the pinhole camera $\eye$ and the object along the ray, i.e. the ray length $y_l$, and second, the surface gradient $y_\nabla$ (a scalar-multiple of the surface normal) at the root, a quantity required for many rendering tasks.
\begin{align*}
    \text{Ray length: \ \ }&y_l = x\\
    \text{Surface gradient: \ \ }&y_\nabla = \nabla f(\eye + x \mathbf{r}; p,w)\\
    \text{s.t.: \ \ } &f(\eye + x \mathbf{r}; p,w) = 0
\end{align*}

\subsection{Root Finding}
Before deriving the JVP of the root implicit layers, a fast algorithm that allows for the extraction of roots along a ray is being introduced that acts directly on the intermediate Taylor representation of $f$. This algorithm does not assume $f$ to be a proper signed distance function in the sense that function values contain exact information about the distance to a root. In general, for the applications described in this paper, it is difficult to guarantee $f$ to be a proper signed distance function because its parameters are being updated by Gradient Descent and projection onto a proper signed distance function is not trivial for arbitrary kernels $\phi$~\citep{carr2001reconstruction}.
\begin{figure}[h]
    \centering
    \begin{subfigure}[t]{0.45\textwidth}
        \centering
        \includegraphics[width=0.95\linewidth]{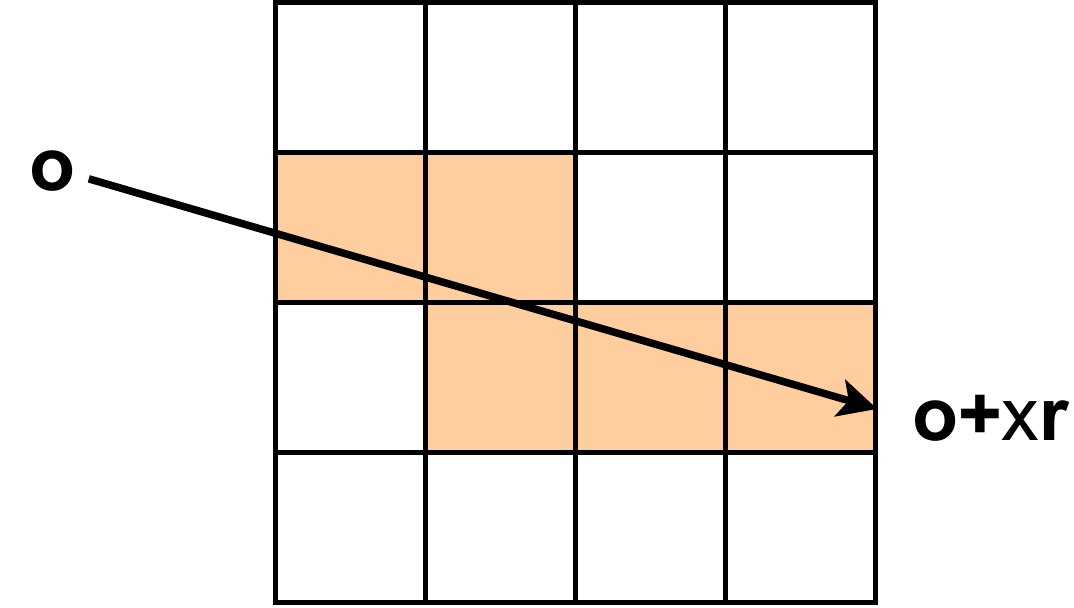} 
        \caption{\textbf{Root Finding:} In order to find a root along a ray from $\eye$ in the direction $\mathbf{r}$ all boxes that intersect the ray are extracted. A 1D polynomial of order $\rho$ is then extracted and checked for a root. The graphic illustrates an example in 2D.} \label{fig:root}
    \end{subfigure}
    \hfill
    \begin{subfigure}[t]{0.45\textwidth}
        \centering
        \includegraphics[width=0.95\linewidth]{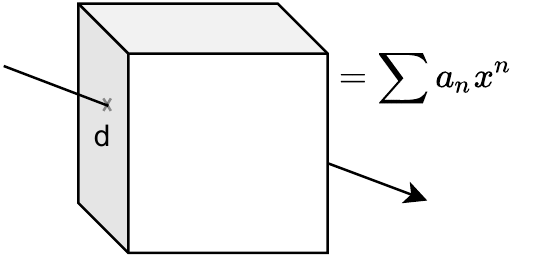} 
        \caption{\textbf{Line2Poly:} Any ray through a box representing a 3D Taylor series expansion can be converted to a 1D polynomial of order $\rho$. For 1D polynomials of order $\rho \leq 4$ analytic and fast solutions for roots are available.} \label{fig:line2poly}
    \end{subfigure}
    \caption{The root finding algorithm intersects the ray with the boxes and then converts 3D Taylor expansions into univariate polynomials for which analyic solutions exist assuming an order $\leq 4$.}
    \label{fig:root_finding}
\end{figure}
Recall that the intermediate representation of the $FC^2T^2$ expansion outputs a grid whose cells contain a 3D Taylor series expansion at its center. Because the expansion is in 3D, we refer to a cell on the grid as a box or voxel. The algorithm finds the first root along a ray by enumerating the boxes that are intersected by the ray. As we will show shortly, any line through a 3D Taylor expansion can be converted to a univariate polynomial of the same order. Furthermore, for orders $\leq 4$, analytic closed-form solutions for the roots exist~\citep{ferrari}. We assume familiarity with voxel-ray intersection and root-finding algorithms which entails that the only missing ingredient for the algorithm is the conversion of a line through a 3D Taylor expansion to a 1D polynomial.

\subsubsection{Line to polynomial}
\label{prelim:r2p}
As described earlier, the intermediate representation of the $FC^2T^2$ approximates a function globally with a series of local 3D Taylor expansions distributed on a grid. Every 3D Taylor expansion stores partial derivatives up to a certain order evaluated at the center of the cell. Let $L_{n_1,n_2,n_3}$ be the coefficient representing $\partial^{n_1,n_2,n_3} f$ and $\mathbf{c}$ the center of the box which entails that:
\begin{align*}
    f(q) = \sum_{n_1+n_2+n_3 \leq \rho} \frac{L_{n_1,n_2,n_3}}{n_1!\ n_2!\ n_3!} (q_1-c_1)^{n_1}(q_2-c_2)^{n_2}(q_3 - c_3)^{n_3}\\
\end{align*}

Assume that the ray $\eye + \mathbf{r}x$ intersects with the box at location $\mathbf{d}$ in the coordinate frame of the box (center of box is origin) as shown in Figure \ref{fig:line2poly}. We are interested in the coefficients of the following univariate polynomial:
\begin{align*}
    f(\mathbf{d} + \mathbf{r}x) =& \sum_{n_1+n_2+n_3 \leq \rho} \frac{L_{n_1,n_2,n_3}}{n_1!\ n_2!\ n_3!} (r_1x+d_1)^{n_1}(r_2x+d_2)^{n_2}(r_3x + d_3)^{n_3}\\
\intertext{Making use of the binomial theorem gives us:}
=& \sum_{n_1+n_2+n_3 \leq \rho} \frac{L_{n_1,n_2,n_3}}{n_1!\ n_2!\ n_3!} \sum _{i,j,k=0}^{n_1,n_2,n_3}{n_1 \choose i}d_1^{n_1-i}r_1^{i} {n_2 \choose j}d_2^{n_2-j}r_2^{j}{n_3 \choose k}d_3^{n_3-k}r_3^{k} x^{i+j+k}\\
    =& \sum_{n=0}^\rho a_n x^n
\intertext{with}
    a_n =& \sum_{n_1+n_2+n_3 \leq \rho} \frac{L_{n_1,n_2,n_3}}{n_1!\ n_2!\ n_3!} \left(\sum_{i+j+k = n} (d_1^{n_1-i} r_1^{i}) (d_2^{n_2-j} r_2^{j}) (d_3^{n_3-k} r_3^{k}) {n_1 \choose i}{n_2 \choose j}{n_3 \choose k}\right)
\end{align*}


For $\rho = 4$, a na\"ive implementation of this operation requires 1465 FLOPs. When applying sympy's~\citep{sympy} common subexpression elimination, the computational costs can be reduced to $668$FLOPs.

\subsubsection{Fast root finding on ray}

The following pseudo-code summarizes the root finding algorithm.
\begin{minted}{python}
    def find_root_on_ray(ray, L):
        for box on ray:
            poly = line2poly(L[box], ray)
            found_root, root = ferrari(poly)
            if found_root:
                return True, root
        return False, 0.
\end{minted}

 Note that Ferrari's method~\citep{ferrari} for finding roots of quartic polynomials requires $136$FLOPs. Any ray through an $N \times N \times N$ grid intersects with at most $3N$ boxes. This entails that in the worst case scenario, i.e. when no root was found, the root finding procedure requires $3N(668 + 136)$FLOPs per ray. For a grid granularity level of 6 ($128 \times 128 \times 128$ boxes), this is equal to $308,736$FLOPs. Recall that a single evaluation of a Neural Network with the DeepSDF~\citep{Park_2019_CVPR} architecture requires $3,674,112$FLOPs. Thus, once the price for the expansion has been paid, finding a root along a ray requires about $10\times$ less FLOPs than a single evaluation of the Neural Network.

\subsection{Jacobian Vector Product}

In the following, the Jacobian Vector Products required for Gradient Based Learning are derived. Because the output of the layer is defined implicitly, we will make use of the Implicit Function (or Dini's) Theorem (IFT)~\citep{krantz2013implicit} similar to the approach described in~\citep{amos2017optnet}. Under mild conditions, the IFT states that:
\begin{align*}
    y(z) &= x \text{\ \ s.t.\ \ } f(x,z) = 0\\
    f(x,z) &= 0 \rightarrow f(y(z),z) = 0 \rightarrow \fracpartial{f}{x}\fracpartial{y}{z} + \fracpartial{f}{z} = 0\\
    \text{therefore, } \fracpartial{y}{z} &= - \left[\fracpartial{f}{x}\right]^{-1} \fracpartial{f}{z}
\end{align*}
This derivation reveals some of the conditions for the IFT to apply, e.g. that $\fracpartial{f}{x}$ be invertible.\\

Let $f$ be of the functional form that its evaluation can be sped up by a variant of the Fast Multipole Method, i.e.:
\begin{align*}
    f(q;p,w) = \begin{bmatrix}
    \sum_n^N \phi(q_1,p_n) w_n, & \dots, & \sum_n^N \phi(q_m,p_n) w_n\\
    \end{bmatrix} \in \mathbb{R}^M
\end{align*}
Note that the IFT does not hold for arbitrary roots of $f$, since:
\begin{align*}
    y(p,w) &= q \text{\ \ s.t.\ \ } f(q;p,w) = 0\\
    \fracpartial{y}{p} &= - \left[\fracpartial{f}{q}\right]^{-1} \fracpartial{f}{p}\\
    \text{but } \fracpartial{f}{q} &\in \mathbb{R}^{3M\times M}
\end{align*}
This entails that $\fracpartial{f}{q}$ is not invertible because it is underdetermined. However, as we will we will show in the next section, invertibility can be ensured by assuming that the root lays on a ray. This is well known in the Graphics/Vision literature~\citep{foley1996computer}.

\subsubsection{Ray length JVP}
Let $y_l(p,w) = x \text{\ \ s.t.\ \ } f(\eye + x \mathbf{r};p,w) = 0$ and $\overline{y} \in \mathbb{R}^M$ be the tangent vector (the incoming error propagated backwards from the subsequent layer), then the following JVPs need to be derived for Gradient Based Learning:
\begin{align*}
    \overline{p} = \overline{y_l} \fracpartial{y_l}{p};\ \ \overline{w} = \overline{y_l} \fracpartial{y_l}{w}
\end{align*}
In the following, the JVP w.r.t. to $w$ is derived but the JVP w.r.t. $p$ is analogous and follows the same pattern as described in section \ref{jvp:explicit}.
\begin{align*}
    y_l(p,w) &= x \text{\ \ s.t.\ \ } f(\eye + x \mathbf{r};p,w) = 0\\
    \fracpartial{y_l}{w} &= - \left[\fracpartial{f}{x}\right]^{-1} \fracpartial{f}{w}
\end{align*}
$\displaystyle\fracpartial{f}{w}$ has been derived in section \ref{jvp:explicit}. The only missing ingredient therefore is $\displaystyle\left[\fracpartial{f}{x}\right]^{-1}$.
\begin{align*}
    \fracpartial{f}{x} = \fracpartial{}{x} \sum_n^N \phi(\eye + x \mathbf{r},p_n) w_n
\end{align*}
Making use of the assumption that $\phi(p,q) = \psi(p_1 - q_1, p_2-q_2, p_3-q_3)$
\begin{align*}
    \fracpartial{}{x}\phi(\eye + x \mathbf{r},p_n) &= \fracpartial{}{x} \psi(o_1 + xr_1 - p_1, o_2 + xr_2 - p_2, o_3 + xr_3 - p_3)\\
    & = r_1 \partial^{1,0,0} \psi(\eye + x \mathbf{r}, p_n) + r_2 \partial^{0,1,0} \psi(\eye + x \mathbf{r}, p_n) + r_3 \partial^{0,0,1} \psi(\eye + x \mathbf{r}, p_n)\\
    & = \bdot{\mathbf{r}}{\nabla \phi(\eye + x \mathbf{r},p_n)}\\
    \text{thus, } \fracpartial{f}{x} &= \bdot{\mathbf{r}}{\nabla f_q} \text{ with } \nabla f_q = \nabla f(\eye + x \mathbf{r}; p,w) \rightarrow \left[\fracpartial{f}{x}\right]^{-1} = \frac{1}{\bdot{\mathbf{r}}{\nabla f_q}}
\end{align*}
As mentioned earlier, the derivations assume a single ray but for most practical applications, there are hundreds of thousands of rays. However, because there is no `cross-talk' between rays, the derivations remain unchanged when multiple rays are assumed, i.e. the projection can be performed `element-wise'. Similar to the explicit layer, even direct computation of the JVP is intractable. However, employing the same strategy, the $FC^2T^2$ expansion can be employed to approximate the JVP. Also note that $\nabla f_q$ can be computed from the expansion of the forward pass.
\begin{align*}
    \text{Let } \overline{\overline{y_l}} &= \frac{-\overline y_l}{\bdot{\mathbf r}{\nabla f_q}} \text{ and } q = \eye + x \mathbf{r} \text{ then, } \\
    \overline{w} &= f(p; q,\overline{\overline{y_l}}); \ \  \overline{p} = w \odot \nabla f(p;q,\overline{\overline{y_l}})
\end{align*}
Thus, it is possible to obtain the JVP using the $FC^2T^2$ expansion and projecting the gradients according to the IFT comes at \emph{almost} no additional cost in comparison to the explicit layer since $\nabla f_q$ can be computed from the forward-pass expansion.

\subsubsection{Surface gradient JVP}

For many applications in Vision and Graphics such as e.g. Inverse Rendering~\citep{marschner1998inverse}, knowledge about surface normals is paramount. Surface normals play an important role in many shading models such as e.g. the (Blinn-)Phong model~\citep{tuong1973illumination,blinn1977models}. As the name suggests, surface normals constitute the gradient at the surface of an object normalized to unit length. In the following, the JVPs required to update model parameters based on surface gradients are introduced. We derive the surface gradient instead of normal JVP mostly for convenience and the fact that the normalization step can be performed trivially in auto-differentiation frameworks. Similar to the previous layer, the surface of an object is encoded as the root of a function $f$, however, instead of outputting the distance between $\eye$ and the object, the surface gradient is returned.
\begin{align*}
    y_\nabla(p,w) &= \nabla f(\eye + x\mathbf{r}; p,w) \text{ s.t. } f(\eye + x\mathbf{r}; p,w) = 0\\
    &= \nabla f(\eye + y_l(p,w)\mathbf{r}; p,w)
\end{align*}
Similar to the previous layer, the JVPs required for Gradient Based Learning are:
\begin{align*}
    \overline{w} = \overline{y_\nabla} \fracpartial{y_\nabla}{p};\ \ \overline{p} = \overline{y_\nabla} \fracpartial{y_\nabla}{w}
\end{align*}
Again, we will derive the JVP w.r.t. $w$ because the JVP w.r.t. $p$ is analogous. Applying the chain rule of derivatives yields,
\begin{align*}
    \overline{y_\nabla} \fracpartial{y_{\nabla}}{w} &= \overline{y_\nabla} \left(\fracpartial{\nabla f}{y_l}\fracpartial{y_l}{w} + \fracpartial{\nabla f}{w}\right)\\
    &= -\overline{y_\nabla} \fracpartial{\nabla f}{y_l} \left[\fracpartial{f}{y_l}\right]^{-1} \fracpartial{f}{w} + \overline{y_\nabla} \fracpartial{\nabla f}{w}
\end{align*}
The quantities that have not been derived previously are $\displaystyle \fracpartial{\nabla f}{y_l}$ and $\displaystyle\fracpartial{\nabla f}{w}$. Recall that, $\nabla f = [\partial^{1,0,0} f, \partial^{0,1,0} f, \partial^{0,0,1} f]$ and let $\nabla^{i,j,k} f = [\partial^{i+1,j,k} f, \partial^{i,j+1,k} f, \partial^{i,j,k+1} f]$.
\begin{align*}
    \fracpartial{\nabla f}{y_l} &= [\bdot{\nabla^{1,0,0} f_q}{\mathbf{r}}, \bdot{\nabla^{0,1,0} f_q}{\mathbf{r}}, \bdot{\nabla^{0,0,1} f_q}{\mathbf{r}}]\\
    \fracpartial{\nabla f}{y_l} \left[\fracpartial{f}{y_l}\right]^{-1} &= \left[\frac{\bdot{\nabla^{1,0,0} f_q}{\mathbf{r}}}{\bdot{\nabla f_q}{\mathbf{r}}}, \frac{\bdot{\nabla^{0,1,0} f_q}{\mathbf{r}}}{\bdot{\nabla f_q}{\mathbf{r}}}, \frac{\bdot{\nabla^{0,0,1} f_q}{\mathbf{r}}}{\bdot{\nabla f_q}{\mathbf{r}}}\right] := \Delta f_q
    \intertext{Note that $\overline{y_\nabla} \in \mathbb{R}^3$ and let $\overline{\overline{y_\nabla}} = \bdot{\Delta f_q}{-\overline{y_\nabla}}$ which entails that, }
    \overline{y_\nabla} \fracpartial{\nabla f}{y_l} \left[\fracpartial{f}{y_l}\right]^{-1} \fracpartial{f}{w} &= f(p; q, \overline{\overline{y_\nabla}}).
\end{align*}
For the remaining quantity to be derived the following holds:
\begin{align*}
    \overline{y_\nabla} \fracpartial{\nabla f}{w} = \partial^{1,0,0} f(p; q, \overline{y}_1) + \partial^{0,1,0} f(p; q, \overline{y}_2) + \partial^{0,0,1} f(p; q, \overline{y}_3)
\end{align*}
In summary,
\begin{align*}
    \overline{y_\nabla} \fracpartial{y_\nabla}{w} &= f(p; q, \overline{\overline{y_\nabla}}) + \partial^{1,0,0} f(p; q, \overline{y}_1) + \partial^{0,1,0} f(p; q, \overline{y}_2) + \partial^{0,0,1} f(p; q, \overline{y}_3)\\
    \overline{y_\nabla} \fracpartial{y_\nabla}{p} &=  w \odot (\nabla f(p; q, \overline{\overline{y_\nabla}}) + \nabla^{1,0,0} f(p; q, \overline{y}_1) + \nabla^{0,1,0} f(p; q, \overline{y}_2) + \nabla^{0,0,1} f(p; q, \overline{y}_3))
\end{align*}

Thus, computing the JVPs for the surface gradients requires two expansions for the backward pass. One projection expansion and one three channel gradient expansion. The backward pass is therefore slower in comparison to the explicit and ray-length layer introduced earlier but still reasonably fast.
\subsection{Applications}

In the following, we showcase four potential applications of the root-implicit layer. The objective of the first two experiments is to extract a 3D representation from depth information. The third experiment combines the explicit and root-implicit layer to model RGBD images while the last application makes use of the surface normal gradients in the context of inverse rendering.

\subsubsection{Learning a depth field}
\begin{figure}[h]
    \centering
    \begin{subfigure}[t]{0.95\textwidth}
        \centering
        \includegraphics[width=\linewidth]{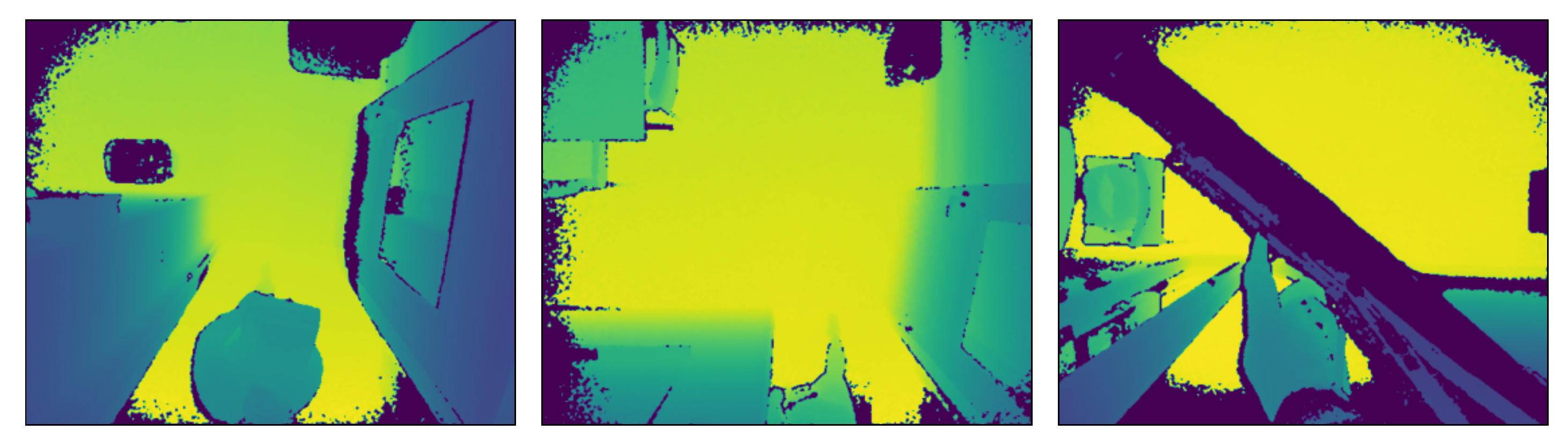} 
        \caption{Ground truth depth measurements taken from~\citep{flores2019dataset}. Dark blue pixel indicate missing or corrupted data.} \label{fig:cmu_example}
    \end{subfigure}
    \hfill
    \begin{subfigure}[t]{0.95\textwidth}
        \centering
        \includegraphics[width=\linewidth]{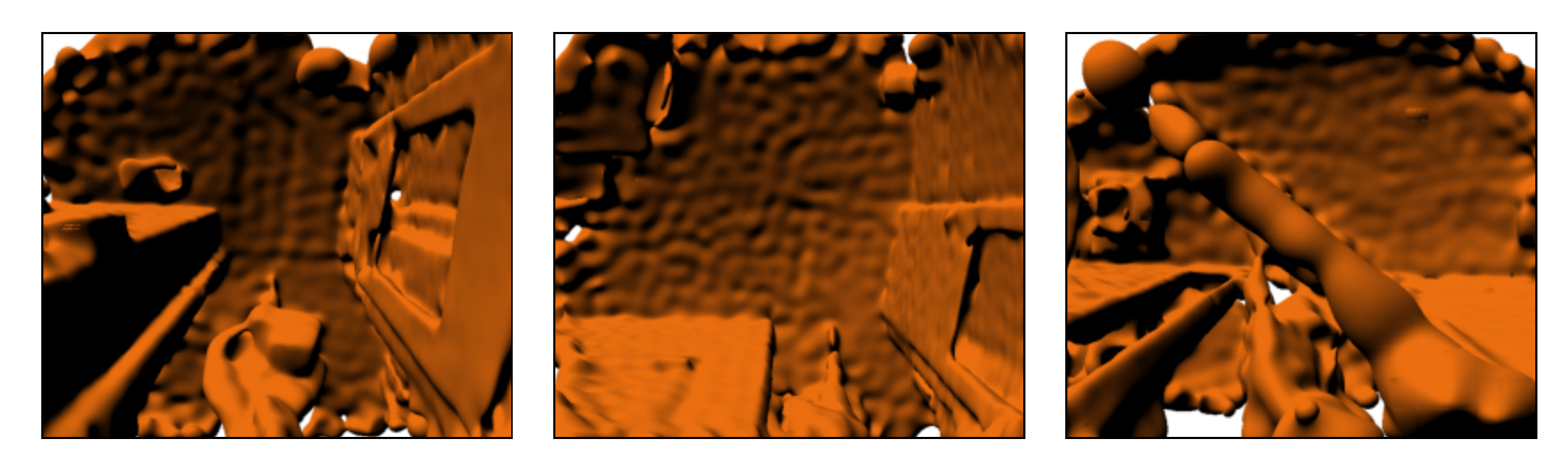} 
        \caption{A rendering of the depth field modeled with the $FC^2T^2$ depth layer. The respective errors are 0.5\%, 0.64\% and 0.54\%.} \label{fig:cmu_single_depth}
    \end{subfigure}
    \caption{Results when training the depth layer to model data collected from a vertically mounted Kinect sensor.}
    \label{fig:cmu_single_depth_fig}
\end{figure}
In a first experiment we extract a coherent 3D representation from images collected with a depth camera. The experiment is based on a data set collected by a vertically mounted depth sensor (Microsoft Kinect) above a class room door~\citep{flores2019dataset} intended to improve building energy efficiency and occupant comfort by estimating occupancy patterns. Figure \ref{fig:cmu_example} shows three example frames of the data set. The goal of the first experiment is to demonstrate the algorithms ability to extract coherent 3D representations given noisy depth information. The depth field collected by the camera is normalized to fit into the domain of the expansion, i.e. $(-1,1)$. The output of a depth camera measures the distance to objects in the field of view and is therefore amenable to the root-implicit layer that outputs ray length. Because the data is fairly low in resolution, we choose a level 5 expansion with a Gaussian kernel ($\alpha = 1000$) and minimize the mean absolute error for 300 iterations to an average error of approximately $0.5\%$. One iterations takes approximately 250ms implying a total training time of 75s per image. Special attention needs to be given to the initialization of $p$ and $w$. If the root-finding algorithm is unable to locate a root within the domain or if $f$ is negative at the first intersection of the domain and ray, the output of the layer and therefore its gradient for the corresponding pixel is undefined. In order to avoid `dead pixels', $p$ and $w$ should be initialized in such a way that every ray has a proper root within the domain. We achieve this by introducing a bias term and initializing $w$ to be relatively small. Furthermore, in order to suppress artifacts, we additionally regularize $w$ (L1) and the reader is referred to the code for implementation details. Figure \ref{fig:cmu_single_depth_fig} summarizes the results.

The code below shows how to use the depth layer. The layer needs to be provided with $p$, $w$, a bias term and the position of the pinhole camera (\emph{eye}) and viewing direction (\emph{gaze}).
\begin{samepage}
\begin{minted}{python}
from fc2t2.root_implicit import get_depth_layer
func = lambda pkg: lambda x,y,z: pkg.exp(-1000*(x**2 + y**2 + z**2))
depth_layer, expand, A = get_depth_layer(func, 5, 512)

def lossfn(p,w):
    depth_hat = depth_layer(p,w,0.05,eye,gaze) #0.05 is bias
    return jnp.mean(jnp.abs(depth_hat - depth_target))
\end{minted}
\end{samepage}

\subsubsection{Learning to learn a depth field}
\label{learn_to_learn}
The previous experiment demonstrated the ability of the proposed technique to model real-world and noisy depth data. However, processing a single frame required more than 1min rendering the approach too slow for any real-time application. In this experiment, we built upon the previous experiment but task a Neural Network with inferring optimal $p$ and $w$ that induce a given depth field. The Neural Network is trained in an auto-encoder fashion, i.e. it is presented with the desired depth field, produces parameters $p$ and $w$ which are fed into the depth layer. The mean absolute error between the desired depth field and the output of the depth field is then minimized. Because a Neural Network is tasked with predicting the parameters of another module, the experiment could be considered an instance of a hypernetwork~\citep{hypernetwork} or to use a more sober description, an operator.
\begin{figure}
    \centering
    \includegraphics[width=\linewidth]{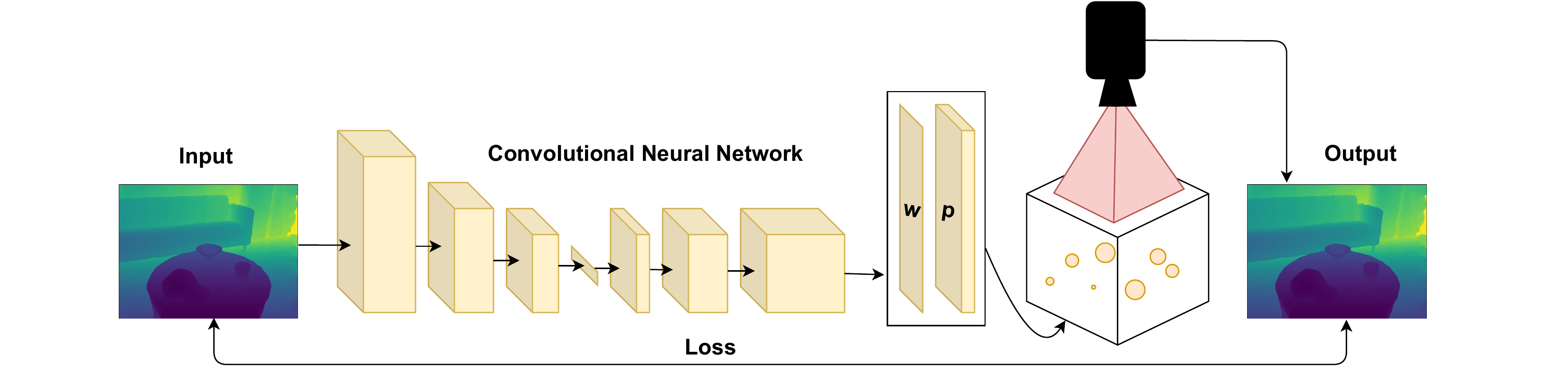}
    \caption{The autoencoder that employs the $FC^2T^2$ depth layer as its decoder. A regular Convolutional Neural Network is tasked with predicting optimal $p$ and $w$ that induce a given depth field.}
    \label{fig:auto_enc_bild}
\end{figure}
The pseudo code below shows the loss function which minimizes an auto-encoder loss.
\begin{minted}{python}
def lossfn(params, depth_target):
    p,w = depthNet.apply(params, depth_target)
    depth_hat = depth_layer(p,w,0.05,eye,gaze)
    err = jnp.mean(jnp.abs(depth_hat - depth_target))
    return err
\end{minted}

\begin{figure}
    \centering
    \includegraphics[width=\linewidth]{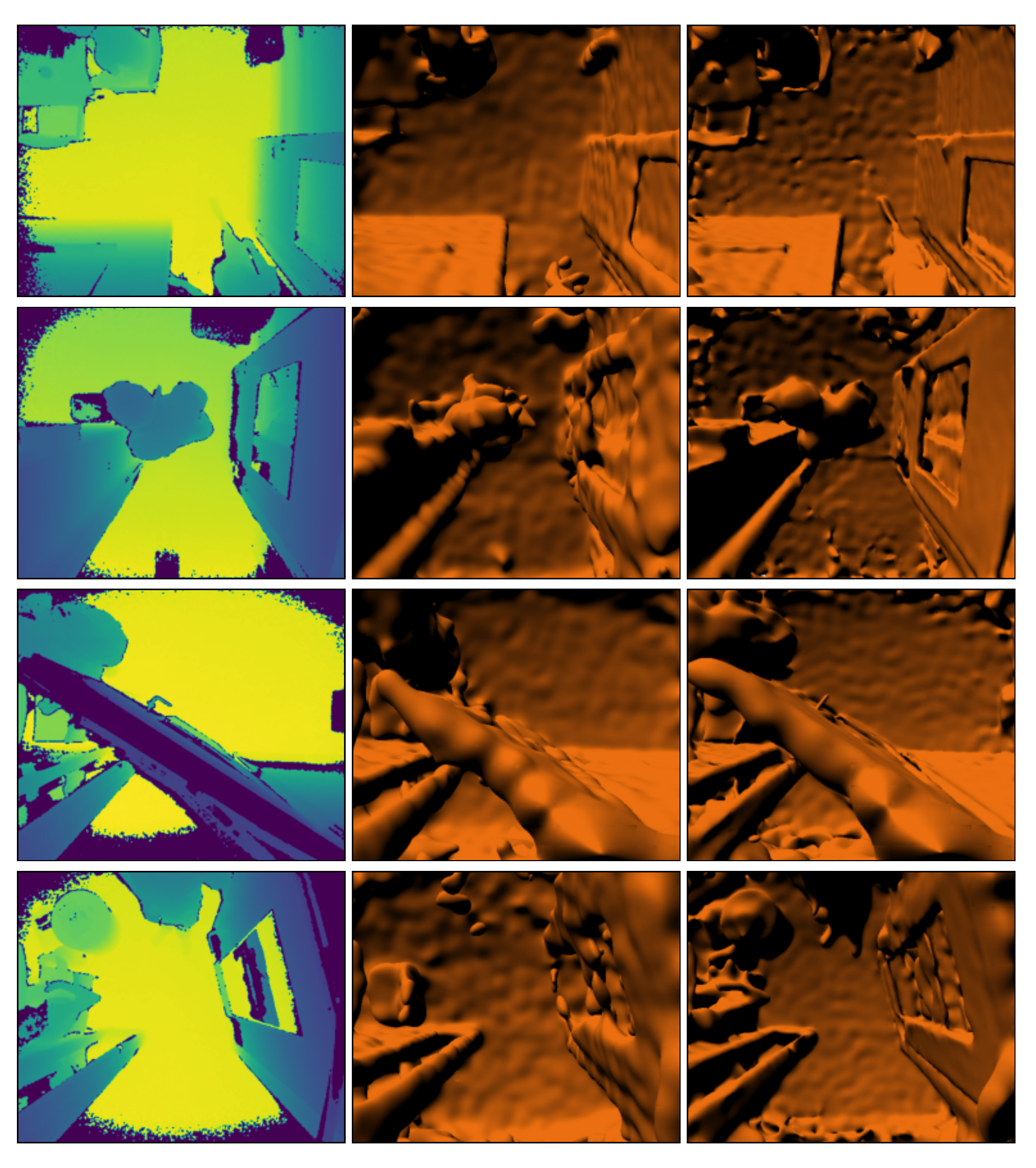}
    \caption{The left column shows the input (and target) to the deep Convolutional Neural Network (CNN). The middle column shows the depth field induced by the parameters predicted by the CNN and right column shows the depth field after being retouched by a few extra steps of Gradient Descent. Retouching reduces the error by approximately one order of magnitude, i.e. from an average error of 5.3\% to 0.63\%.}
    \label{fig:depth_net_results}
\end{figure}

Processing a single frame, i.e. forward and backward pass including the CNN, take approximately 800-900ms and once the CNN is trained, rendering a single frame, i.e. a forward pass of the CNN followed by expansion and rendering, takes approximately 400-450ms. The average MAE is 4\% and 5.6\% on training and test set, respectively. Employing the Neural Network to predict optimal $p$ and $w$ that induce a given depth field therefore results in an average speed up of approximately $190\times$ at the cost of increasing the error about 9 fold. Figure \ref{fig:depth_net_results} shows the results on four different frames on the test set. The CNN architecture is simple and can most likely be improved significantly by future work.

\subsubsection{Sensor fusion: RGBD}

\begin{figure}[h]
    \centering
    \includegraphics[width=\linewidth]{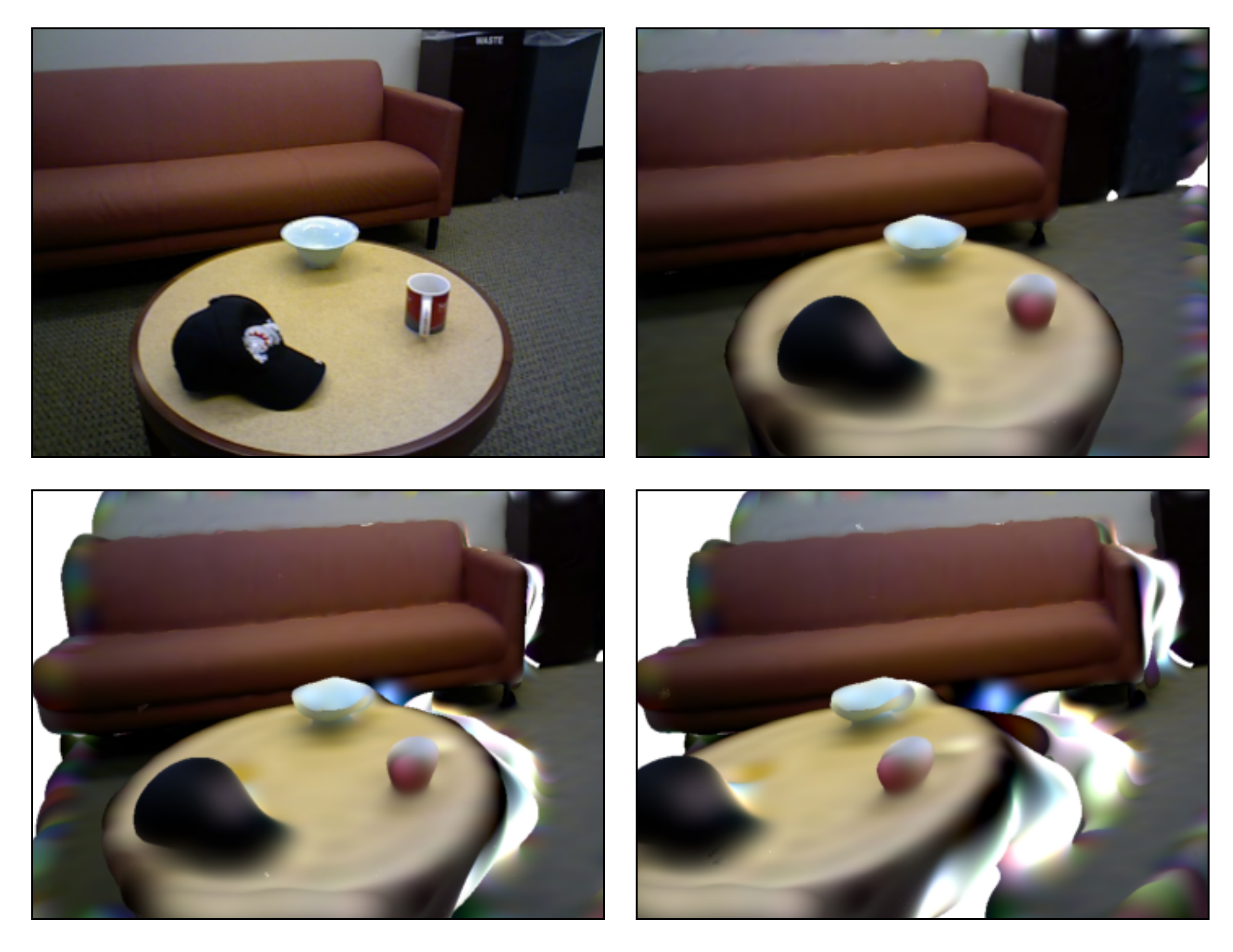}
    \caption{Top left image shows the groud truth RGB data. To the right of the ground truth is the output of the RGBD layer from the same view point as the ground truth. The two frames at the bottom show renderings from novel view points.}
    \label{fig:rgbd_couch}
\end{figure}

In this experiment, we combine the explicit and the depth layer to represent images collected with RGBD cameras. We make use of a single frame collected for the dataset described in ~\citep{lai2014unsupervised}. The dataset contains depth and color information. We model depth data with the proposed depth layer and additionally make use of the explicit layer introduced earlier to model color. Specifically, the combined layer outputs depth and color at the root. Training on a single frame takes approximately 2.5min from scratch but could potentially be sped up by a Neural Network in a similar fashion as described in the previous experiment. Figure \ref{fig:rgbd_couch} summarizes the results and show renderings of the RGBD image from novel view points.

\subsubsection{Inverse rendering}

\begin{figure}
    \centering
    \includegraphics[width=\linewidth]{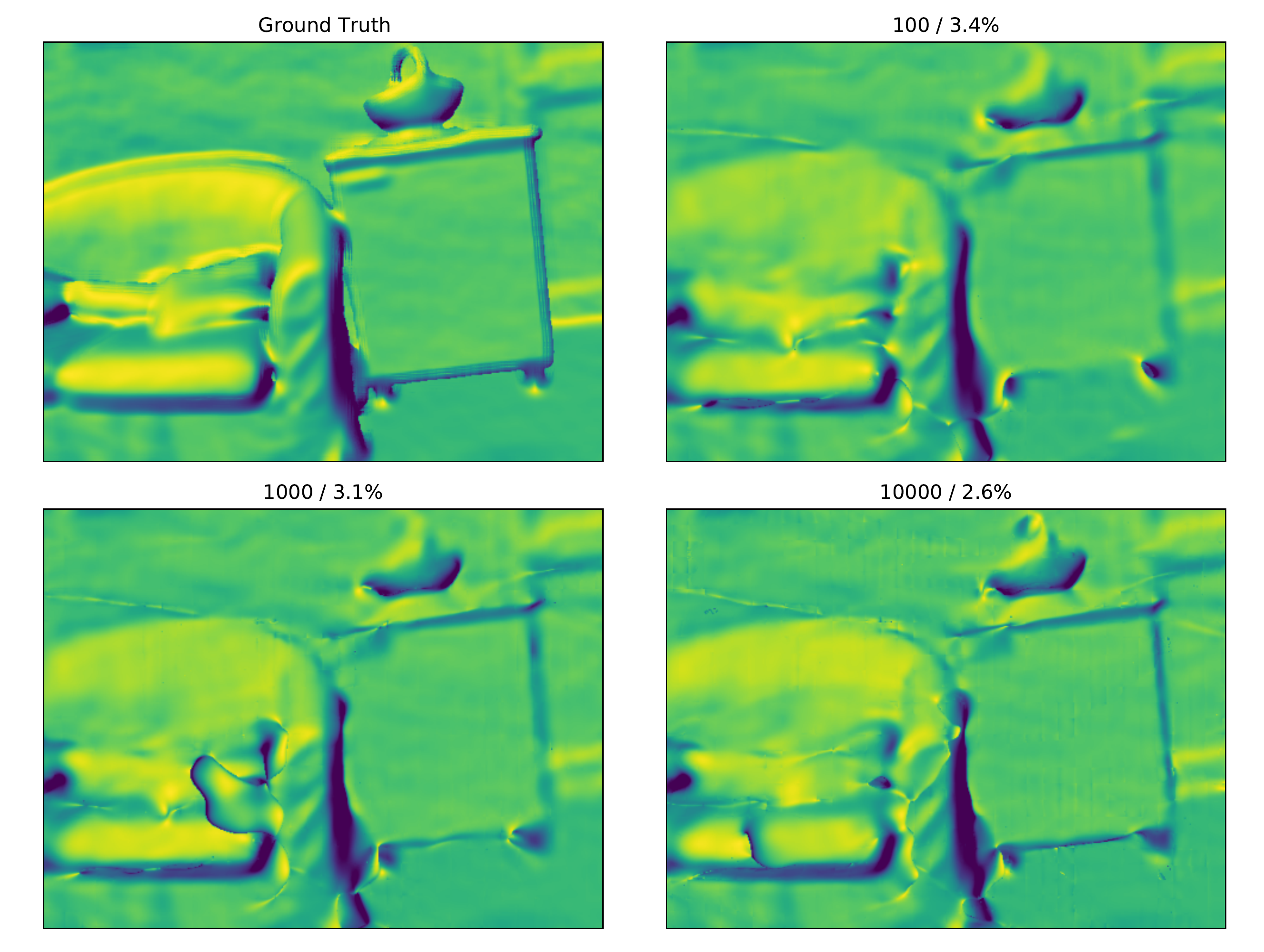}
    \caption{Results of the surface gradient layers after 100, 1000 and 10000 epochs respectively. The top left frame shows the ground truth.}
    \label{fig:boring}
\end{figure}

In order to showcase the performance of the layer that outputs surface gradients at a root, we conduct an arguably very simple experiment in the realm of Inverse Rendering~\citep{marschner1998inverse}. We make use of use of a datum from the `Reconstruction Meets Recognition Challenge 2014'~\citep{lai2014unsupervised}. The data set contains ground truth measurements of surface normals extracted from RGBD data collected with a Microsoft Kinect sensor. We `render' the surface normals by assuming a single light source and no color, i.e. the resulting image contains a single value per pixel that constitutes the dot product of the surface normal and the imaginary light source. Because the proposed layer outputs surface gradients as opposed to normals, the output is first normalized before it is dot-multiplied by a free parameter describing a light source. We minimize the mean absolute error for 10,000 epochs over the entire image of resolution $420 \times 560$. Because the image does not contain much detail, we choose a grid granularity level of 5 with a Gaussian kernel ($\alpha = 1000$). A single epoch takes approximately 600ms which entails a total training time of approximately 100min. Even though a single epoch is reasonably fast, we found the model to converge slowly to a solution with limited but reasonable accuracy with an average error of $2.6\%$. Figure \ref{fig:boring} shows the results at various stages of training. We reduce the learning rate during training and the reader is referred to the code for implementation details.

\newpage
\section{Integral Implicit Taylor Layer}
In the following, integral implicit layers are introduced and strategies to approximate the JVPs required for Gradient Based Learning are derived. The layer outputs line integrals along a ray, i.e. similar to the root implicit layer, let $\eye$ and $\mathbf{r}$ be a position and direction vector that encode the position and viewing direction of a pinhole camera respectively. We begin by introducing fast algorithms that allow for the analytic computation of integrals along rays that act directly on the intermediate Taylor representation. 

\subsection{Line Integration}
\subsubsection{Line Integral}
We start with a simple integral of the type below and build on top of the findings:
\begin{align*}
    y(p,w) = \int_0^{\infty} f(\eye + \mathbf{r}x; p, w) dx
\end{align*}
Analogously to the root finding algorithm, we iterate over all boxes that the ray intersects with and convert the ray through the box to a univariate polynomial as described in Figure \ref{fig:root_finding}. Essentially, the function along the ray is a piece-wise polynomial and its integral can be computed by splitting the integral at ray-box intersections, i.e. $\int_a^b f(x)dx = \int_a^n f(x)dx + \int_n^b f(x)dx$. Assume a function \verb+intersect+ that returns two scalars $x_1$ and $x_2$ such that $\eye + x_1 \mathbf{r}$ and $\eye + x_2 \mathbf{r}$ describe ray-box intersection points. We assume familiarity with basic polynomial arithmetic such as evaluation, integration, addition, multiplication and composition. The pseudo-code below shows how a simple line integral can be computed along a ray:
\begin{minted}{python}
    def integrate_ray(r, L):
        D = 0
        for box on ray:
            x1,x2 = intersect(box, r*x + eye)
            s = x2-x1
            d = x1*r + eye - box.center
            poly = line2poly(d,r,L[box])
            D += val(integrate(poly),s)
        return D
\end{minted}
Note that the similarly to the root implicit layer, computing the output of the forward pass does not make use of the $L2P$ procedure and that the FLOPs required for the polynomial arithmetic are minimal. Assuming $\rho=4$, integrations requires 5 FLOPs whereas evaluating the integral requires 17 FLOPs per box. 

\subsubsection{Volumetric Rendering Integral}
Recently, volumetric rendering has experienced a resurgence in popularity due to the success of Neural Radiance Fields (NeRF)~\citep{mildenhall2020nerf}. The volumetric rendering equation is a specific type of line integral defined as follows:

\begin{align*}
    y(p,w) &= \int_0^\infty T(x)\sigma(x)c(x) dx\\
    \sigma(x) &= f(\eye + \mathbf{r}x; p, w^{(1)}); w^{(1)} \in R^{N \times 1}\\
    c(x) &= f(\eye + \mathbf{r}x; p, w^{(2)}); w^{(2)} \in R^{N \times 3}\\
    T(x) &= \exp\left[-\int_0^x \sigma(x') dx'\right]\\
    \text{with: } \sigma(x) &> 0
\end{align*}

In this case, $\sigma(x)$ and $c(x)$ describe particle density and particle color at a specific spatial location. $T(x)$ can be interpreted as the probability that the ray has not yet hit a particle. For most rays, $T(x)$ decreases monotonically to 0 and ensures that the camera cannot see past objects or through dense fog. Traditionally volumetric rendering was employed to render effects like fog, smoke or steam but is nowadays also used to render solid objects for which $\sigma(x)$ increases sharply at object surfaces. One of the challenges of computing the volumetric rendering integral is the $T(x)$ term as there is no analytic (and therefore fast) solution to $exp[-f(x)]$ when $f(x)$ is a polynomial. We alleviate this difficulty by approximating $exp[-x]$ for $0 \leq x \leq 5.$ by a polynomial and assume that $exp[-x] = 0$ for $x > 5.$. Let \verb$mexp_poly$ be the coefficients of this polynomial approximation of order 4. The following pseudo-code shows how the volumetric rendering integral can be computed \emph{almost} exactly without the need for numerical integration based on the $FC^2T^2$ expansion:
\begin{minted}{python}
    def box_polys_fwd(s_poly, c_polys, D):
        Sig = integrate(s_poly + D)
        T_poly = compose(Sig, mexp_poly)
        for c in [R,G,B]:
            C_polys[c] = integrate(mul(mul(c_poly[c], s_poly),T_Tpoly))
        return C_polys, Sig
\end{minted}
Let \verb+s_poly+ and \verb+c_poly+ be the polynomials describing $\sigma(x)$ and $c(x)$ respectively, then:
\begin{samepage}
\begin{minted}{python}
    def integrate_ray(ray, L):
        C = [0,0,0]
        D_ = 0 #accumulator for density
        for box on ray:
            x1,x2 = intersect(box, r*x + eye)
            s = x2-x1
            d = l1*r + eye - box.center
            s_poly, c_polys = line2poly(d,r,L[box])
            C_polys, Sig = box_polys_fwd(s_poly, c_polys, D_)
            for c in [R,G,B]:
                C[c] += val(C_polys[c],s)
            D_ = val(Sig,s) #no need to increment because D_ was added in box_polys
            if D_ > 4.5: #poly fit to exp(-x) is only valid until 5
                break
        return C, D
\end{minted}
\end{samepage}

\subsection{Jacobian Vector Product}

In the context of Gradient Based Learning, the previous section introduced efficient algorithms to evaluate the forward pass of a computational layer that outputs integrals along rays. In the following, strategies to approximate the associated JVPs are introduced. As we will show shortly, computing these quantities is not trivial but nevertheless possible by making use of the $FC^2T^2$ expansion. Interestingly, these strategies do not require the $P2M$ procedure, i.e. at no point are particles inserted into the expansion.

\subsubsection{Line Integral JVP}

Again, we start by deriving the JVP for the simple line integral of the type below and build on these findings:
\begin{align*}
    y(p,w) = \int_0^{\infty} f(\eye + \mathbf{r}x; p, w) dx
\end{align*}
For the backward pass, we need to approximate the following JVPs:
\begin{align*}
    \overline{w} = \overline{y} \fracpartial{y}{w}; \ \ \overline{p} = \overline{y} \fracpartial{y}{p}
\end{align*}
Because $\overline{p}$ is analogous, we focus on $\overline{w}$:
\begin{align*}
    \overline{w} = \overline{y} \fracpartial{y}{w} \text{\ with\ } \fracpartial{y}{w} = \fracpartial{\int_0^{\infty} f(\eye + \mathbf{r}x; p, w) dx}{w}
\end{align*}
Expanding the Jacobian yields:
\begin{align*}
    &\fracpartial{\int_0^{\infty} f(\eye + \mathbf{r}x; p, w)dx}{w} = \begin{bmatrix}
    \int_0^\infty \phi(\eye + \mathbf{r}x, p_1)dx\\
    \int_0^\infty \phi(\eye + \mathbf{r}x, p_2)dx\\
    \dots \\
    \int_0^\infty \phi(\eye + \mathbf{r}x, p_N)dx
    \end{bmatrix},\\
    &\overline{w} = \int_0^\infty f(p; \eye + \mathbf{r}x, \overline{y})dx
\end{align*}

However, the resulting expression for $\overline{w}$ seems problematic because the integration variable appears behind the semi-colon of $f$, i.e. it acts on quantities required to compute the expansion. How can such an integral be evaluated?\\

In order to understand the intuition behind our solution, let us rewrite the integral as a summation, i.e.:
\begin{align*}
    &\overline{w} = \lim_{\Delta x \rightarrow 0} \sum_{i=0}^\infty f(p; \eye + \mathbf{r}(i \Delta x), \overline{y}) \Delta x
\end{align*}

In essence, this expression tells us that we would need to compute infinitely-many expansions where for each expansion, a single point would be inserted at location $\eye + \mathbf{r}(i \Delta x)$ with weight $\overline{y}$. $x$ would be incremented infinitesimally from expansion to expansion. Considering that computing expansions is computationally expensive, how can this insight help us in approximating $\overline{w}$? Recall that Taylor expansions are additive and that we can collect all of the infinitely-many expansions into one. The trick to do this analytically and therefore efficiently is to `hack' the $P2M$-step of the algorithm. Instead of inserting a single point, we insert infinitely-many points along a ray, i.e. we insert an entire ray into the $\mathcal{M}$-expansion.\\

Recall that for the regular $P2M$-step, for the box centered at $p'$ the following quantity is inserted: $$\mathcal{M}(\mathbf{n}, p') = \sum_{i \in I(p')} (p_{i,1}- p'_{1})^{n_1}(p_{i,2}-p'_{2})^{n_2}( p_{i,3} - p'_{3})^{n_3} w_i$$ Thus, if we wanted to insert infinitely many points along a ray into an infinitely wide box with weight $\overline{y}$, we would need to insert:
\begin{align*}
    \mathcal{M}(\mathbf{n}, p') &= \lim_{\Delta x \rightarrow 0} \sum_{i=0}^\infty (o_1 + (i \Delta x) r_1- p'_{1})^{n_1}(o_2 + (i \Delta x) r_2 - p'_{2})^{n_2}(o_3 + (i \Delta x) r_3 - p'_{3})^{n_3} \overline{y} \Delta x\\
    & = \overline{y} \int_{0}^{\infty} (o_1 + x r_1- p'_{1})^{n_1}(o_2 + x r_2 - p'_{2})^{n_2}(o_3 + x r_3 - p'_{3})^{n_3} dx
\end{align*}
Applying the binomial theorem and assuming that the first intersection of ray and box is located at $\mathbf{d}$ and the length of the ray segment is $s$ gives:
\begin{align*}
    \mathcal{M}(\mathbf{n}, p') &= \overline{y} \int_{0}^{s} (x r_1 + d_1)^{n_1}(x r_2 + d_{2})^{n_2}(x r_3 + d_{3})^{n_3} dx\\
    &= \overline{y} \int_{0}^{s} \sum _{i,j,k=0}^{n_1,n_2,n_3}{n_1 \choose i}d_1^{n_1-i}r_1^{i} {n_2 \choose j}d_2^{n_2-j}r_2^{j}{n_3 \choose k}d_3^{n_3-k}r_3^{k} x^{i+j+k} dx\\
    & = \overline{y} \sum_{i,j,k=0}^{n_1,n_2,n_3} \frac{{n_1 \choose i}{n_2 \choose j}{n_3 \choose k}}{i+j+k+1} d_1^{n_1-i}r_1^{i} d_2^{n_2-j}r_2^{j}d_3^{n_3-k}r_3^{k} s^{i+j+k+1}
\end{align*}

Let \verb$line2taylor(d,r,s)$ be a function that evaluates the line integral to be inserted for all $\mathbf{n}$. The following pseudo-code summarizes the $ray2M$ procedure:
\begin{minted}{python}
    def ray2M(ray, dy, M):
        for box on ray:
            x1,x2 = intersect(box, r*x + eye)
            s = x2-x1
            q = l1*r + eye
            M[box] += line2taylor(q-box.center,r,s)
        return M
        
    M = empty M-expansion
    for (ray,dy) in (rays,dys):
        M+= ray2M_h(ray,dy, L_fwd, M, func) 
\end{minted}

Once the $ray2M$-step has been carried out for all rays, the resulting $\mathcal{M}$-expansion can be transformed into a $\mathcal{L}$-expansion with the regular machinery described in section \ref{sec:fc2t2}. Obtaining $\overline{p}$ and $\overline{w}$ remains unchanged as well, i.e. in order to obtain JVPs the formulas introduced in \ref{jvp:explicit} are still valid.

\subsubsection{Volumetric Rendering JVP}

Deriving the JVPs for the volumetric rendering integral follows a similar strategy. By application of the chain-rule, the Jacobians for volumetric rendering can be derived as:
\begin{align}
    \fracpartial{y}{w^{(2)}} &= \int_0^\infty\underbrace{\fracpartial{c(x)}{w^{(2)}} \sigma(x)T(x) }_{\text{Case 1}} \label{eq:partialw2}dx\\
    \fracpartial{y}{w^{(1)}} &= \int_0^\infty\underbrace{\fracpartial{T(x)}{w^{(1)}} \sigma(x)c(x)}_{\text{Case 2a}} + \underbrace{\fracpartial{\sigma(x)}{w^{(1)}}T(x)c(x)}_{\text{Case 2b}} dx
\end{align}

As shorthand, let $\bx = \eye + x \mathbf{r}$ and $\bx'$ analogously.

\textbf{Case 1:}\\
This case can be summarized as computing $\int_0^{\infty} f(\bx; p, w) h(x)dx$. Following the strategy of $ray2M$ and expanding the Jacobian:

\begin{align*}
    &\fracpartial{\int_0^{\infty} f(\bx; p, w)h(x)dx}{w^{(2)}} = \begin{bmatrix}
    \int_0^\infty \phi(\bx, p_1)h(x)dx\\
    \int_0^\infty \phi(\bx, p_2)h(x)dx\\
    \dots \\
    \int_0^\infty \phi(\bx, p_N)h(x)dx
    \end{bmatrix},\\
    &\overline{w}^{(2)} = \int_0^\infty f(p; \bx, \overline{y}h(x))dx
\end{align*}

Thus, now the integration variable $x$ appears in both arguments behind the semi-colon. However, as long as $h(x)$ is a polynomial (or can quickly be approximated by one), this makes computing the JVP only marginally more difficult. Instead of inserting infinitely-many points with a fixed value $\overline{y}$ into the expansion, this value is weighted by $h(x)$ for all $x$ along the ray. Let $h(x) = \sum_{m=0} h_m x^m$.

The quantity needed to be inserted can therefore be derived as:
\begin{align*}
    \mathcal{M}(\mathbf{n}, p') &= \overline{y} \int_{0}^{s} (x r_1 + d_1)^{n_1}(x r_2 + d_{2})^{n_2}(x r_3 + d_{3})^{n_3} h(x) dx\\
    & = \overline{y} \sum_{i,j,k=0}^{n_1,n_2,n_3} \frac{{n_1 \choose i}{n_2 \choose j}{n_3 \choose k}}{i+j+k+1} d_1^{n_1-i}r_1^{i} d_2^{n_2-j}r_2^{j}d_3^{n_3-k}r_3^{k} \sum_{m=0} s^{m+i+j+k+1} h_m
\end{align*}

Analogously, let \verb$line2taylor_h(d,r,s,h)$ be a function that evaluates the line integral to be inserted for all $\mathbf{n}$. The following pseudo-code summarizes the weighted $ray2M$ procedure assuming that \verb+func+ returns the coefficients of $h(x)$:

\begin{samepage}
\begin{minted}{python}
    def ray2M_h(ray, dy, L_fwd, M, func):
        #L_fwd is the cached Taylor expansion cached from the forward pass
        for box on ray: #in any order
            x1,x2 = intersect(box, r*x + eye)
            s = x2-x1
            q = l1*r + eye
            s_poly, c_polys = line2poly(q-box.center,r,L_fwd[box])
            h_poly = func(s_poly, c_polys)
            M[box] += line2taylor_h(q-box.center,r,s, h_poly)*dy
        return M
        
    M = empty M-expansion
    for (ray,dy) in (rays,dys):
        M+= ray2M_h(ray,dy, L_fwd, M, func) 
\end{minted}
\end{samepage}

\textbf{Case 2:}\\
Case 2a deals with the case of computing the Jacobian w.r.t. $$\int_0^{\infty} \exp\left[-\int_0^x f(\bx'; p, w) dx'\right] h'(x)dx$$

Let $h(x) = T(x)h'(x)$, then,
\begin{align*}
   \fracpartial{\int_0^{\infty} \exp\left[-\int_0^x f(\bx'; p, w) dx'\right] }{w}h'(x)dx &= - \int_0^{\infty} \fracpartial{\int_0^x f(\bx'; p, w)dx'}{w^{(1)}} h(x) dx\\
   &= -\begin{bmatrix}
    \int_0^\infty \int_0^x\phi(\bx', p_1))dx' h(x)dx\\
    \int_0^\infty \int_0^x\phi(\bx', p_2))dx' h(x)dx\\
    \dots \\
    \int_0^\infty \int_0^x\phi(\bx', p_N)dx' h(x)dx
    \end{bmatrix},\\
    \overline{w} =& -\int_0^\infty \int_0^x f(p; \bx', \overline{y}h(x))dx' dx
\end{align*}

Now we are confronted with solving a double integral. However, with a few simple manipulations, this double integral can be brought into the form of Case 1 and can therefore be dealt with in the same manner. We begin by changing the order of integration:
\begin{align*}
    \overline{w}^{(1)} =& -\int_0^\infty \int_0^x f(p; \bx', \overline{y}h(x))dx' dx\\
        =& -\int_0^\infty  f(p; \bx', \overline{y}\int_x^\infty h(x)dx) dx' 
\end{align*}
 Expanding $h(x)$ yields:
 \begin{align*}
     \int_x^\infty h(x')dx' &= \int_x^\infty T(x')\sigma(x')\mathbf{c}(x')dx'\\
    &= \underbrace{\int_0^\infty T(x')\sigma(x')\mathbf{c}(x')dx'}_{= \mathbf{C} \text{\ (from forward pass)}} - \int_0^x T(x')\sigma(x')\mathbf{c}(x')dx'\\
    &= \mathbf{C} - \int_0^x T(x')\sigma(x')\mathbf{c}(x')dx' = k(x)
 \end{align*}

Swapping the integration direction from $x \rightarrow \infty$ to $0 \rightarrow x$ avoids the awkward problem of having to compute an acausal quantity, i.e. a quantity that requires knowledge of the `future' of the ray.

Combining this with Case 2b yields:
\begin{align*}
    \overline{w}^{(1)} &= \int_0^\infty f(p; \bx, \overline{y} T(x)c(x))  - \int_0^\infty f(p; \bx, \overline{y} k(x)) dx\\
    &= \int_0^\infty f(p; \bx, \overline{y} (T(x)c(x) - k(x))) dx \\
    &= \int_0^\infty f\left(p; \bx, \overline{y} \left(\int_0^x T(x')\sigma(x')\mathbf{c}(x')dx' + T(x)c(x) - \mathbf{C}\right)\right) dx
\end{align*}

In contrast to the Case 1, in order to solve Case 2, a ray-integral is inserted into the $\mathcal{M}$-expansion for the backward pass. Note that the resulting algorithm bears resemblance to the adjoint sensitivity method~\citep{cao2003adjoint} that finds application in the NeuralODE~\citep{chen2018neural} algorithm. Similar to the adjoint method, in order to perform the backward pass, a problem of similar difficulty as the forward pass needs to be solved. This is in contrast to the root-implicit layers introduce in the previous section whose backward pass is considerably faster and easier to evaluate as it only requires a simple projection step.

The derivations lead to the following algorithms in pseudo-code:
\begin{minted}{python}
    def box_polys_bwd(s_poly, c_polys, D):
        Sig = integrate(s_poly + D)
        Tpoly = compose(Sig, mexp_poly)
        Tc_polys = [mul(c_poly[c], T_poly) for c in [R,B,G]]
        Tsig_poly = mul(s_poly, T_poly)
        return Tc_polys, Tsig_poly
\end{minted}

The output of the forward pass is a tuple representing RGB color for every ray. This entails that the tangent at which the JVP is evaluated is also a three-tuple representing the error for each color channel. The following pseudo-code assumes that the tangent vector \verb+dy+ and the $\mathcal{L}/\mathcal{M}$-expansions for the four channels representing RGB and density can be accessed with square bracket notation, i.e. \verb+M[box,D]+ represent the $\mathcal{M}$-expansions for the density channel and $\verb+dy[R]+$ the error for the red-channel.

\begin{samepage}
\begin{minted}{python}
    def integral_ray2M(ray, L_fwd, M,
                       C_fwd, #result of forward pass
                       dy):
        C_bwd = [0,0,0]
        D_ = 0 #accumulator for density
        for box on ray: #since we integrate, order matters
            l1,l2 = intersect(box, r*x + eye)
            s = l2-l1
            d = l1*r + eye - box.center
            #generating the required polynomials
            s_poly, c_polys = line2poly(d,r,L_fwd[box])
            C_polys, Sig = box_polys_fwd(s_poly, c_polys, D_)
            Tc_polys, Tsig_poly = box_polys_bwd(s_poly, c_polys, D_)
            for c in [R,G,B]:
                #Case 1:
                M[box,c] += line2taylor_h(d,r,s,Tsig_poly)*dy[c]
                #Case 2:
                h_poly = Tc_polys[c] + C_polys[c] + C_bwd[c] - C_fwd[c]
                M[box,D] += line2taylor_h(d,r,s,h_poly)*dy[c]
                C_bwd[c] += val(C_polys[c],s)
            D_ = val(Sig,s) #D was added in box_polys_fwd
            if D_ > 4.5: #poly fit to exp(-x) is only valid until 5
                break
        return M
\end{minted}
\end{samepage}

\textbf{Incorporating background color:}\\
In many scenarios, because the domain is assumed to be finite, the likelihood that a ray has hit a particle might not be 1 or inversely, $\lim_{x \rightarrow \infty} T(x)$ is not 0. In these cases, one may want to incorporate knowledge about the ambient background color. Let $\mathbf{c_{bgr}}$ be the ambient background color then for the forward pass the following holds:
\begin{align*}
    y = \int_0^\infty T(x)\mathbf{c}(x)\sigma(x) dx + \mathbf{c}_{bgr}\lim_{x \rightarrow \infty} T(x)
\end{align*}

Thus, we need to additionally handle the Jacobian of $\fracpartial{\mathbf{c}_{bgr}\lim_{x \rightarrow \infty} T(x)}{w^{(1)}}$. This quantity is easy to derive and only affects the $\mathcal{M}$-expansion of the density channel:
\begin{align*}
    \overline{y} \fracpartial{\mathbf{c}_{bgr}\lim_{x \rightarrow \infty} T(x)}{w^{(1)}} = \bdot{\overline{y}}{\mathbf{c}_{bgr}} \lim_{x \rightarrow \infty} T(x)
\end{align*}
Note that $\lim_{x \rightarrow \infty} T(x) =: T_\infty$ is known from the forward pass and let \verb+bg_offset+ denote $\bdot{\overline{y}}{\mathbf{c}_{bgr}}T_\infty$. The only changes to the $ray2M$ procedure are therefore:

\verb#h_poly = h_poly + bg_offset#\\

\begin{figure}
    \centering
    \includegraphics[width=0.6\linewidth]{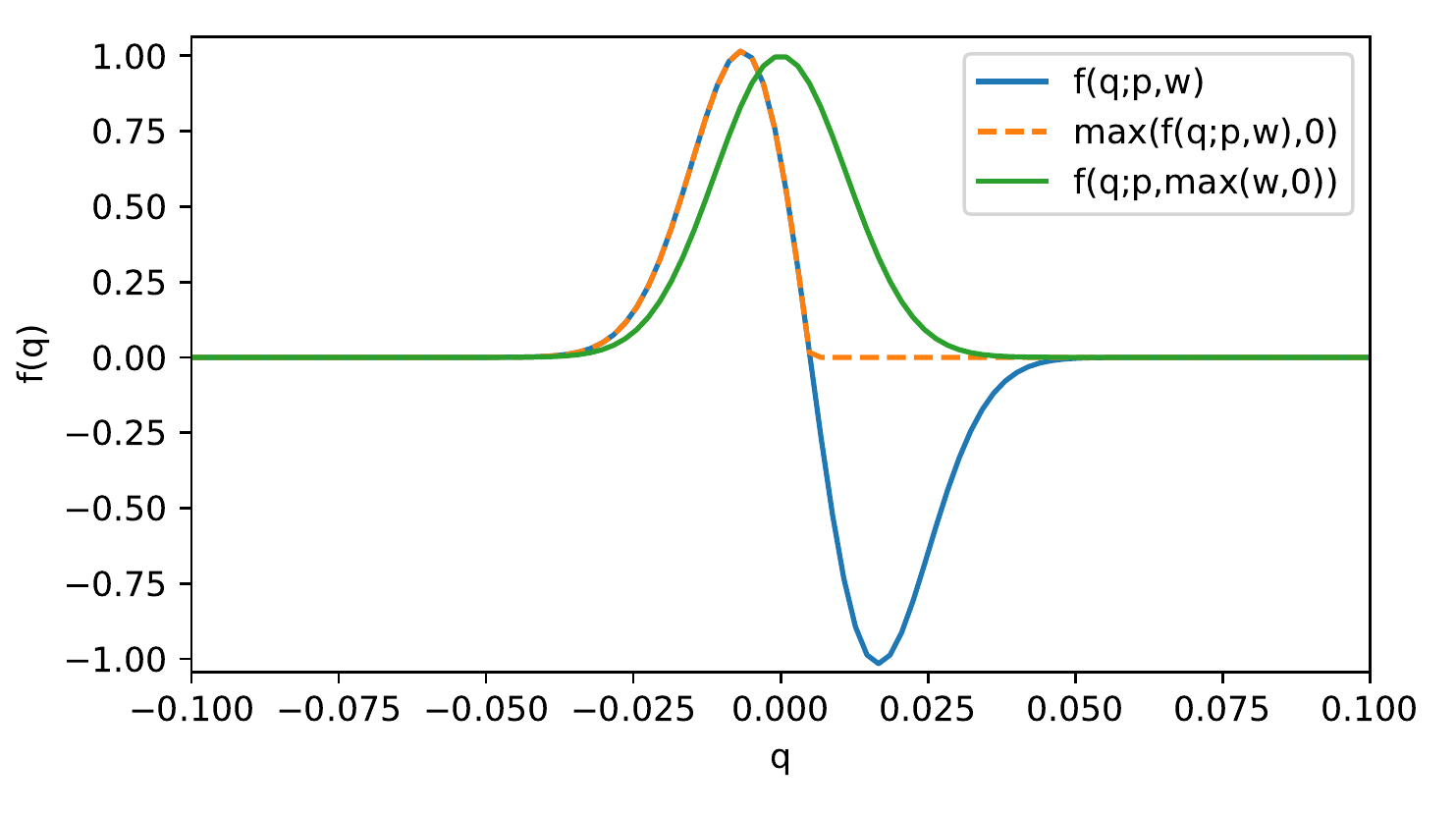}
    \caption{The effects of enforcing non-negativity in different ways. Applying the $relu$ non-linearity to particle weights $w$ instead of the output results in a loss of sharpness and more smoothness.}
    \label{fig:nonneg}
\end{figure}

\textbf{Enforcing non-negativity:}\\
As described earlier, the particle density must be non-negative, i.e. $\sigma(x) \geq 0$ for all $x$. We enforce this constraint in an arguably crude way by applying the $relu$ non-linearity to $w^{(1)}$. Note that this is not equivalent to applying the $relu$ non-linearity to $\sigma$. Figure \ref{fig:nonneg} sketches the effects of this and shows that when enforcing non-negative $\sigma(x)$ by ensuring that $w^{(1)}$ is non-negative sharpness is lost, i.e. $f(q; p,max(w,0))$ is in general smoother compared to $max(f(q; p,w),0)$ which might be discontinuous at zero-crossings of $f$.

\textbf{Counting FLOPs:}\\
For the forward pass, for every box, the Taylor expansion needs to be converted to a line polynomial for 4 channels. This requires $4 * 668FLOPs$. The polynomial arithmetic described in \verb+box_polys_fwd+ in turn requires $1,563FLOPs$ per box and ray. Assuming a level 6 expansion, this results in $1.6MFLOPs$ per ray in the worst case, i.e. approximately half of a single evaluation of the DeepSDF Neural Network. Assuming an $800 \times 800$ image this results in $1,040 GFLOPs$. When adding the cost for the expansion  ($4*154 GFLOPs$ or $4*1,317 GFLOPs$ depending on whether or not the kernel can be factored) this results in a total number of FLOPs of either $1,654$ or $6,308GLOPs$ in total.

The backward pass is slightly more expensive. Again, $4*668FLOPs$ are required for the conversion of the 4-channel Taylor expansion to line polynomials. However, the polynomial arithmetic in \verb+box_polys_bwd+ is more expensive at $2,452FLOPs$ per ray and box. Additionally, solving the integral for the $ray2M$ procedure incurs an additional cost of $1489FLOPs$ per ray and box. The worst case total number of FLOPs required for the backward pass assuming a level 6 expansion and a resolution of $800 \times 800$ is therefore $2,231/6,912GFLOPs$ depending on whether or not the kernel can be factored. 

The total number of FLOPs per iteration on an entire $800 \times 800$ image are therefore approximately $3.8$ or $13.3TFLOPs$ depending on the kernel.

\subsection{Applications}

\subsubsection{Radiance fields}

\begin{figure}[h]
    \centering
    \includegraphics[width=0.75\linewidth]{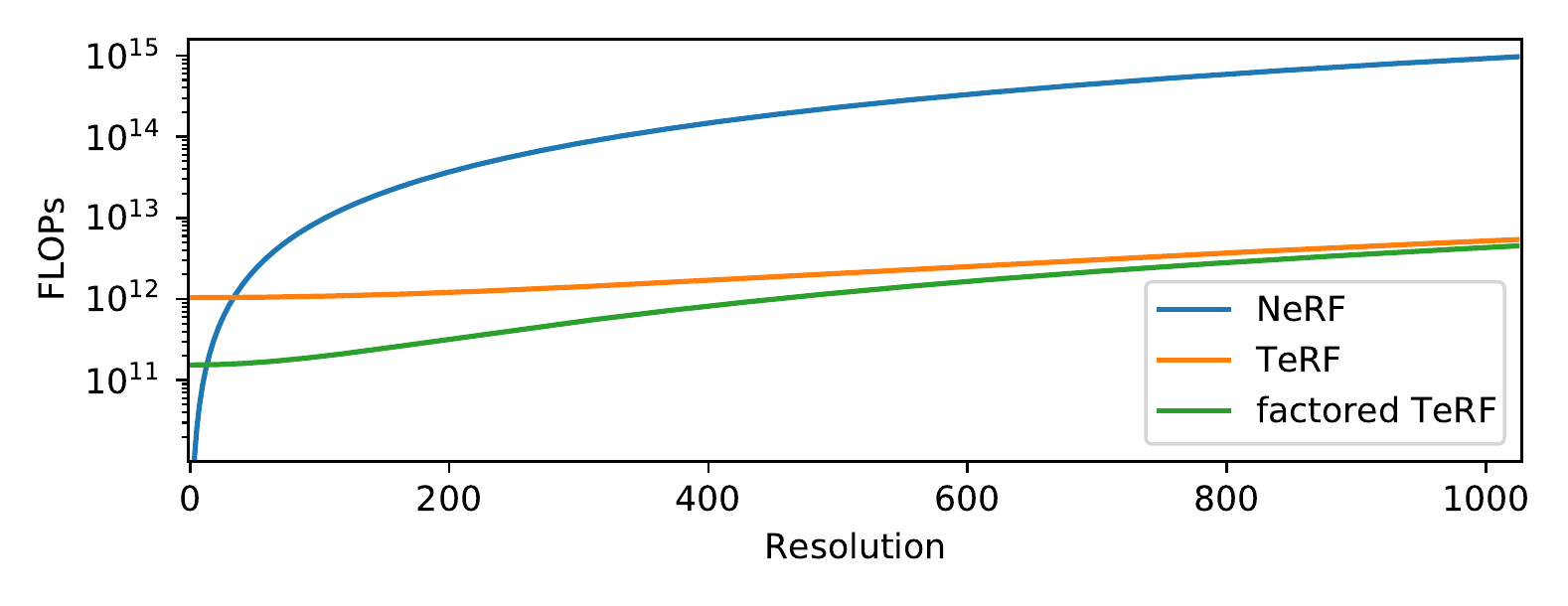}
    \caption{A comparison of the FLOPs required for the forward and backward pass as a function of the image's resolution. Note that the intercept for TeRF is non-zero because of the FLOPs required for the expansion and that the non-factored TeRF breaks even with NeRF at a resolution of $34\times 34$ or approximately 1000 rays.}
    \label{fig:nerf_terf_flops}
\end{figure}

We recreate the synthetic experiment proposed in ~\citep{mildenhall2020nerf}. The experiment assumes knowledge of a data set that contains tuples of RGB images and poses (a tuple of pinhole camera location and viewing direction). The goal is to infer a 3D representation of the object given this data set. The reader is referred to the original paper for details. In ~\citep{mildenhall2020nerf}, the authors propose NeRF, a neural network approach resembling the DeepSDF network~\citep{Park_2019_CVPR}. This Neural Network is trained by solving the volumetric rendering integrals numerically which results in many (usually 128) Neural Network evaluations per ray. The paper additionally proposes heuristics to encourage higher spatial frequencies and to reduce the FLOPs required for integration. We compare this approach to the integral implicit layer introduced earlier that we call TeRF. TeRF evaluates the integral analytically instead of numerically but is still approximate in nature as it makes use of the $FC^2T^2$ procedure internally. TeRF dramatically reduces the FLOPs required for the forward and backward pass. Assuming 128 evaluations per ray, the NeRF approach requires approximately $300 TFLOPs$ per pass so approximately $600TFLOPs$ in total. This entails a $45\times$ or $157\times$ reduction in FLOPs depending on whether or not the kernel can be factorized. Figure \ref{fig:nerf_terf_flops} compares NeRF and TeRF w.r.t. FLOPs for varying image resolutions.

\begin{figure}[h]
    \centering
    \begin{subfigure}[t]{0.48\textwidth}
        \centering
        \includegraphics[width=0.85\linewidth]{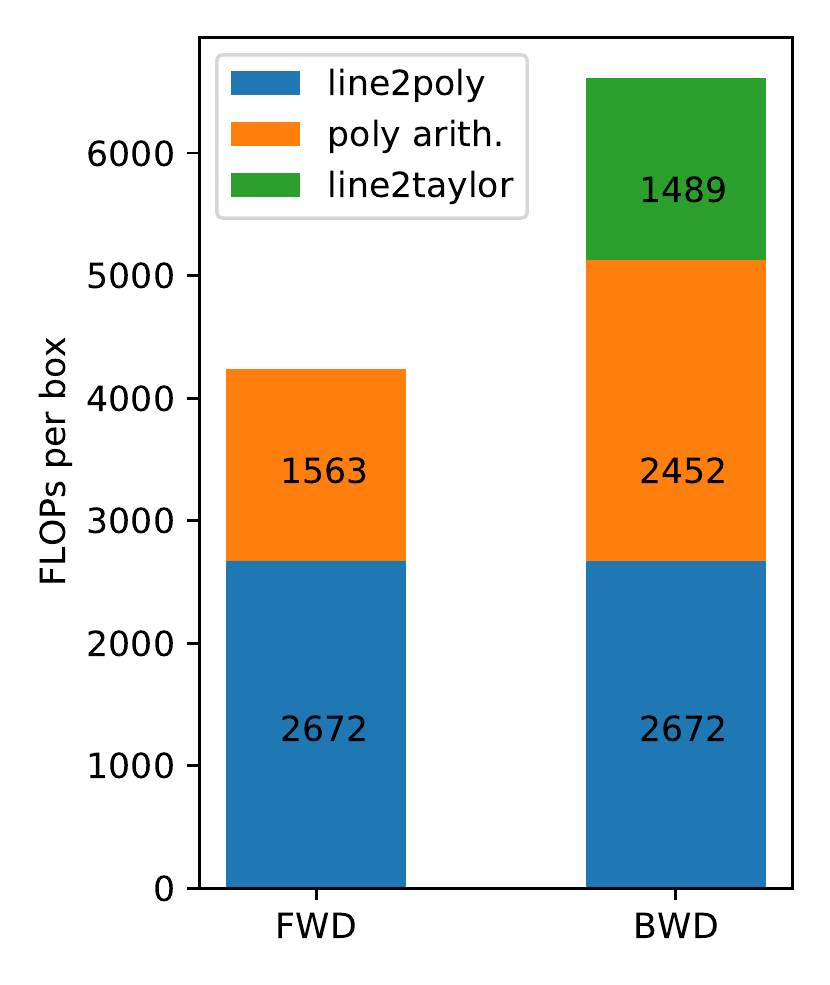} 
        \caption{A breakdown of the FLOPs required for the forward (FWD) and backward pass (BWD) per box and ray.} \label{fig:terf_flops}
    \end{subfigure}
    \hfill
    \begin{subfigure}[t]{0.48\textwidth}
        \centering
        \includegraphics[width=0.95\linewidth]{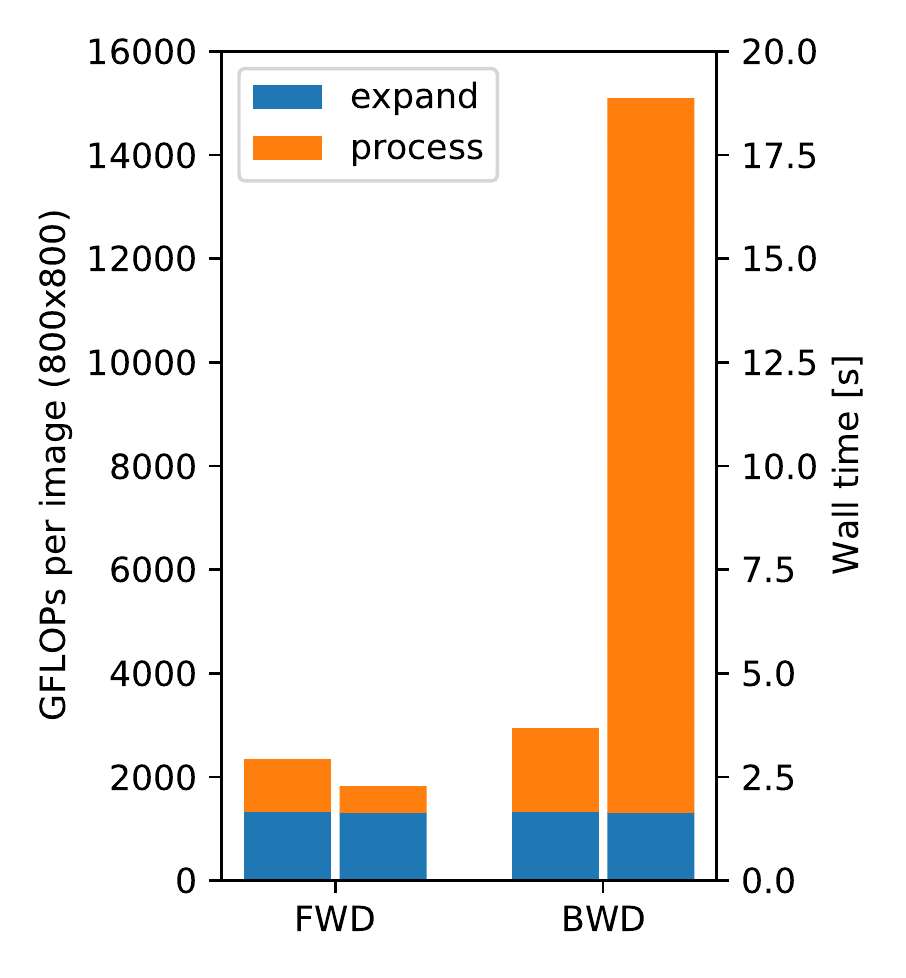} 
        \caption{Striped bars indicate wall time. Even though the backward pass is only slightly more expensive w.r.t. FLOPs, the required wall time is orders of magnitudes worse.} \label{fig:terf_wall}
    \end{subfigure}
    \caption{A comparison of the forward (FWD) and backward (BWD) pass of TeRF in terms of FLOPs and walltime. The current implementation of the backward pass is unfortunately very FLOP inefficient.}
    \label{fig:nerf_terf_walltime_flops}
\end{figure}

The current JAX implementation is unfortunately unable to capitalize on this reduction in FLOPs. Rendering an $800 \times 800$ image with NeRF takes approximately 35s compared to TeRF's 2.2s (1.6s for expanding and 610ms for rendering), a speed up of approximately only $14.5\times$. However, things get worse for the backward pass. Even though the FLOPs required for the backward pass only slightly increase compared to the forward pass, the wall time jumps from an expected 900ms to 17.5s resulting in a speed-up of only $2 \times$. Figure \ref{fig:nerf_terf_walltime_flops} shows a comparison of the forward and backward pass in terms of FLOPs and wall time. To make things even worse, we are unable to even $jit$-compile the loss function for JAX versions newer than $0.2.7$. At this moment, it is unclear whether or not an efficient implementation of TeRF is possible in JAX (or any other currently available high level general purpose GPU language).

\begin{figure}[h]
    \centering
    \includegraphics[width=0.75\linewidth]{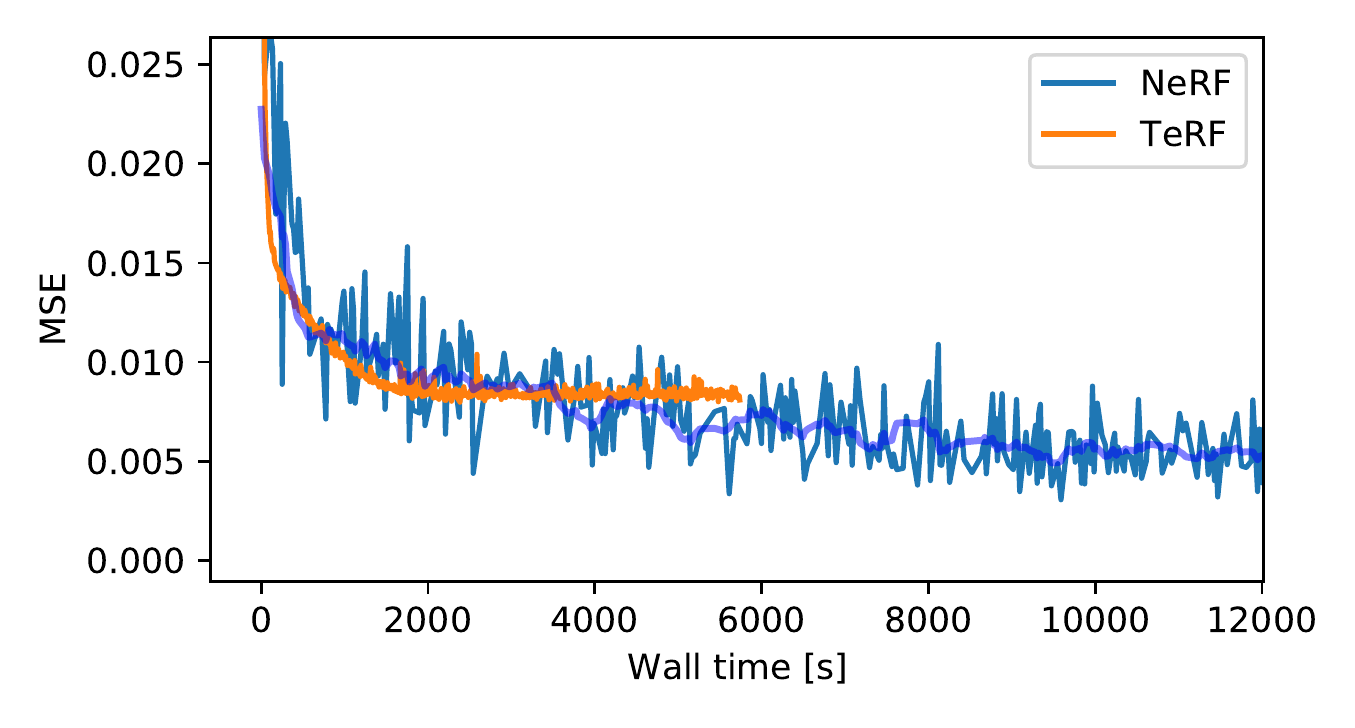}
    \caption{A comparison of NeRF and TeRF in terms of mean squared error as a function of wall time. TeRF converges slightly faster in the beginning but levels out soon.}
    \label{fig:nerf_terf_walltime_mse}
\end{figure}

\begin{figure}[h!]
    \centering
    \begin{subfigure}[t]{0.48\textwidth}
        \centering
        \includegraphics[width=0.95\linewidth]{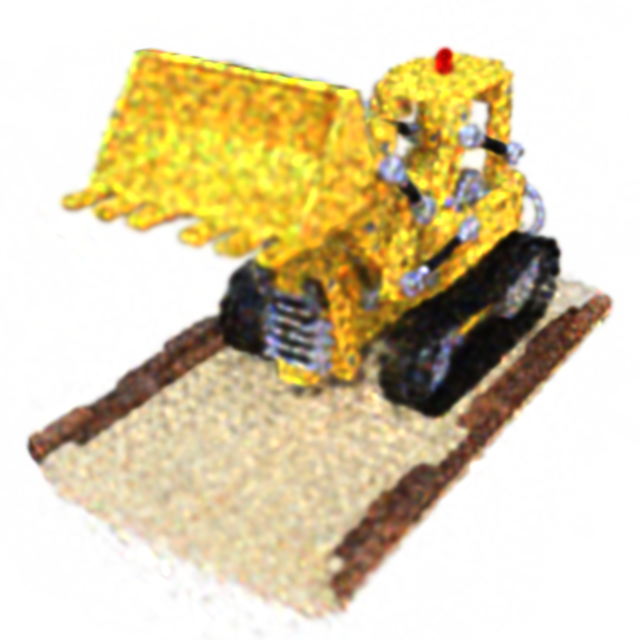}
        \caption{TeRF at 10min, MSE of 0.0101: The image looks overly `blotchy'. The overall shape is clearly visible and some detail is present.} \label{fig:terf_10min}
    \end{subfigure}
    \hfill
    \begin{subfigure}[t]{0.48\textwidth}
        \centering
        \includegraphics[width=0.95\linewidth]{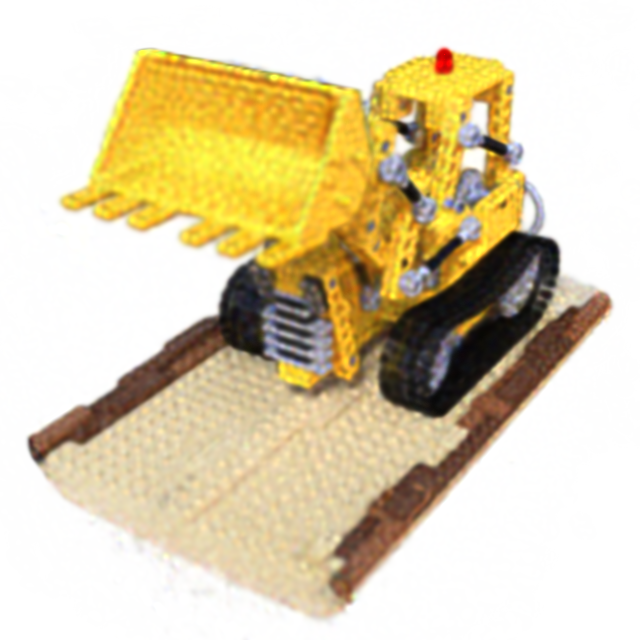}
        \caption{TeRF at 20min, MSE of 0.0087: The image looks less `blotchy' and some high frequency details are clearly visible as e.g. holes in the Lego pieces.} \label{fig:terf_20min}
    \end{subfigure}
    \centering
    \begin{subfigure}[t]{0.48\textwidth}
        \centering
        \includegraphics[width=0.95\linewidth]{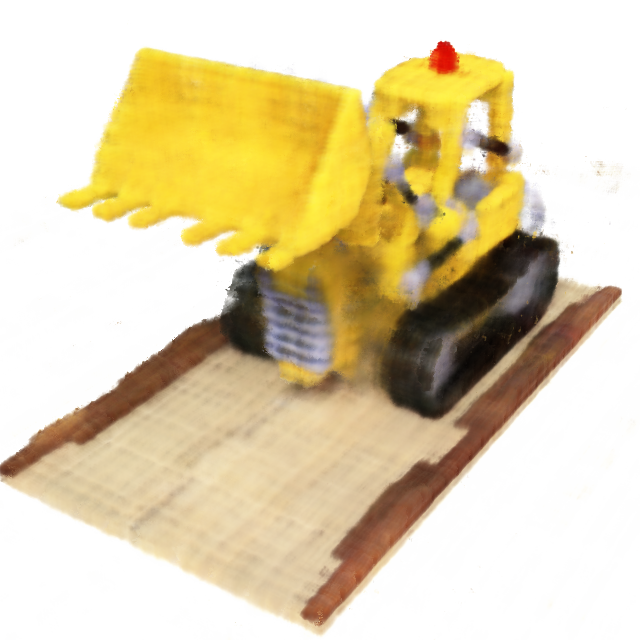}
        \caption{NeRF at 10min, MSE of 0.012: The image looks blurry but the overall shape is clearly visible. Some high frequency details seem visible.} \label{fig:nerf_10min}
    \end{subfigure}
    \hfill
    \begin{subfigure}[t]{0.48\textwidth}
        \centering
        \includegraphics[width=0.95\linewidth]{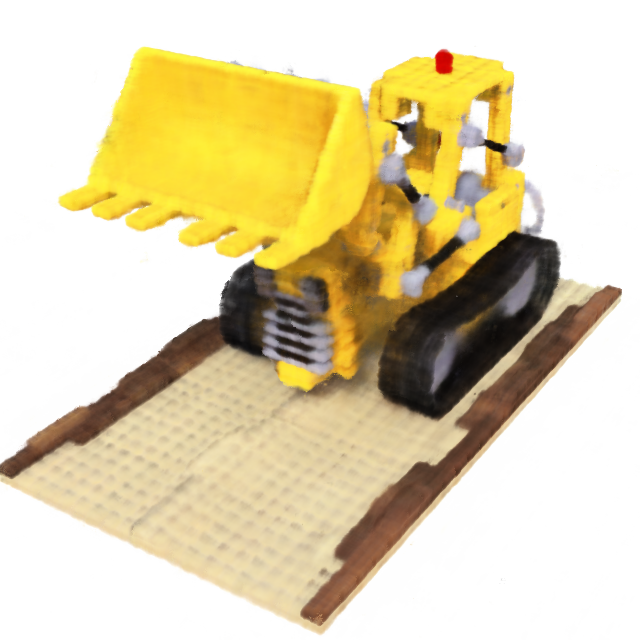}
        \caption{NeRF at 20min, MSE of 0.009: The image looks less blurry and more high frequency details seem visible. Details like holes in the Lego pieces seem missing.} \label{fig:nerf_20min}
    \end{subfigure}
    \caption{A comparison of NeRF and TeRF after 10 and 20 minutes of training. NeRF seems to show a bias toward smooth- or blurryness whereas TeRF seems biased towards `blotchy'ness or high frequency noise. TeRF seems to have learned more high frequency details at the end of training.}
    \label{fig:nerf_terf_20_10}
\end{figure}

NeRF produces high quality images but requires multiple hours of training. In Figure \ref{fig:nerf_terf_20_10}, we visually compare the output of NeRF and TeRF after 10 and 20 minutes of training respectively, i.e. long before NeRF has converged. The reader is referred to ~\citep{mildenhall2020nerf} for renderings of NeRF after convergence. In the experiment, because of the slow run time of TeRF's backward pass, we only process 100,000 rays per iteration (or per expansion). TeRF is approximately $4\times$ faster in processing 100,000 rays, however, this does not result in a $4\times$ reduction in wall time. In general, TeRF converges quickly but to a significantly worse solution compared to NeRF. Figure \ref{fig:nerf_terf_walltime_mse} shows the MSE of both approaches as a function of wall time. Even though not significant, TeRF seems to converge quicker in the beginning but levels out quickly. TeRF was trained with a level 6 expansion and 8m source points. However, after training only 2.35 out of the 8m source locations are associated with a non-negative density. Thus only about 35\% of the source points contribute to the spatial density distribution.

\subsection{Future Work}

\subsubsection{Non-Negativity Constraint}
As shown earlier, the Neural Network-based approach NeRF converges to a significantly better solution in terms of mean squared error. We conjecture this gap in accuracy not to stem from capacity or accuracy limitations of a level 6 expansion but rather from how non-negativity is enforced. Currently, non-negativity is enforced by ensuring non-negative weights $w$. We previously showed that this results in a reduction in sharpness and is most likely the cause for the 'pearls on a wireframe'-look that the rendering with the TeRF layer seem to exhibit. Ideally, one would allow for non-negative weights and enforce non-negative densities, i.e. the idea is to compute $\max(f(q;p,w),0)$ instead of computing $f(q;p,\max(w,0))$. During the forward pass, this is easy to implement by first finding roots and jumping to the next root if the function value is currently negative. However, this is not true for the backward pass. Whilst $g_1(x) = f(\eye + x\mathbf{r};p,\max(w,0))$ can be converted to a polynomial and therefore be inserted into an $\mathcal{M}$-expansion easily, the same is not true for $g_2(x) = \max(f(\eye + x\mathbf{r};p,w),0)$. It is not trivial to insert a clipped polynomial into an $\mathcal{M}$-expansion because the property of polynomial closure is lost, i.e. $g_2$ cannot be expressed as a polynomial. One possible way to circumvent this problem is to first perform a polynomial approximation of $f(x) = max(x,0)$ and compose the resulting polynomials. As polynomial composition increases the order, this strategy might lead to a significant increase in computational cost. Another strategy is to solve the rendering equation by sampling as proposed in the original NeRF paper. This strategy results in a significant memory overhead but allows to enforce non-negativity easily.

\subsubsection{Non-Lambertian Effects}
Currently, we assume the object to exclusively have Lambertian surfaces~\citep{basri2003lambertian} in the sense that its perceived color is invariant to the viewing direction, i.e. the surface is perfectly diffuse. This assumption is violated in the experiments conducted above, i.e. the Lego excavator has reflective surfaces. The question arises how to generalize TeRF to non-Lambertian effects. NeRF achieves this by parameterizing a 5D instead of 3D Neural Network. For TeRF such a strategy is most likely ill-advised because it would dramatically increase memory and computational requirements. One strategy could be to use a Neural Network to produce $w^{(2)}$, i.e. the source weights responsible for color, as a function of the viewing direction. This would result in an algorithm similar to the one presented in section \ref{learn_to_learn}. Instead of a CNN that takes a desired depth field as input, this Neural Network would take a 2D viewing direction as input and produce optimal $w^{(2)}$. It might be advisable to learn a perturbation from the Lambertian colors, i.e. $w^{(2)} = w^{(2)}_L + f(\mathbf{r})$ with $w^{(2)}_L$ denoting the Lambertian colors that TeRF currently extracts and $f(\mathbf{r})$ being the perturbation to these colors that result from the viewing direction $\mathbf{r}$.

\section{Conclusion}

This work is of exploratory nature and investigates the efficacy of fast summation algorithms and series expansions in the context of gradient based learning with applications to vision and graphics. We demonstrate the value of an intermediate Taylor representation of convolutional operators and introduce algorithms that act directly on `Taylor space' to approximate the quantities required for the forward and backward pass of the backpropagation algorithm. These algorithms offer enormous reductions in the computational resources required for training and inference. However, the current implementation of the techniques is unable to fully capitalize on these reductions. In the following, we summarize the FLOP efficiency of the current implementation of these layers on some of the tasks discussed in this paper.

\begin{table}[h]
\begin{tabular}{|l|l|l|l|l|}
\hline
\textbf{Task}        & \textbf{Explicit} & \textbf{Root-Impl.} & \textbf{Expl. + R-Impl.} & \textbf{Integral Impl.} \\ \hline
             &  \includegraphics[width=0.12\linewidth]{result_plots/einstein_6.png}        & \includegraphics[width=0.12\linewidth]{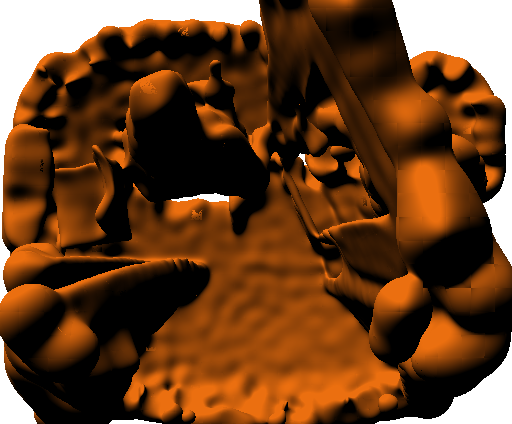}   & \includegraphics[width=0.12\linewidth]{result_plots/rbgd_placeholder.png}  & \includegraphics[width=0.12\linewidth]{result_plots/lego_placeholder.png}   \\ \hline
\textbf{TFLOPs/epoch} & 0.31               & 0.041           & 0.22          & 3.8          \\ \hline
\textbf{Time/epoch} (current)   & 1.2               & 0.24           & 0.7          & 21           \\ \hline
\textbf{Efficiency}   & 2.6\%              & 2.9\%        & 2.8\%       & 1.8\%        \\ \hline
\textbf{Time/epoch} (ideal)   & 0.03               & 0.01           & 0.02          & 0.38           \\ \hline
\end{tabular}
\caption{A comparison of the FLOP efficiency of the current implementation. An improved implementation of the techniques introduced in this paper can offer additional speed-ups of 30 to 90$\times$ depending on the task.}
\label{sadness}
\end{table}

We believe this work to be a first step. The current implementation should be considered a proof-of-concept but it shows that even a very simple variant of the techniques can already deliver significant reductions in computational cost. Furthermore, we believe that future improvements such as e.g. a sparse grid and variable kernels sizes can result in even greater speed-ups and improvements to accuracy. We believe that the techniques introduced in this paper can potentially allow for training radiance fields with a similar accuracy as the Neural Network approach in the matter of 1-2 minutes as opposed to multiple hours. However, this requires careful implementation of the techniques in low level GPU languages such as CUDA which is beyond the scope of this paper. If widely adopted, the techniques described in this paper could enable breakthroughs in the realms of computer vision or robotics. Note that the techniques introduced in this paper allow for controlling the computational cost in a straight-forward way which might enable applications on low-powered or resource constraint devices. We hope that this work is the beginning of a wider adoption and increased research activity of fast algorithms in the context of gradient based learning.

\acks{We would like to thank Shlok Mohta, Jeong Joon Park and Travis Askham for valuable discussions.}


\bibliography{sample}

\end{document}